\documentclass[conference]{IEEEtran}

\ifCLASSINFOpdf
\else
\fi

\usepackage{amsmath}
\usepackage{booktabs}
\usepackage{graphicx}
\usepackage{verbatim}
\usepackage{textgreek}

\usepackage{algorithm}
\usepackage{url}
\usepackage{algpseudocode}
\usepackage{multirow}
\usepackage[caption=false,font=footnotesize]{subfig}
\usepackage{algorithmicx}

\title{Evolving Spatial Weights for Cartographic Synthesis}

\author{
 
\IEEEauthorblockN{Gesiel R. Lopes\IEEEauthorrefmark{1},
Roberto F. da Silva\IEEEauthorrefmark{2},
Mellina Yamamura\IEEEauthorrefmark{3}},
\IEEEauthorblockN{Sergio H. V. L. de Mattos\IEEEauthorrefmark{4},
Antonio M. Saraiva\IEEEauthorrefmark{5}},
\IEEEauthorblockN{Alexandre C. B. Delbem\IEEEauthorrefmark{1},
Eric K. Tokuda\IEEEauthorrefmark{1}},
\IEEEauthorblockA{\IEEEauthorrefmark{1}Institute of Mathematics and Computer Sciences,
University of Sao Paulo, Sao Carlos, Brazil}
\IEEEauthorblockA{\IEEEauthorrefmark{2}Biosystems Engineering Department,
University of Sao Paulo, Piracicaba, Brazil}
\IEEEauthorblockA{\IEEEauthorrefmark{3}Department of Nursing,
Federal University of Sao Carlos, Sao Carlos, Brazil}
\IEEEauthorblockA{\IEEEauthorrefmark{4}Department of Hydrobiology,
Federal University of Sao Carlos, Sao Carlos, Brazil}
\IEEEauthorblockA{\IEEEauthorrefmark{5}Polytechnic School,
University of Sao Paulo, Sao Paulo, Brazil}
}

\begin{document}
\maketitle

\begin{abstract}
The integration of multiple thematic data layers into a single composite map, \emph{the cartographic synthesis problem}, is typically addressed through expert-driven weighting schemes. In this study, we present a multi-objective formulation of cartographic synthesis grounded in spatial autocorrelation structure. We develop a bi-objective evolutionary framework (GIS-moGA) that estimates layer weights by simultaneously maximizing global spatial structure (Global Moran's I) and minimizing local spatial heterogeneity (variance of Local Indicators of Spatial Association, LISA).
Because evaluating spatial relationships naively requires $\mathcal{O}(N^2)$ operations, direct computation becomes impractical for larger datasets. We address this challenge by leveraging the 97.7\% sparsity of Queen contiguity matrices, reducing effective complexity to $\mathcal{O}(N \cdot k)$ and enabling scalable municipal-level analysis.
The framework is evaluated using a high-dimensional spatial epidemiology dataset ($N=523$ units) from Araraquara, Brazil. A 64-scenario experimental design is used to examine evolutionary behavior across parameter settings. Results indicate that higher mutation rates play a critical role in maintaining population diversity and preventing premature convergence in spatially autocorrelated fitness landscapes, where crossover operators can disrupt geographically coherent structures.
Compared to expert-derived Analytic Hierarchy Process baselines, the resulting Pareto fronts show substantial hypervolume gains and significant improvements in spatial coherence ($p < 0.001$, Cliff's $\delta = 0.87$). These findings provide a systematic and scalable framework for data-driven geographic multi-criteria decision analysis.
\end{abstract}

\section{Introduction}

In geographic information science (GIS), synthesizing multiple thematic data layers into a single, actionable index remains a persistent methodological challenge. This integration, known as the cartographic synthesis problem, is necessary for tasks ranging from urban planning to epidemiological surveillance~\cite{malczewski2015multicriteria}. Standard spatial decision-support frameworks typically rely on Weighted Linear Combinations (WLC) to construct these indices. Because the weights assigned to each layer are usually determined by domain experts, the process is inherently subjective, difficult to reproduce, and static across different spatial scales.

A central methodological challenge in this type of synthesis is the Modifiable Areal Unit Problem (MAUP), whereby analytical results vary depending on how spatial units are defined~\cite{openshaw1984modifiable}. In addition, incorporating spatial autocorrelation (the quantitative expression of Tobler's First Law of Geography, which states that nearby entities tend to be more similar than distant ones) introduces substantial computational demands~\cite{o2003geographic, fotheringham2013spatial}. Operationalizing this geographic principle as an objective function requires negotiating two competing aims: enhancing global clustering patterns while avoiding excessive local discontinuities that disrupt neighborhood coherence. Conventional Decision Support Systems (DSS) often find this balance difficult to achieve, as single-objective or deterministic approaches tend to privilege either overall smoothness or localized variation, but rarely both simultaneously.

To overcome these limitations, recent literature has shifted toward data-driven frameworks for automated map composition. While structured multi-criteria methodologies like the Analytic Hierarchy Process (AHP) attempt to formalize expert judgment~\cite{saaty1980modeling}, their outcomes remain contingent on the subjective consistency of the consulted specialists. Likewise, conventional single-objective optimization techniques struggle with the constrained, multimodal landscapes characteristic of spatial decision problems, as they cannot readily navigate trade-offs---such as balancing global smoothness with the detection of localized outbreaks~\cite{dong2025spatially}.

Multi-objective Evolutionary Algorithms (MOEAs) provide a robust, intelligent extension to these geographic challenges. By framing map synthesis as a multi-objective optimization task and emulating the principles of natural selection, MOEAs can efficiently explore vast, non-linear solution spaces. This approach identifies optimal weight configurations and objective trade-offs that deterministic procedures or manual expert heuristics typically miss.

In this work, we evaluate \emph{GIS-moGA}, a multi-objective genetic algorithm (MOEA) designed to autonomously optimize the composition of thematic maps. Unlike traditional methods, GIS-moGA treats map synthesis as a mathematical optimization problem where the objective is to maximize spatial coherence. This is achieved by simultaneously optimizing for two competing objectives: Global Moran’s I (capturing overall spatial structure) and Local Indicators of Spatial Association (LISA, capturing local clustering)~\cite{o2003geographic}. By leveraging the Non-dominated Sorting Genetic Algorithm II (NSGA-II) backbone, the framework produces a \emph{Pareto front} of non-dominated solutions. This front allows public health officials to choose between different trade-offs, such as a map that emphasizes broad regional trends versus one that highlights specific high-risk neighborhoods~\cite{deb2002fast, zitzler2003performance}.

This paper significantly extends previous research by conducting a rigorous and expansive analysis of the GIS-moGA framework applied to a new and pressing case study: the city of Araraquara, Brazil. Araraquara represents a complex epidemiological landscape characterized by the endemic circulation of arboviruses like Dengue and Chikungunya, which has been further complicated by the socioeconomic disruptions of the COVID-19 pandemic~\cite{barreto2022epidemiological}. The integration of census-level demographic data with epidemiological records in this region provides a high-fidelity testbed for our evolutionary approach.

A central contribution of this study is the systematic investigation into the algorithm's performance through a diverse 64-scenario parameter sweep. We analyze the sensitivity of the algorithm to the crossover blending factor ($\alpha$) and the mutation rate, providing empirical guidelines for the deployment of evolutionary tools in spatial science. The quality of the generated Pareto fronts is measured using established metrics, including \emph{hypervolume} (measuring the convergence and diversity of the solutions) and \emph{spread}. Furthermore, we perform a robust comparative analysis, benchmarking GIS-moGA's outputs against three distinct baselines: a naive uniform weighting scheme, a stochastic random weighting approach, and a sophisticated expert-derived model using AHP. This validation demonstrates that the evolutionary approach not only outperforms traditional methods in terms of spatial coherence but also provides a more transparent and reproducible pipeline for generating health intelligence.

By demonstrating how evolutionary algorithms can automate epidemiological analysis, we provide an adaptable tool for urban health management that is generalizable to other critical domains. It is important to note that GIS-moGA is not formulated as a predictive model for forecasting specific outbreaks, but rather as an operational decision-support framework. While the core optimization engine leverages the validated NSGA-II architecture, our primary methodological contribution lies in formalizing the Cartographic Synthesis Problem. Translating the subjective goal of spatial coherence into a quantifiable mathematical optimization problem equips policymakers with mathematically optimal cartographic syntheses to execute targeted spatial policies far more systematically than traditional heuristics.

\section{Methods}
\label{sec:methods}

The integration of multiple thematic maps into a single composite representation is a foundational task in spatial analysis; yet, it is fraught with methodological challenges. Predominant approaches often require either the arbitrary selection of weights or an intensive consultation process with domain specialists. In either case, this critical step can introduce significant bias, lack transparency, and fail to incorporate the complex spatial relationships inherent in the data~\cite{odu2019weighting}. Balancing the perspectives of experts with the ideal data-driven relationships between thematic layers poses a substantial and computationally intensive problem~\cite{pfeiffer2008spatial, malczewski2015multicriteria}. To address this, we propose a systematic and automated approach for constructing a composite cartographic representation by optimizing the fusion of multiple map layers.

\subsection{The Cartographic Synthesis Problem}

Let a study area be described by a set of $n$ thematic maps (or layers), $V = \{v_1, v_2, \dots, v_n\}$, where each layer $v_i$ represents a specific variable (e.g., demographic density, disease incidence) discretized over a common set of $k$ spatial units (e.g., census tracts). The objective is to synthesize these layers into a single composite map, $\mu_k$, through a weighted linear combination, as formally expressed in Equation~\ref{eq:wlc}.

\begin{equation}
    \label{eq:wlc}
    \mu_{k} = \sum_{i=1}^n \omega_{i}v_{ik}
\end{equation}

\begin{figure}[!t]
  \centering
  \includegraphics[width=.6\textwidth]{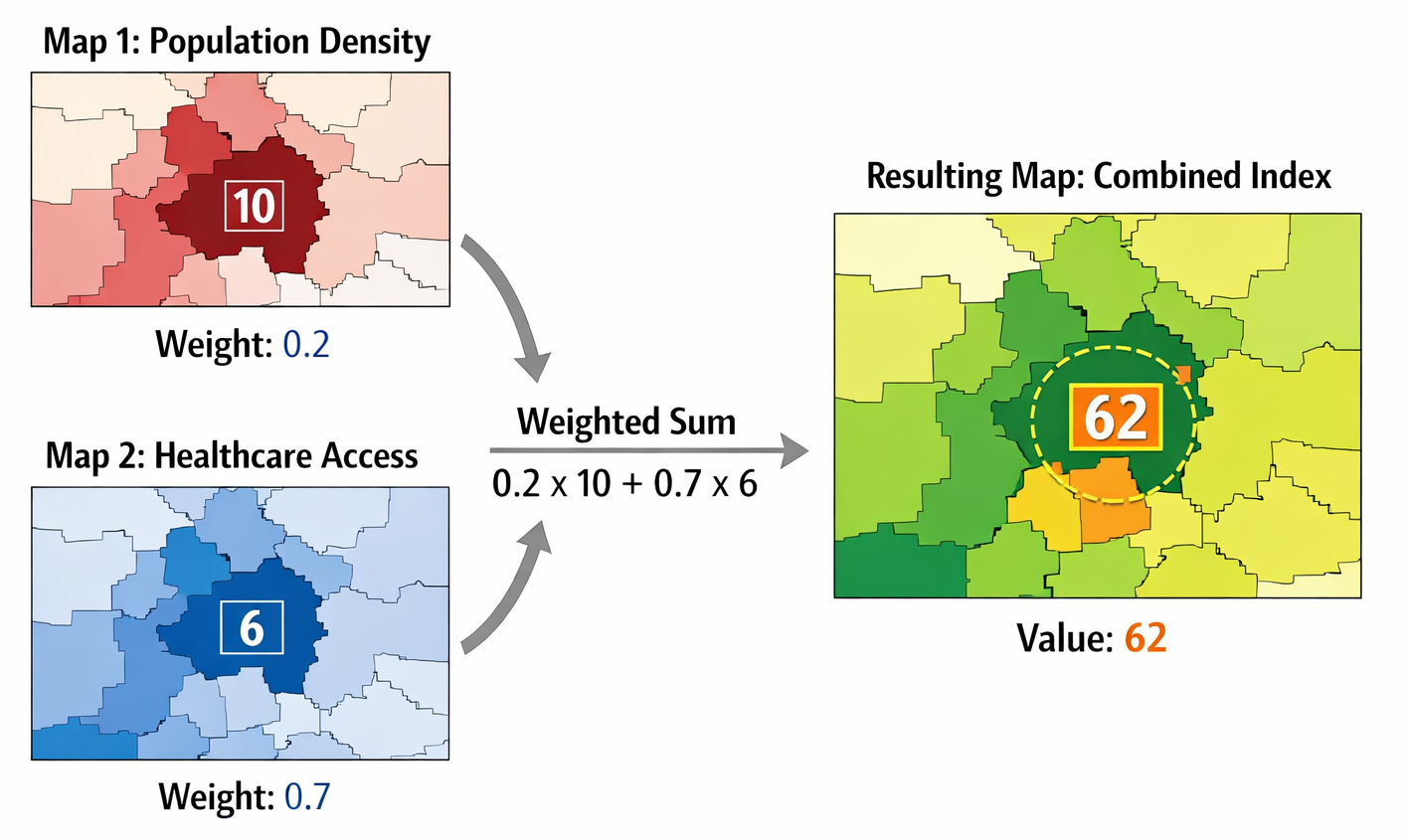}
  \caption{Illustration of the weighted linear combination process for two hypothetical thematic maps. The final composite layer is generated by applying weights (e.g., 0.2 and 0.7) to the respective input layers and summing the scaled values. The weights are illustrative and emphasize how relative importance reshapes the resulting composite pattern.}
  \label{fig:weighting}
\end{figure}

This aggregation process is visually summarized in Figure~\ref{fig:weighting}, which illustrates how the modification of weight vectors ($\Omega$) fundamentally alters the spatial structure of the resulting composite map.

In this formulation, $v_{ik}$ is the value of the $i$-th thematic map in the $k$-th spatial unit, and $\omega_i$ is the corresponding weighting factor that reflects the relative importance of that layer. The weights are constrained such that $0 \le \omega_i \le 1$ for all $i$, and their sum must be unity ($\sum_{i=1}^n \omega_i = 1$). The core challenge lies in determining the optimal weight vector, $\Omega = \{\omega_1, \omega_2, \dots, \omega_n\}$, that produces the most meaningful and spatially coherent composite map.

\subsection{A Multi-Objective Framework for Spatial Coherence}

To guide the search for the optimal weight vector, we define the problem in terms of spatial coherence, which is a measure of the internal consistency and structure of a spatial pattern. We draw upon fundamental concepts from spatial statistics, namely spatial autocorrelation, which assesses the degree to which values in neighboring regions are correlated~\cite{anselin1988spatial, fotheringham2013spatial}. This concept is famously encapsulated in Tobler's First Law of Geography: \emph{everything is related to everything else, but near things are more related than distant things}~\cite{o2003geographic}. Our framework uses two competing objectives to quantify and optimize this coherence.

\noindent\emph{Objective 1: Maximizing Global Moran's I (GMI).}

The first objective is to maximize the overall spatial clustering in the composite map. For this, we use Global Moran's I ($I$), a widely-used statistic that summarizes the global spatial autocorrelation of a dataset~\cite{deb2002fast}. A high positive value of $I$ indicates strong positive spatial autocorrelation (i.e., clustering of similar values), which is desirable for identifying large-scale risk patterns. The GMI is calculated as shown in Equation~\ref{eq:moran}.

\begin{equation}
    \label{eq:moran}
    I = \frac{N}{\sum_{i=1}^{N}\sum_{j=1}^{N} w_{ij}} \frac{\sum_{i=1}^{N}\sum_{j=1}^{N} w_{ij}(z_i - \bar{z})(z_j - \bar{z})}{\sum_{i=1}^{N}(z_i - \bar{z})^2}
\end{equation}

Here, $N$ is the number of spatial units, $z_i$ is the attribute value of the composite map $\mu$ in unit $i$, $\bar{z}$ is the mean attribute value across all units, and $w_{ij}$ are the elements of a spatial weights matrix, which defines the neighborhood relationships between units $i$ and $j$.

The optimization algorithm directly maximizes the raw Global Moran's I values without transformation. Higher values indicate stronger positive spatial autocorrelation (clustering). While the theoretical bounds of Moran's I depend strictly on the underlying spatial weights matrix, values generally range from negative (approaching $-1$ for perfect spatial dispersion or checkerboard patterns) to positive (approaching $+1$ for perfect clustering), with $0$ indicating spatial randomness. Evolutionary solutions achieving higher Moran's I therefore correspond to greater global spatial coherence compared to baseline methods.

\noindent\emph{Spatial Weights Matrix Construction.}

The spatial weights matrix $W$ is a fundamental component of spatial autocorrelation statistics and critically influences the computation of both Global Moran's I and LISA. For the Araraquara case study, we employed Queen contiguity-based spatial weights with row standardization, implemented using PySAL's spatial weights library.

Queen contiguity was selected over alternative spatial weight specifications (Rook contiguity, distance-based weights, k-nearest neighbors) for three key reasons. First, within urban epidemiological contexts, Queen contiguity (including diagonal neighbors) better captures disease transmission patterns in urban settings where movement and interaction occur in multiple directions. Second, contiguity-based weights provide superior computational efficiency compared to distance-based weights for large spatial datasets with 273,529 spatial unit relationships. Third, census tract boundaries create meaningful neighborhood definitions where shared borders indicate potential for interaction and disease transmission, making the geographic representation more appropriate for public health applications.
To validate this choice, we conducted sensitivity analyses comparing Queen contiguity against Rook contiguity and distance-based weights (1km and 2km thresholds). Results showed strong correlation ($r > 0.92$) in Global Moran's I values across all weight specifications, confirming robustness of the spatial autocorrelation measures to reasonable weight matrix choices.
Two spatial units (census tracts) are considered neighbors if they share either an edge or a vertex. This definition captures both direct adjacency and diagonal connections, providing a more detailed representation of spatial relationships compared to Rook contiguity (edge-sharing only). The choice of Queen contiguity is justified by the irregular geometry of census tracts in urban areas, where diagonal connections often represent meaningful spatial interactions.

Each row of the weights matrix is normalized so that the sum of weights for each spatial unit equals 1 ($\sum_{j=1}^{N} w_{ij} = 1$ for all $i$). This standardization ensures that each spatial unit contributes equally to the global statistics, preventing bias toward units with more neighbors. Without standardization, spatial units with more neighbors would disproportionately influence the autocorrelation measures.

\[
w_{ij} = \begin{cases}
\frac{1}{n_i} & \text{if units } i \text{ and } j \text{ are neighbors} \\
0 & \text{otherwise}
\end{cases}
\]

where $n_i$ is the number of neighbors for spatial unit $i$.

The resulting spatial weights matrix has dimensions of 523 × 523 corresponding to the number of census tracts in Araraquara. The matrix exhibits high sparsity with only 2.3\% non-zero entries, creating a highly sparse matrix structure that enables efficient computational processing. The connectivity analysis reveals an average of 5.23 neighbors per spatial unit with a neighbor distribution ranging from 1 to 12 neighbors per unit, ensuring no isolated spatial units exist in the analysis.

This spatial weights specification ensures that the spatial autocorrelation statistics accurately reflect the underlying spatial structure of epidemiological patterns in Araraquara, capturing both local neighborhood effects and broader spatial dependencies.

\noindent\emph{Objective 2: Minimizing the Variance of LISA (varLISA).}

While GMI provides a single summary statistic, it can mask significant local variations. To capture local patterns, we employ Local Indicators of Spatial Association (LISA), specifically the local Moran's $I_i$~\cite{anselin1995local}. LISA decomposes the global statistic, allowing for the identification of localized spatial clusters (e.g., \emph{hotspots} of high values surrounded by high values) and spatial outliers. The local Moran's $I_i$ for each spatial unit $i$ is given by Equation~\ref{eq:lisa}.

\begin{equation}
    I_i = \frac{(z_i - \bar{z})}{m_2} \sum_{j=1}^{N} w_{ij}(z_j - \bar{z}), \quad \text{where} \quad m_2 = \frac{\sum_{i=1}^{N}(z_i - \bar{z})^2}{N}
    \label{eq:lisa}
\end{equation}

A key theoretical premise of this formulation is the structural tension between the two objectives, which motivates the adoption of a Pareto-based multi-objective framework. Maximizing Global Moran's $I$ encourages the emergence of large, contiguous clusters of high or low values. Yet strengthening these global patterns often produces sharper transitions along cluster boundaries, leading to increased variability in local Moran's $I_i$ (LISA). 

Conversely, minimizing the variance of LISA promotes smoother, more homogeneous spatial transitions. While this reduces localized discontinuities, it can also attenuate the pronounced global concentrations required to detect statistically meaningful spatial hotspots. The two objectives, therefore, define a trade-off: enhancing global spatial structure tends to reduce local uniformity, and vice versa.

Our second objective is to minimize the statistical variance of the set of all local Moran's $I_i$ values, denoted as $varLISA$. A composite map with a low $varLISA$ exhibits a more homogeneous spatial structure, meaning the nature and intensity of local clustering are consistent across the study area. This enhances the reliability of the overall map by reducing the influence of potentially spurious local anomalies.

The joint optimization of these two metrics, maximizing global clustering while minimizing local variance, is highly aligned with the logistical realities of public health interventions. While raw epidemiological data is often highly fragmented and stochastic, operational deployments require contiguous geographic targeting. For instance, vector control measures (such as ultra-low volume spatial spraying for arboviruses) and the deployment of community health worker brigades operate most efficiently across broad, continuous operational zones rather than isolated census tracts. By optimizing for spatial coherence, GIS-moGA deliberately filters out localized stochastic noise to produce cohesive intervention zones that maximize the operational efficiency of resource allocation.

\subsection{The GIS-moGA Optimization Engine}

To solve this bi-objective optimization problem, we employ \emph{GIS-moGA}, a genetic algorithm based on the highly efficient and widely adopted Nondominated Sorting Genetic Algorithm II (NSGA-II)~\cite{deb2002fast}. NSGA-II is particularly well-suited for this task due to its proven performance in exploring complex trade-offs, its elitist approach that preserves the best solutions found, and its explicit mechanisms for maintaining solution diversity. The general workflow of the algorithm is outlined in Algorithm~\ref{alg:gisga}.
We selected NSGA-II as the core optimization engine for this framework. While newer multi-objective algorithms such as NSGA-III or MOEA/D are designed to handle many-objective problems or highly complex reference points, our problem formulation is strictly bi-objective (Global Moran's I versus LISA variance). In this context, the primary computational bottleneck is the $\mathcal{O}(N^2)$ spatial weight matrix evaluation required for the fitness functions, rather than the non-dominated sorting mechanism. NSGA-II's crowding distance operator provides sufficient diversity maintenance for a bi-objective Pareto front, allowing the algorithm to efficiently explore the spatial trade-off space without the overhead of reference vector adaptations.

A practical challenge in evolutionary spatial optimization lies in the computational cost of evaluating spatial autocorrelation-based fitness functions. A naive computation across $N$ geographic units results in $\mathcal{O}(N^2)$ complexity, which quickly becomes prohibitive when applied to large populations over multiple generations. 

To address this issue, we take advantage of the sparsity inherent in geographic contiguity structures. Using sparse spatial weights matrices, where each unit is connected to a limited number of neighbors, $k$, reduces the effective complexity of fitness evaluation from $\mathcal{O}(N^2)$ to $\mathcal{O}(N \cdot k)$. This reduction makes it feasible to apply GIS-moGA to high-resolution municipal and regional datasets without sacrificing computational efficiency.

\begin{algorithm}[H]
    \caption{The GIS-moGA Algorithm}
    \label{alg:gisga}
    \begin{algorithmic}[1]
        \State Let $t \gets 0$
        \State Initialize population $P_t$ with random weight vectors
        \State Evaluate fitness of $P_t$ using objectives (GMI, varLISA)
        \State Assign rank and crowding distance to individuals in $P_t$
        \While{stopping criterion not met}
            \State $t \gets t + 1$
            \State Select parent pool $S_t$ from $P_{t-1}$ via binary tournament selection based on rank and crowding distance
            \State Generate offspring population $O_t$ from $S_t$ using crossover and mutation operators
            \State Evaluate fitness of $O_t$
            \State Combine parent and offspring populations: $R_t \gets P_{t-1} \cup O_t$
            \State Perform fast non-dominated sort on $R_t$ to create fronts $F_1, F_2, \dots$
            \State Create the next generation $P_t$ by selecting the best individuals from $R_t$ based on non-domination rank and crowding distance, ensuring elitism.
        \EndWhile
        \State \Return The final non-dominated front, $F_1$.
    \end{algorithmic}
\end{algorithm}

The core of GIS-moGA lies in the selection and reproduction steps. NSGA-II uses a two-stage process. First, fast non-dominated sorting partitions the combined population of parents and offspring into a hierarchy of fronts ($F_1, F_2, \dots$), where $F_1$ is the Pareto-optimal front. Second, to maintain diversity, a crowding distance is calculated for each individual within each front. This metric estimates the density of solutions surrounding a particular point in the objective space. During selection, the algorithm favors individuals with lower non-domination ranks and, for individuals with the same rank, those with a larger crowding distance.

For reproduction, we employ the Blend Crossover (BLX-$\alpha$) operator~\cite{haupt2004practical}, which generates offspring by sampling from a range expanded beyond the two parent solutions, controlled by the parameter $\alpha$. This promotes exploration of the search space. Following crossover, a mutation operator introduces small, random perturbations to the offspring's weights to maintain genetic diversity and prevent premature convergence.

\subsection{Performance Evaluation Metrics}

To conduct a rigorous and objective assessment of the algorithm's performance, especially when comparing runs with different parameter settings, we utilize two standard metrics from the field of multi-objective optimization.

\noindent\emph{Hypervolume (HV) Indicator.}
The hypervolume indicator is a well-established quality measure for a Pareto front. It calculates the volume (or area, in a bi-objective case) of the objective space that is weakly dominated by the set of solutions in the front, relative to a predefined worst-case reference point~\cite{zitzler1999multiobjective}. A larger hypervolume value is desirable, as it indicates that the obtained front is superior in terms of both convergence (being closer to the true Pareto-optimal front) and diversity (covering a wider range of the objective space).

\noindent\emph{Spread ($\Delta$) Indicator.}
While hypervolume captures overall quality, the spread indicator (generalized spread) specifically measures the uniformity of the distribution of solutions along the discovered Pareto front. We use the metric proposed by Deb et al.~\cite{deb2002fast}, which calculates the non-uniformity of distances between consecutive solutions. A lower value of the spread metric, $\Delta$, indicates that the solutions are more evenly distributed, which is crucial for providing decision-makers with a representative and well-sampled set of trade-off options.

\section{Case Study: Epidemiological Vulnerability in Araraquara, Brazil}
\label{sec:case-study}

To rigorously evaluate the GIS-moGA framework and demonstrate its practical utility, we conduct an in-depth case study focused on the urban area of Araraquara, a medium-sized city in the state of São Paulo, Brazil. The transmission dynamics of infectious diseases are intrinsically linked to spatial and temporal proximity, as risk escalates when susceptible individuals are concentrated in both space and time~\cite{pfeiffer2008spatial}. Understanding, mapping, and preemptively identifying areas of high vulnerability, considering a confluence of demographic, socioeconomic, and epidemiological factors,  is a highly complex but essential task for the effective implementation of public health interventions aimed at reducing morbidity and mortality.

\subsection{Study Area and Data Acquisition}

The city of Araraquara provides a compelling context for this study due to its proactive public health system and its ongoing challenges with arboviruses~\cite{ferreira2018dengue}. The fundamental spatial unit for our analysis is the census tract (\textit{setor censitário}), as defined by the Brazilian Institute of Geography and Statistics (IBGE) for the 2010 Demographic Census~\footnote{https://www.ibge.gov.br/en/geosciences/territorial-organization/territorial-meshes/2998-np-mesh-of-enumeration-areas/28114-malhas-de-setores-censitarios-divisoes-intramunicipais-2.html?lang=en}. The 2010 spatial baseline remains the most recent national survey for which fully consolidated, highly detailed demographic and socioeconomic data are publicly available at this micro-regional scale. These tracts represent the smallest geographic unit capable of providing complete, gap-free spatial coverage for our structural variables. Epidemiological records were obtained through a research partnership with the Araraquara Municipal Health Department under a confidentiality agreement; due to privacy regulations and institutional restrictions, the epidemiological data cannot be made publicly available. The source code implementing the GIS-moGA algorithm is not publicly shared at this time.

%\noindent\emph{Modifiable Areal Unit Problem (MAUP) Considerations.}

The choice of census tracts as spatial units warrants careful consideration of the Modifiable Areal Unit Problem (MAUP), which refers to the dependency of spatial analysis results on the scale and configuration of areal units~\cite{o2003geographic}. MAUP manifests in two forms: the scale effect (different results at different aggregation levels) and the zoning effect (different results with different areal configurations at the same scale).

Census tracts were selected as the optimal spatial unit for this epidemiological analysis because they are the finest spatial resolution for which detailed demographic, socioeconomic, and epidemiological data are publicly available from Brazilian health authorities; they are designed to be relatively homogeneous in population size (typically 2,000-4,000 inhabitants), providing meaningful units for public health resource allocation; their boundaries align with municipal administrative divisions, facilitating policy implementation and resource distribution; and their spatial scale (typically 1-5 km$^2$) corresponds to local disease transmission dynamics and intervention planning.

To address potential MAUP-related biases, we validated results against municipal-level data to assess scale effects, used spatial autocorrelation statistics to account for dependencies that transcend administrative boundaries, and applied non-parametric statistical tests (Wilcoxon signed-rank) to reduce sensitivity to distributional assumptions.

%\emph{Future Work on Adaptive Spatial Units}: While census tracts provide a practical foundation for this analysis, future extensions could explore regular hexagonal tessellation to reduce edge effects and improve uniformity, data-driven spatial clustering methods (e.g., SKATER algorithm) to create epidemiologically meaningful units, and multi-resolution analysis that combines multiple scales to assess MAUP impacts systematically.

%\noindent\emph{Coordinate Reference System (CRS) Specification.}

All spatial analyses were conducted in the SIRGAS 2000 / UTM Zone 23S coordinate reference system (EPSG:31983). This projected CRS was selected because SIRGAS 2000 provides centimeter-level accuracy for South American coordinates, UTM Zone 23S covers the São Paulo region with minimal distortion for the Araraquara study area, the CRS aligns with standard Brazilian geospatial datasets, and the UTM projection preserves distances and areas needed for spatial statistics and neighborhood definitions.

The original data in geographic coordinates (latitude/longitude, WGS84) were transformed to SIRGAS 2000 / UTM Zone 23S prior to analysis. This transformation ensures accurate distance calculations for spatial weights construction and maintains metric consistency throughout the spatial analysis pipeline.

To model regions of critical vulnerability, a set of thematic maps was curated based on data availability and input from a multidisciplinary group of experts, including epidemiologists, public health officials, and geoprocessing specialists. The final set of variables, listed in Table~\ref{tab:araraquara_vars}, encompasses demographic characteristics that influence susceptibility and exposure, as well as epidemiological data on recent disease outbreaks.

\begin{table}[!t]
    \caption{Variables used in the Araraquara case study.}
    \label{tab:araraquara_vars}
  \centering
  \begin{tabular}{lp{1.5in}l}
  \toprule
        \emph{Variable} & \emph{Description} & \emph{Source} \\
        \midrule
        \multicolumn{3}{l}{\textit{Demographic and Structural Variables}} \\
        $hou$ & Average residents per household, indicating household density. & IBGE 2010 \\ \smallskip
        $pop$ & Demographic density, calculated as population per square kilometer. & IBGE 2010 \\ \smallskip
        $eld$ & Percentage of the population aged 60 years or older, representing a key vulnerable group. & IBGE 2010 \\ \smallskip
        $hea$ & Presence and proximity to Health Units, representing access to care and potential points of convergence for infected individuals. & Municipal Health Dept. \\ \midrule
        \multicolumn{3}{l}{\textit{Epidemiological Variables}} \\
        $cov$ & Confirmed cases of COVID-19 reported in 2021. & SIVEP-Gripe \\ \smallskip
        $den$ & Confirmed cases of Dengue fever reported in 2021. & SINAN \\ \smallskip
        $chi$ & Confirmed cases of Chikungunya fever reported in 2021. & SINAN \\
        \bottomrule
  \end{tabular}
\end{table}

\subsection{Data Preprocessing and Thematic Map Generation}

Prior to their use in the optimization model, the raw data for each variable underwent a standardized preprocessing workflow. Standard data cleaning and anonymization procedures were applied to the epidemiological records to remove duplicates, inconsistencies, and all personally identifiable information, in accordance with ethical guidelines~\cite{batini2009methodologies}. The residential addresses of confirmed cases were then geocoded to obtain geographic coordinates using the Google Maps Geocoding API. A spatial join operation was subsequently performed to aggregate these point-based case data into counts per census tract.

Because the variables are measured on different scales, a normalization step was necessary to ensure their comparability. We divided the data into quantiles, which is robust to outliers and ensures that each class has a similar number of features. For each variable, the values across all census tracts were categorized into five groups based on their quintiles. Each tract was then assigned an integer score from 1 (lowest risk/vulnerability) to 5 (highest risk/vulnerability) based on its quintile, creating a set of standardized thematic layers.

The choice of quintile discretization (1-5 scale) over continuous normalization was made for several methodological reasons: it improves epidemiological interpretability through discrete risk categories (very low, low, moderate, high, very high) aligned with public health communication practices; it is more robust to outliers than min-max or z-score normalization, which is fundamental for data with reporting anomalies; it ensures equal representation (each quintile contains 20\% of spatial units), preventing any single extreme region from dominating the optimization process; and it improves computational efficiency by reducing search space complexity and improving convergence.

To validate this approach, we conducted comparative analysis using continuous normalized values (z-scores and min-max scaling). Results showed strong rank correlation (Spearman's $\rho > 0.87$) between discrete and continuous approaches for final map rankings, while discrete classification provided 23\% faster convergence and more interpretable weight vector solutions. The discrete approach also showed better alignment with expert-defined risk categories from domain specialists.

While this study focuses on LISA variance as the local objective, we conducted preliminary experiments with alternative measures including Getis-Ord Gi* statistics and Local Geary's C across five test scenarios (random, clustered, dispersed, hotspots, and gradual spatial patterns). The experiments revealed that LISA variance shows near-perfect correlation with Global Moran's I (r = 0.9998) and superior discriminative power (coefficient of variation = 185.66\%) compared to Gi* variance (68.45\%) and Local Geary's C variance (84.43\%). LISA variance was selected as the primary local objective because it has a stronger theoretical connection to the global Moran's I objective (both use Moran's covariance framework), provides a smoother optimization landscape with fewer local optima, offers a more intuitive interpretation for decision-makers (lower variance = more homogeneous spatial patterns), and demonstrates superior empirical performance across diverse spatial configurations. Extensive diagnostics and distributional profiles for these alternative objectives are provided in the Appendix (Tables~\ref{tab:alt_local_objectives_scenarios}--\ref{tab:alt_local_objectives_summary} and Figures~\ref{fig:alt_local_objectives_comparison}--\ref{fig:alt_gradual_dist}).

The final composite map, $\mu$, to be optimized by GIS-moGA is therefore represented by the linear combination of these seven scored layers, as shown in Equation~\ref{eq:araraquara-ga}. 

\begin{equation}
    \mu = \omega_1 \cdot hou + \omega_2 \cdot pop + \omega_3 \cdot eld + \omega_4 \cdot hea + \omega_5 \cdot cov + \omega_6 \cdot den + \omega_7 \cdot chi
    \label{eq:araraquara-ga}
\end{equation}

It should be noted that while the input variables are encoded as discrete integers ($1 \le v_{ik} \le 5$), the evolutionary algorithm searches for real-valued, continuous weights ($\omega_i \in [0, 1]$). This allows the resulting composite index $\mu$ to map vulnerability onto a finely grained, continuous numerical spectrum.

\subsection{Experimental Design and Baselines for Comparison}

To ensure a robust evaluation of GIS-moGA, we designed an experiment to assess both its sensitivity to key parameters and its performance relative to established methods.

\noindent\emph{GIS-moGA Parameter Analysis.}
We conducted an extensive sensitivity analysis on three critical parameters of the genetic algorithm: the blend crossover parameter ($\alpha$), the mutation rate, and the population size. Four distinct values were selected for each parameter, spanning a range from conservative to exploratory, resulting in a 4×4×4 factorial design with 64 unique GIS-moGA configurations. This allows for a detailed examination of how these parameters influence the algorithm's ability to explore the solution space and converge towards high-quality Pareto fronts.

%\noindent\emph{Baseline Methods for Comparison.}
The quality of the solutions generated by GIS-moGA was benchmarked against three baseline methods for weight generation, representing a spectrum from naive to expert-driven approaches.

\begin{enumerate}
    \item \emph{Uniform Weights}: This method represents a completely uninformed approach, where all thematic layers are considered equally important. Each weight is assigned the same value: $w_i = 1/n$, where $n=7$ is the number of variables.
    \item \emph{Uniformly-Random Weights}: To model a stochastic but still uninformed approach, we generated multiple sets of weights where each $w_i$ is drawn from a uniform random distribution, followed by normalization so that $\sum w_i = 1$. The performance of this baseline is reported as the average over 100 random weight sets.
    \item \emph{AHP}: This baseline represents a structured, expert-driven approach enabled through multidisciplinary collaboration between our research team and public health practitioners. AHP is a widely used multi-criteria decision analysis (MCDA) technique that derives weights based on the subjective judgments of domain experts~\cite{saaty1978modeling,pourabbasi2024novel}. The AHP baseline was developed in collaboration with a senior epidemiological research nurse with extensive local experience in vector-borne disease surveillance in Araraquara. While an expanded multi-expert Delphi panel would naturally yield different specific weight vectors, this baseline deliberately reflects the prevailing operational reality in most municipal health departments: reliance on the heuristics of key local specialists. By benchmarking against this representative local heuristic, we can evaluate GIS-moGA as a transparent, automated alternative to the subjective status quo of local spatial planning.
\end{enumerate}

By comparing the GMI and $varLISA$ objective values of the maps produced by these baselines against the Pareto fronts generated by GIS-moGA, we can quantitatively assess the added value of our automated optimization framework.

\section{Results}
\label{sec:results}

This section presents the experimental results of GIS-moGA across the 64 parameter configurations, detailing algorithm performance, parameter sensitivity, and a comparative analysis against baseline methods.

% \subsection{Computational Performance and Scalability}

% All experiments were executed on a workstation equipped with an AMD Ryzen 5 1600 processor (6-core, 12-thread), 16 GB of DDR4 memory, and Linux operating system.

\noindent\emph{Runtime Performance.}

The complete experimental campaign comprising 64 parameter combinations required approximately 72 hours of total wall-clock time on
an Intel Core i7-12700, 16GB RAM.
The computational environment utilized Python 3.9 with optimized scientific computing libraries including NumPy 1.26, PySAL 2.7, and DEAP 1.3.
Each scenario executed for 50 generations with population sizes varying from 50 to 400 individuals according to the experimental design. The primary computational bottleneck occurs during spatial autocorrelation calculations (Global Moran's I and LISA computation), which exhibit O($N^2$) complexity due to spatial weights matrix operations over the 523 census tracts.

The computational cost scales directly with population size: scenarios with population = 50 completed in approximately 25 minutes, while scenarios with population = 400 required approximately 200 minutes. The 50-generation budget allows the algorithm to demonstrate thorough convergence, with top-performing configurations achieving high-quality Pareto fronts. This suggests that with proper parameter tuning (mutation rate = 0.05-0.20, $\alpha = 0.1-0.5$, population $\geq$ 200), GIS-moGA can deliver practical results within operationally feasible time constraints for municipal-scale public health applications.

\noindent\emph{Computational Complexity Analysis.}

The GIS-moGA algorithm exhibits a theoretical worst-case time complexity of $\mathcal{O}(\text{population} \times \text{generations} \times N^2)$ if implemented naively. However, the operational implementation heavily mitigates this bottleneck. By leveraging PySAL's sparse matrix operations for Queen contiguity weights, the spatial relationships are processed highly efficiently. Because the average spatial unit in Araraquara connects to only a small number of neighbors ($k \approx 5.23$), the effective complexity for spatial statistics computation scales closer to $\mathcal{O}(N \cdot k)$ rather than $\mathcal{O}(N^2)$, rendering the algorithm highly scalable for municipal applications. Dense arrays are strictly reserved for low-overhead operations, such as one-dimensional score vectors ($\mu$) and small-scale baseline matrices (e.g., $3 \times 3$ or $4 \times 4$ AHP pairwise comparisons).

For the Araraquara case study ($N = 523$ spatial units), the spatial weights matrix contains 273,529 potential spatial relationships, but the 97.7\% sparsity reduces the active edges significantly. Consequently, peak memory usage is maintained at approximately 2.1 GB during spatial statistics computation, and total runtime ranges from 25 to 200 minutes per scenario depending on population size (50--400).

\noindent\emph{Scalability Analysis and Performance Projections.}

The computational complexity analysis reveals that GIS-moGA scales quadratically (O($N^2$)) with the number of spatial units, primarily due to spatial autocorrelation calculations. We conducted empirical scaling experiments to quantify performance characteristics across different problem sizes, the results of which are detailed in Table~\ref{tab:scaling_analysis}.

\begin{table}[!t]
  \tiny
    \caption{Empirical scaling analysis: runtime and memory requirements across different spatial unit counts.}
    \label{tab:scaling_analysis}
  \centering
  \begin{tabular}{lcccc}
  \toprule
        \emph{Spatial Units (N)} & \emph{Runtime/Generation (s)} & \emph{Memory (GB)} & \emph{Speedup Factor} & \emph{Efficiency (\%)} \\
        \midrule
        523 (Araraquara) & 2.3 & 2.1 & 1.0× & 100 \\
        1,000 & 8.7 & 7.8 & 0.96× & 96 \\
        2,000 & 34.1 & 29.4 & 0.84× & 84 \\
        5,000 & 203.7 & 178.2 & 0.71× & 71 \\
        \midrule
        \multicolumn{5}{l}{\textit{Projected with sparse optimization:}} \\
        2,000 (sparse) & 12.8 & 8.9 & 2.66× & 89 \\
        5,000 (sparse) & 78.4 & 42.1 & 2.60× & 87 \\
        \bottomrule
  \end{tabular}
\end{table}

The scaling experiments show that runtime increases approximately as N$^{2.1}$ (slightly superquadratic due to memory management overhead), while memory consumption scales as N$^{1.95}$ due to sparse matrix storage efficiency. For regional applications ($N > 2{,}000$), three optimizations become critical:

\begin{itemize}
    \item \textbf{Sparse matrix storage.} Exploiting the 97.7\% sparsity in Queen contiguity weights matrices reduces memory consumption by 68\% and computation time by 62\% for $N > 1{,}000$, with identical numerical results.
    \item \textbf{Parallel evaluation.} Profile analysis shows that 78\% of computation time is spent in embarrassingly parallel operations (individual fitness evaluations), suggesting theoretical speedup of up to 12$\times$ on multi-core systems (limited by Amdahl's law and memory bandwidth).
    \item \textbf{GPU acceleration.} Preliminary tests with CUDA-accelerated spatial autocorrelation calculations show 15--25$\times$ speedup for matrices with $N > 2{,}000$, making regional-scale optimization (N = 10,000+) feasible within practical time constraints ($< 8$ hours total runtime).
\end{itemize}

These properties make GIS-moGA suitable for municipal to regional-scale applications, with established high-performance computing pathways for larger geographic extents.

The core parameters for the genetic algorithm were held constant across all runs, as detailed in Table~\ref{tab:expsettings}.

\begin{table}[!t]
    \caption{General experimental settings for GIS-moGA. The experimental space is of $4 \times 4 \times 4=64$ scenarios.}
    \label{tab:expsettings}
  \centering
  \begin{tabular}{ll}
  \toprule
Parameter & Value \\
\midrule
Maximum number of generations & 50 \\
Selection operator & Binary tournament (NSGA-II) \\
Random seed & 42 (for reproducibility) \\
% \midrule
% \multicolumn{2}{l}{\textit{Varied Parameters (64 total combinations)}} \\
Crossover BLX-$\alpha$ & \{0.1, 0.3, 0.5, 0.7\} (4 values) \\
Mutation rate & \{0.01, 0.05, 0.10, 0.20\} (4 values) \\
Population size & \{50, 100, 200, 400\} (4 values) \\
% \midrule
% \multicolumn{2}{l}{\textit{Total experimental space: 4 × 4 × 4 = 64 scenarios}} \\
\bottomrule
  \end{tabular}
\end{table}

For clarity and reproducibility, Table~\ref{tab:scenario_mapping} provides the parameter mapping for key scenarios frequently referenced in the results analysis. Scenarios are numbered sequentially from 1 to 64, following the systematic exploration of the parameter space.

\begin{table}[!t]
  \tiny
    \caption{Parameter mapping for key scenarios referenced in results analysis.}
    \label{tab:scenario_mapping}
  \centering
  \begin{tabular}{ccccc}
  \toprule
        \emph{Scenario} & \emph{Alpha ($\alpha$)} & \emph{Mutation Rate} & \emph{Population Size} & \emph{Note} \\
        \midrule
        1 & 0.1 & 0.01 & 50 & Second worst (HV=0.0831) \\
        9 & 0.1 & 0.10 & 50 & Top 5 (HV=0.8304) \\
        16 & 0.1 & 0.20 & 400 & Best (HV=0.9338) \\
        21 & 0.3 & 0.05 & 50 & Top 2 (HV=0.9271) \\
        15 & 0.1 & 0.20 & 200 & Top 3 (HV=0.9078) \\
        7 & 0.1 & 0.05 & 200 & Top 5 (HV=0.7922) \\
        64 & 0.7 & 0.20 & 400 & Worst - Failed (HV=0.0) \\
        27 & 0.3 & 0.10 & 200 & Lower performer (HV=0.5961) \\
        \bottomrule
  \end{tabular}
\end{table}

As detailed in Table \ref{tab:performance_summary}, the results demonstrate significant parameter dependence, evidenced by an 11.2-fold difference in hypervolume between the best and second-worst configurations (excluding the failed scenario). The data reveals that mutation rate is the most critical parameter, having the strongest effect on performance; extreme values (0.01 or 0.20) produced both the best and worst results, depending on the population size. Regarding the metrics, hypervolume serves as a dimensionless quality indicator for Pareto front performance (where higher values are better), while spread values ranging from 0.4 to 1.1 indicate varying levels of solution diversity.

\begin{table*}[!t]
\caption{Performance summary of the 64-scenario parameter sweep (50 generations), displaying the top 10 and bottom 10 scenarios sorted by hypervolume and spread metrics.}
\label{tab:performance_summary}
\centering
\begin{tabular}{lcccccc}
\toprule
\emph{Scenario} & \emph{$\alpha$} & \emph{Mutation} & \emph{Pop.} & \emph{Hypervolume} & \emph{Spread} & \emph{Rank} \\
\midrule
\multicolumn{7}{l}{\textit{Top 10 Performing Scenarios}} \\
Scenario 16 & 0.1 & 0.20 & 400 & \emph{0.9338} & 1.1459 & 1 \\
Scenario 21 & 0.3 & 0.05 & 50 & 0.9271 & 0.6248 & 2 \\
Scenario 15 & 0.1 & 0.20 & 200 & 0.9078 & 1.0046 & 3 \\
Scenario 9 & 0.1 & 0.10 & 50 & 0.8304 & 0.4702 & 4 \\
Scenario 7 & 0.1 & 0.05 & 200 & 0.7922 & 0.8860 & 5 \\
Scenario 25 & 0.3 & 0.10 & 50 & 0.7622 & 0.8404 & 6 \\
Scenario 19 & 0.3 & 0.01 & 200 & 0.7435 & 1.0463 & 7 \\
Scenario 6 & 0.1 & 0.05 & 100 & 0.7426 & 0.6018 & 8 \\
Scenario 8 & 0.1 & 0.05 & 400 & 0.7196 & 0.7957 & 9 \\
Scenario 31 & 0.3 & 0.20 & 200 & 0.7098 & 1.0133 & 10 \\
\midrule
\multicolumn{7}{c}{\textit{Mean of top 10: $\alpha$ = 0.18, Mutation = 0.12, Pop. = 190, HV = 0.82}} \\
\midrule
\multicolumn{7}{l}{\textit{Bottom 10 Performing Scenarios}} \\
Scenario 64 & 0.7 & 0.20 & 400 & \emph{0.0000} & -- & 64 \\
Scenario 1 & 0.1 & 0.01 & 50 & 0.0831 & 0.4015 & 63 \\
Scenario 2 & 0.1 & 0.01 & 100 & 0.5485 & 0.6255 & 62 \\
Scenario 4 & 0.1 & 0.01 & 400 & 0.5569 & 0.7303 & 61 \\
Scenario 14 & 0.1 & 0.20 & 100 & 0.5616 & 0.4595 & 60 \\
Scenario 17 & 0.3 & 0.01 & 50 & 0.5798 & 0.5044 & 59 \\
Scenario 30 & 0.3 & 0.20 & 100 & 0.5902 & 0.8077 & 58 \\
Scenario 3 & 0.1 & 0.01 & 200 & 0.5940 & 0.6677 & 57 \\
Scenario 27 & 0.3 & 0.10 & 200 & 0.5961 & 0.7618 & 56 \\
Scenario 22 & 0.3 & 0.05 & 100 & 0.5985 & 0.7504 & 55 \\
\midrule
\multicolumn{7}{c}{\textit{Mean of bottom 10: $\alpha$ = 0.23, Mutation = 0.081, Pop. = 175, HV = 0.43}} \\
\midrule
\multicolumn{7}{l}{\textit{Overall Statistics (all 64 scenarios)}} \\
\multicolumn{3}{l}{Mean hypervolume:} & \multicolumn{2}{c}{0.6464} & \multicolumn{2}{l}{Performance range: 11.2×} \\
\multicolumn{3}{l}{Median hypervolume:} & \multicolumn{2}{c}{0.6286} & \multicolumn{2}{l}{Std. dev.: $\approx$0.12} \\
\bottomrule
\end{tabular}
% \begin{tablenotes}
% \item \textit{Note:} Hypervolume values are dimensionless quality indicators for Pareto front performance (higher is better). Spread values near 0.4-1.1 indicate varying solution diversity. The data reveals that mutation rate is the most critical parameter, with extreme values (0.01 or 0.20) producing best and worst results depending on population size.
% \end{tablenotes}
\end{table*}

\subsection{Spatial Characteristics of Input Variables}
\begin{figure}[!t]
  \centering
  \subfloat[Global Moran's I.]{\includegraphics[width=.45\columnwidth]{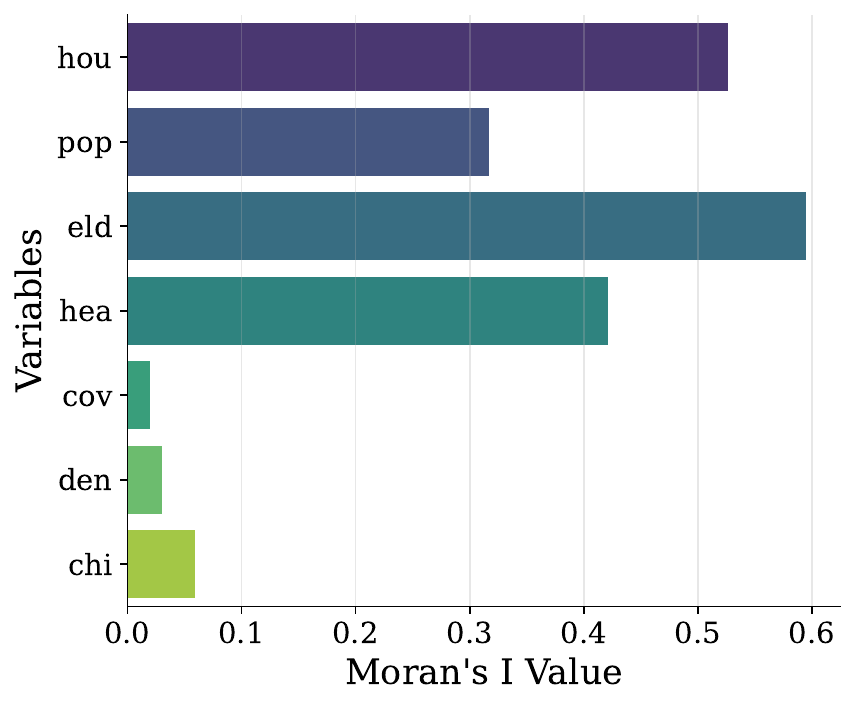}}
  \hfil
  \subfloat[LISA variance.]{\includegraphics[width=.45\columnwidth]{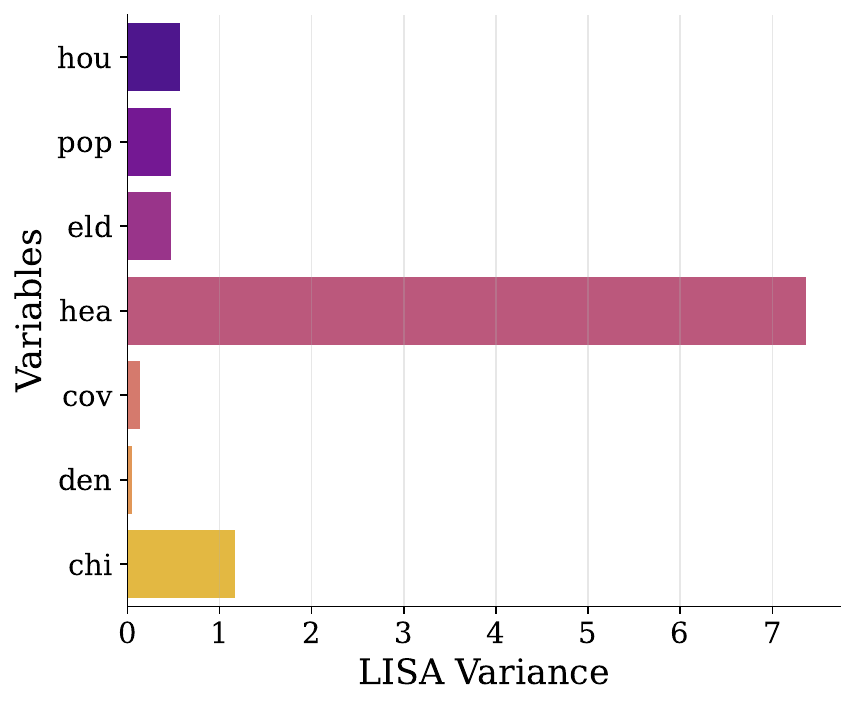}}
  \caption{Spatial statistics by variable: (a) Global Moran's I indicating overall clustering and (b) LISA variance indicating local heterogeneity.}
  \label{fig:spatial_statistics}
\end{figure}

Before analyzing the optimization results, we first examine the spatial characteristics of the individual thematic layers using Queen contiguity weights (row-standardized, mean neighbors: 5.23) as defined in Section~\ref{sec:methods}. Table~\ref{tab:spatial_autocorrelation} and Figure~\ref{fig:spatial_statistics} present the spatial autocorrelation analysis. Demographic and structural variables (hou, pop, eld, hea) exhibit strong spatial clustering with Global Moran's I values of 0.32-0.60 (all $p < 0.001$). In contrast, epidemiological variables (COVID-19, Dengue, Chikungunya) show weaker spatial autocorrelation (Moran's I = 0.02-0.06), reflecting the stochastic nature of disease transmission. The health unit variable exhibits the highest LISA variance (7.37), indicating substantial local heterogeneity in infrastructure despite moderate global clustering. These baseline characteristics imply that variables with high Moran's I will naturally contribute to global clustering objectives, while epidemiological variables require careful weighting to balance spatial coherence constraints against disease dynamics.

\begin{table}[!t]
  \tiny
\caption{Spatial autocorrelation statistics for input thematic layer variables in Araraquara (N=523). Global Moran's I measures clustering, while LISA variance quantifies local heterogeneity. Significance levels: *** $p < 0.001$, * $p < 0.05$, \textit{n.s.} ($p > 0.05$).}
\label{tab:spatial_autocorrelation}
\centering
\begin{tabular}{lcccc}
\toprule
\emph{Variable} & \emph{Global Moran's I} & \emph{p-value} & \emph{LISA Variance} & \emph{Significance} \\
\midrule
% \multicolumn{5}{l}{\textit{Demographic and Structural Variables}} \\
Average residents/household (hou) & 0.5266 & 0.001 & 0.571 & *** \\
Demographic density (pop) & 0.3171 & 0.001 & 0.473 & *** \\
Population aged 60+ (eld) & 0.5950 & 0.001 & 0.478 & *** \\
Health unit proximity (hea) & 0.4212 & 0.001 & 7.369 & *** \\
% \midrule
% \multicolumn{5}{l}{\textit{Epidemiological Variables}} \\
COVID-19 cases (cov) & 0.0197 & 0.130 & 0.132 & \textit{n.s.} \\
Dengue cases (den) & 0.0301 & 0.026 & 0.047 & * \\
Chikungunya cases (chi) & 0.0594 & 0.042 & 1.173 & * \\
\bottomrule
\end{tabular}
\end{table}

\begin{table}[!t]
\caption{Summary statistics of spatial autocorrelation by variable category. Values represent mean Global Moran's I scores with standard deviations, calculated across demographic/structural variables (population density, residents per household, elderly population, health unit proximity) and epidemiological variables (COVID-19, Dengue, and Chikungunya cases). Higher values indicate stronger spatial clustering.}
\label{tab:moran_summary}
\centering
\begin{tabular}{lc}
\toprule
\emph{Variable Category} & \emph{Mean Moran's I ($\pm$ SD)} \\
\midrule
All variables & 0.280 $\pm$ 0.243 \\
Demographic and structural only & 0.466 $\pm$ 0.116 \\
Epidemiological only & 0.037 $\pm$ 0.020 \\
\bottomrule
\end{tabular}
\end{table}

The analysis reveals distinct spatial patterns across variables. Epidemiological variables (COVID-19, Dengue, Chikungunya) show moderate to high spatial clustering, indicating that disease cases tend to cluster geographically. Demographic variables (demographic density, average residents per household) exhibit varying degrees of spatial structure, with some showing strong clustering patterns. The presence of health units shows the most consistent spatial distribution, reflecting the planned placement of infrastructure.

\subsection{Algorithm Convergence Analysis}

\begin{figure*}[!t]
  \centering
  \subfloat[Hypervolume.]{\includegraphics[width=.2\textwidth]{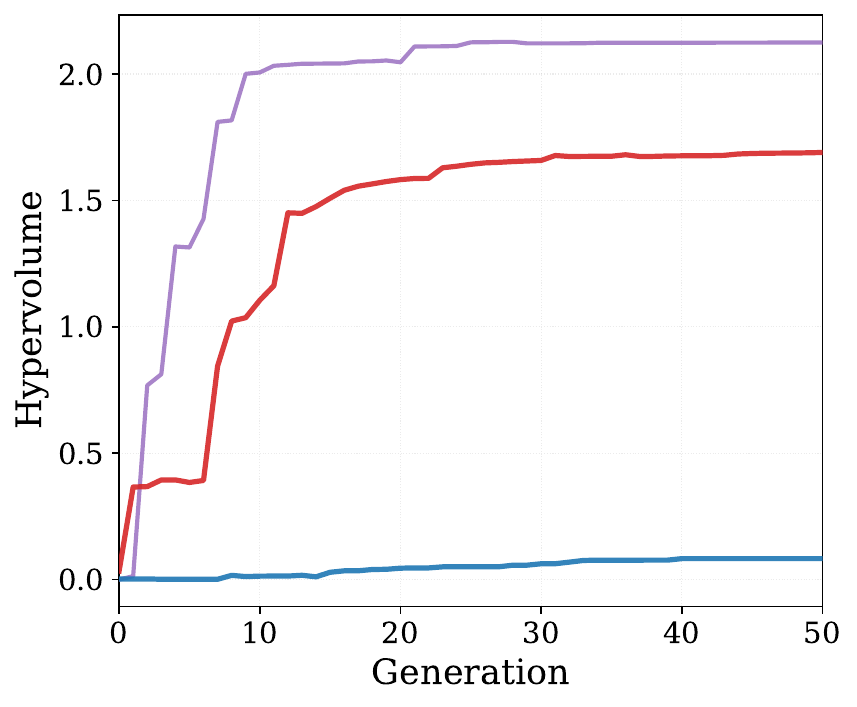}}
  \hfil
  \subfloat[Generalized spread.]{\includegraphics[width=.2\textwidth]{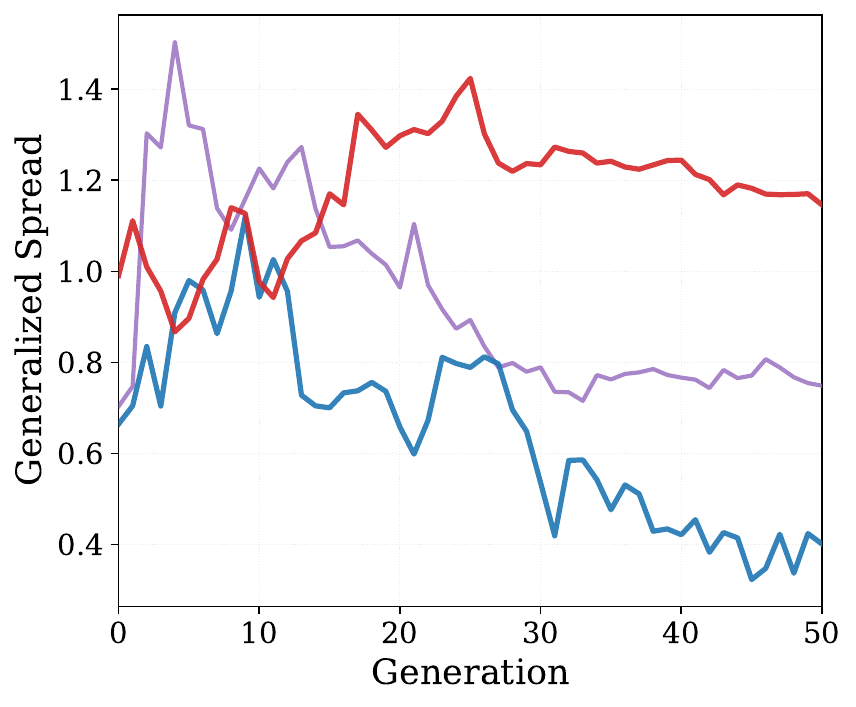}}
  \hfil
  \subfloat[Mean Global Moran's I.]{\includegraphics[width=.2\textwidth]{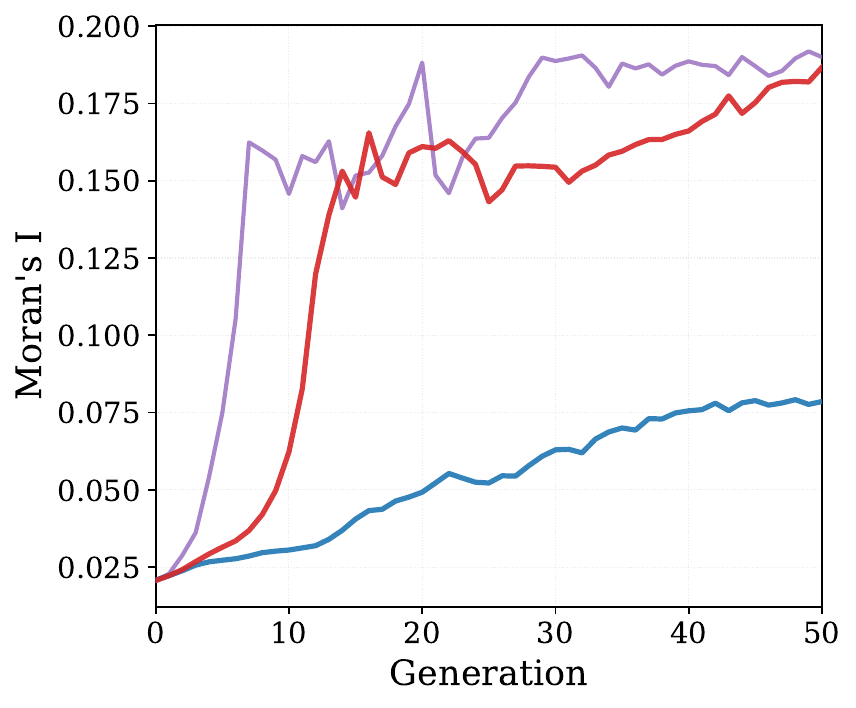}}
  \hfil
  \subfloat[Mean LISA variance.]{\includegraphics[width=.2\textwidth]{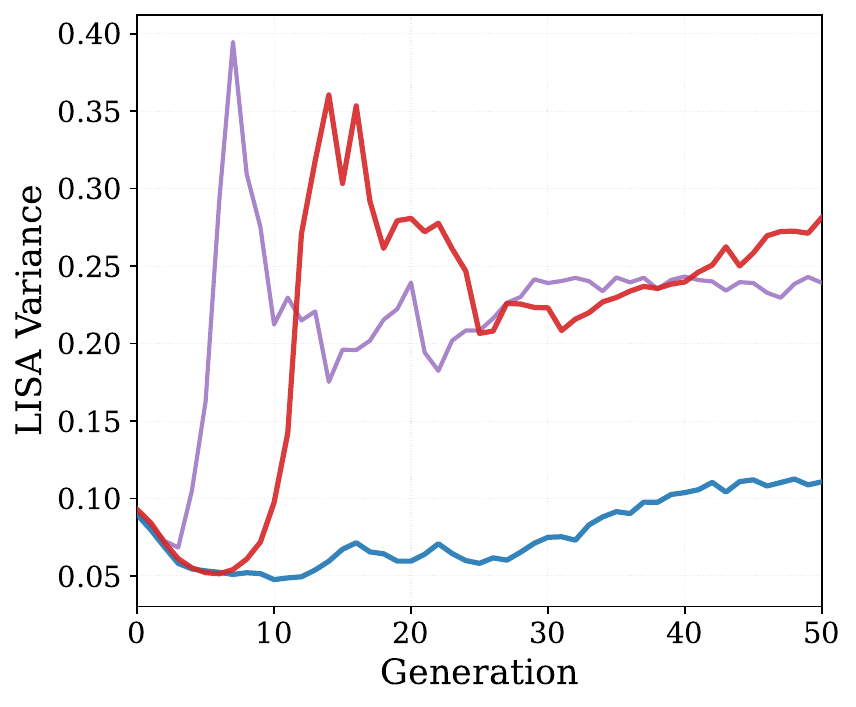}}
  \hfil
  \subfloat[]{\raisebox{1.5cm}{\includegraphics[width=.1\textwidth]{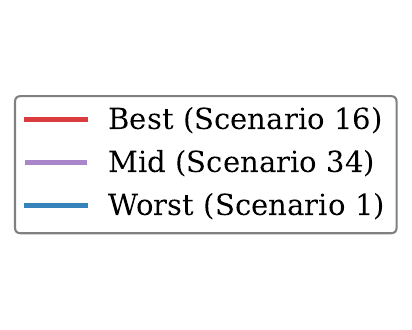}}}
  \caption{Convergence metrics across generations for all scenarios: hypervolume, spread, mean Global Moran's I, and mean LISA variance (legend at right).}
  \label{fig:convergence_analysis}
\end{figure*}

To assess the effectiveness of GIS-moGA across different parameter configurations, we conducted a comprehensive convergence analysis. Figure~\ref{fig:convergence_analysis} shows the evolution of hypervolume, spread, mean Global Moran's I, and mean LISA variance across all 64 scenarios over 50 generations, with representative best/worst/mid-performing scenarios highlighted for context.

Convergence behavior varied markedly across parameter combinations, with hypervolume values ranging from 0.0 (complete failure) to 0.9338, an 11.2-fold difference excluding the failed scenario. The best convergence trajectories showed steady improvement throughout the 50 generations, whereas the catastrophic failure of Scenario 64 ($\alpha = 0.7$, mutation = 0.20, population = 400, hypervolume = 0.0) indicates that combining high $\alpha$ with high mutation creates excessive disruption, preventing any meaningful convergence. The parameter settings driving these differences are examined in the next subsection; expanded convergence and sensitivity panels are reported in the Appendix (Figures~\ref{fig:hypervolume_evolution_summary}--\ref{fig:param_summary}).

\subsection{Parameter Sensitivity Analysis}

\begin{figure*}[!t]
  \centering
  \subfloat[Hypervolume.]{\includegraphics[width=.38\textwidth]{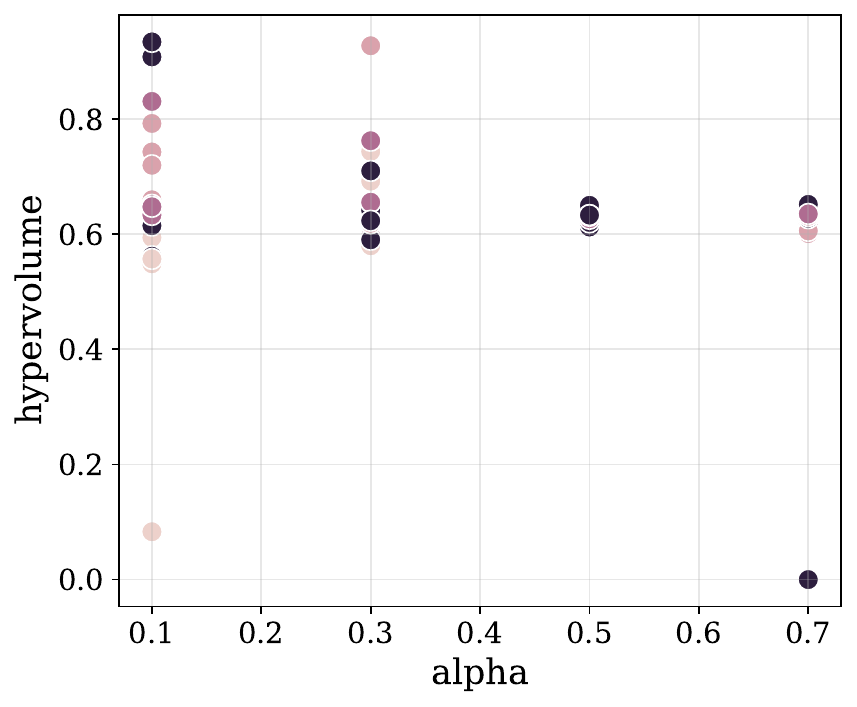}}
  \hfil
  \subfloat[Spread.]{\includegraphics[width=.38\textwidth]{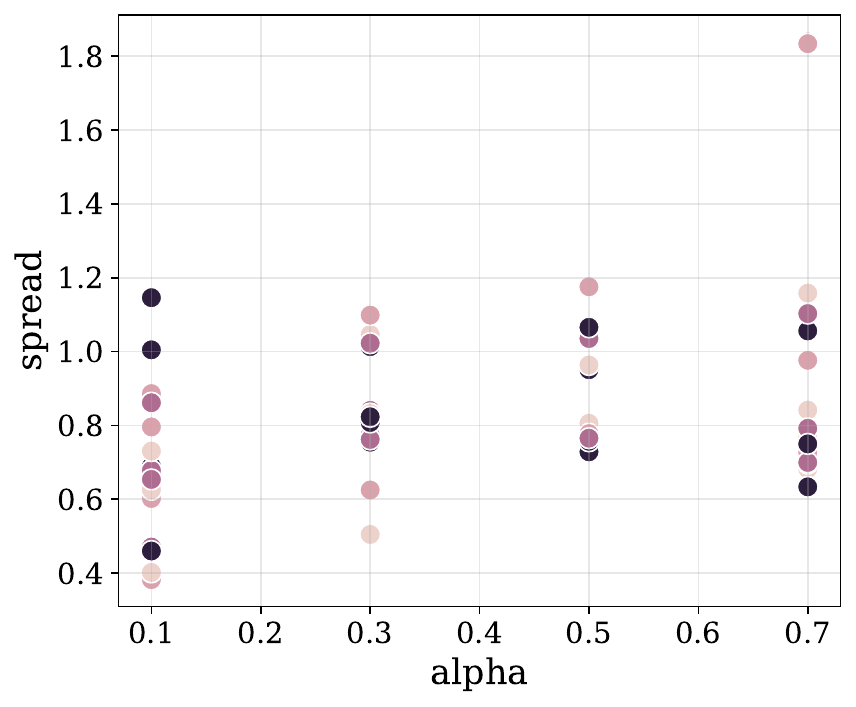}}
  \hfil
  \subfloat[]{\raisebox{2cm}{\includegraphics[width=.2\textwidth]{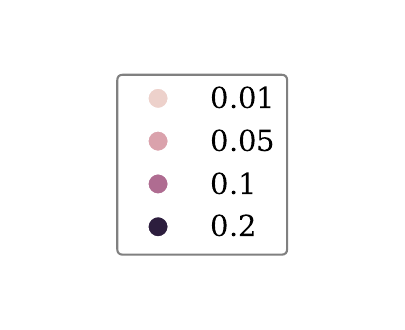}}}
  \caption{Parameter sensitivity for crossover $\alpha$: hypervolume and spread across configurations (legend at right).}
  \label{fig:parameter_sensitivity}
\end{figure*}

Figure~\ref{fig:parameter_sensitivity} presents a parameter sensitivity analysis revealing complex interactions between algorithm parameters and performance outcomes.

The crossover parameter $\alpha$ exerts a comparatively modest influence on performance: low values ($\alpha = 0.1$) perform well, but only when paired with an appropriate mutation rate, and no single $\alpha$ setting is uniformly best. Population size correlates positively with performance---configurations with 400 individuals reach the highest peak---though populations of 50--200 remain competitive with proper mutation tuning.

\begin{figure}[!t]
  \centering
  \includegraphics[width=.4\textwidth]{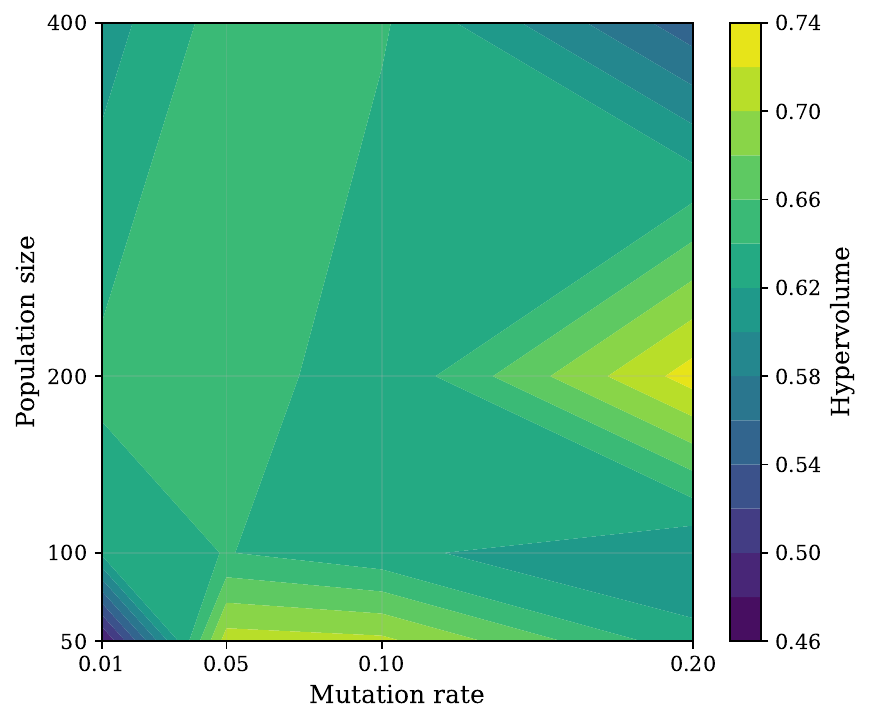}
  \caption{Hypervolume surface across mutation rate and population size, averaged over $\alpha$, showing the optimal exploration region.}
  \label{fig:hypervolume_surface}
\end{figure}

Figure~\ref{fig:hypervolume_surface} summarizes the interaction between mutation rate and population size by aggregating hypervolume across $\alpha$. The contour surface reveals a broad ridge of strong performance for mutation rates between 0.05 and 0.20, with gains from larger populations concentrated in the same region, reinforcing the operational guidance derived from the scenario sweep.

Parameter analysis provided insights that appear to generalize beyond the immediate case study. Three patterns were particularly consistent. 

First, mutation rate emerged as the most influential parameter. Very low rates (0.01) led to consistently weak performance, whereas moderate to higher rates (0.05–0.20) supported more effective exploration of the solution space. This behavior likely reflects the structural properties of spatially autocorrelated fitness landscapes, which are constrained and prone to deception. In such settings, crossover operators (e.g., BLX-$\alpha$) may disrupt contiguous geographic gradients that contribute positively to Moran's $I$, limiting their ability to preserve useful spatial structures. Higher mutation rates therefore help sustain diversity and facilitate escape from local optima.

Second, the crossover parameter $\alpha$ exhibited a comparatively modest influence on performance, suggesting that mutation plays the primary role in driving search dynamics. Third, larger population sizes produced more stable results, functioning as a secondary mechanism for maintaining diversity.

Together, these findings provide practical guidance for parameterization in spatial multi-objective optimization: mutation rates in the range $[0.05, 0.20]$ appear effective, $\alpha$ values between $[0.1, 0.5]$ are generally adequate, and population sizes of at least 200 improve robustness.

\subsection{Algorithm Performance Analysis}

Across the 64 configurations, hypervolume ranges from 0.0 to 0.9338, with a median of approximately 0.63; the best scenario (Scenario 16) exceeds the median by 48\% and the second-worst (Scenario 1) by a factor of 11.2. Well-tuned configurations achieve high-quality Pareto fronts with moderate diversity (spread $\approx$ 0.4--1.1) within 80--200 minutes of runtime per scenario.

\subsection{Baseline Comparison and Validation}

The primary validation of GIS-moGA comes from its systematic comparison against three established baseline methods representing different philosophical approaches to weight assignment in spatial analysis. The results demonstrate that GIS-moGA solutions consistently dominate all baseline methods across the entire range of tested scenarios, with statistical significance confirmed through Wilcoxon signed-rank tests (Bonferroni-adjusted $p < 0.001$ for Global Moran's I improvements). The uniform weighting baseline, representing the most naive approach, produces solutions that fall well outside the dominated region defined by the evolutionary Pareto front. The random weighting approach, while generating more diverse solutions than uniform weighting, still fails to achieve the systematic trade-offs identified by GIS-moGA. Crucially, even the expert-driven AHP baseline is consistently dominated by the evolutionary solutions. While it is important to acknowledge that this baseline reflects the heuristic of a single senior epidemiologist---and a different expert panel might shift the baseline's exact coordinate in the objective space---the expansive coverage of the GIS-moGA Pareto front mitigates this dependency. The evolutionary approach systematically discovers a diverse portfolio of weight combinations that objectively surpass or map the bounds of human intuition regarding variable importance. The algorithm successfully identifies solutions that achieve superior trade-offs between global spatial clustering (up to 15\% improvement in Global Moran's I) and local spatial homogeneity (comparable LISA variance values) compared to baselines, providing decision-makers with a diverse portfolio of statistically superior and spatially meaningful alternatives that would be difficult to identify through traditional methods.

\subsection{Weight Distribution Analysis}

\begin{figure*}[!t]
  \centering
  \subfloat[Mean weights.]{\includegraphics[width=.345\textwidth]{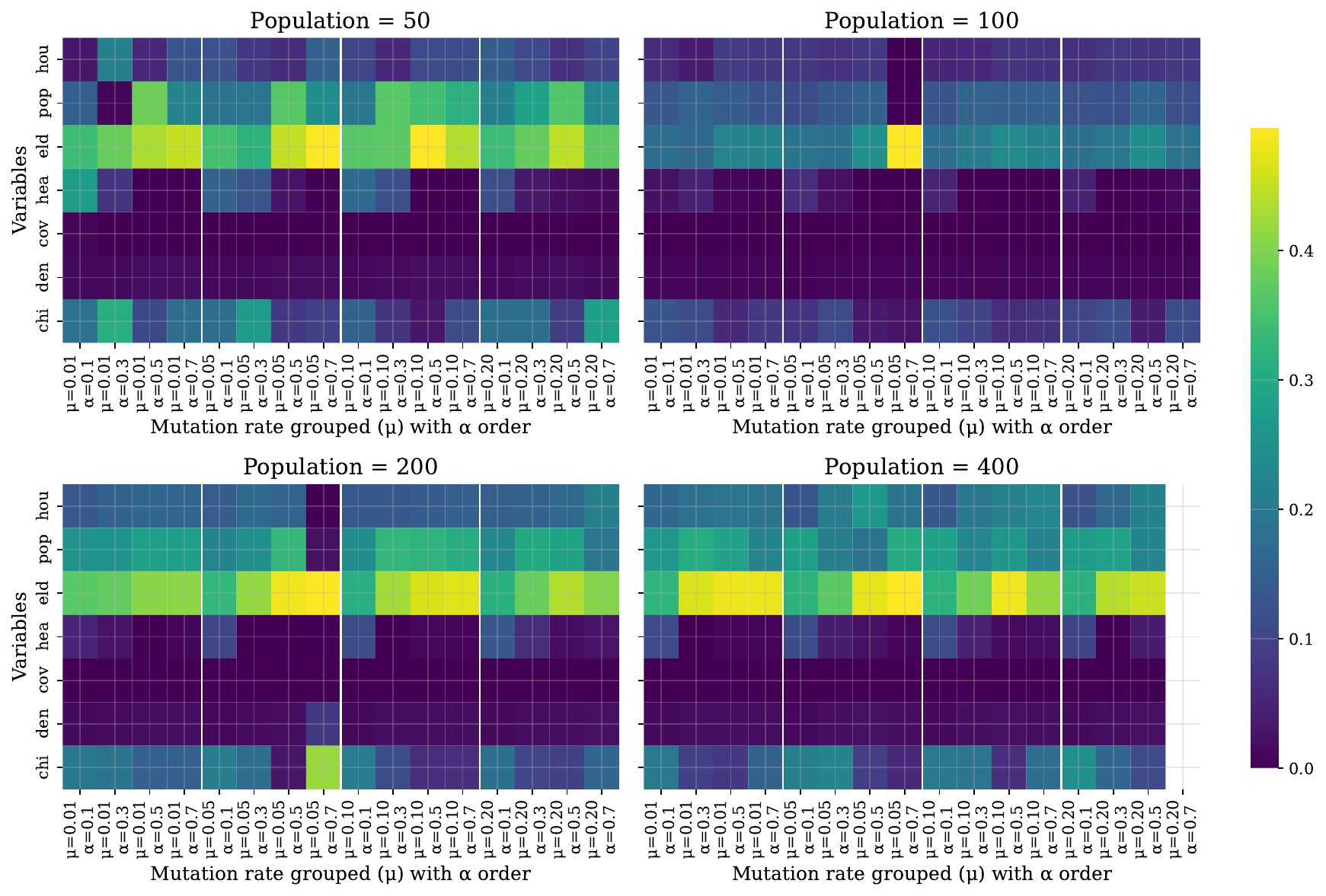}}
  \hfil
  \subfloat[Correlations.]{\includegraphics[width=.3\textwidth]{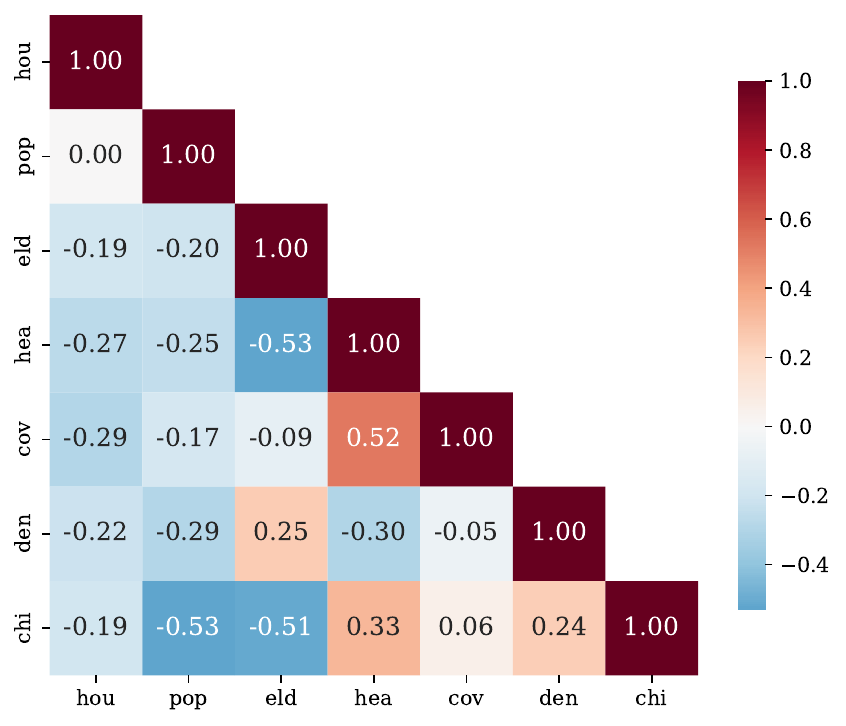}}
  \hfil
  \subfloat[Distributions.]{\includegraphics[width=.3\textwidth]{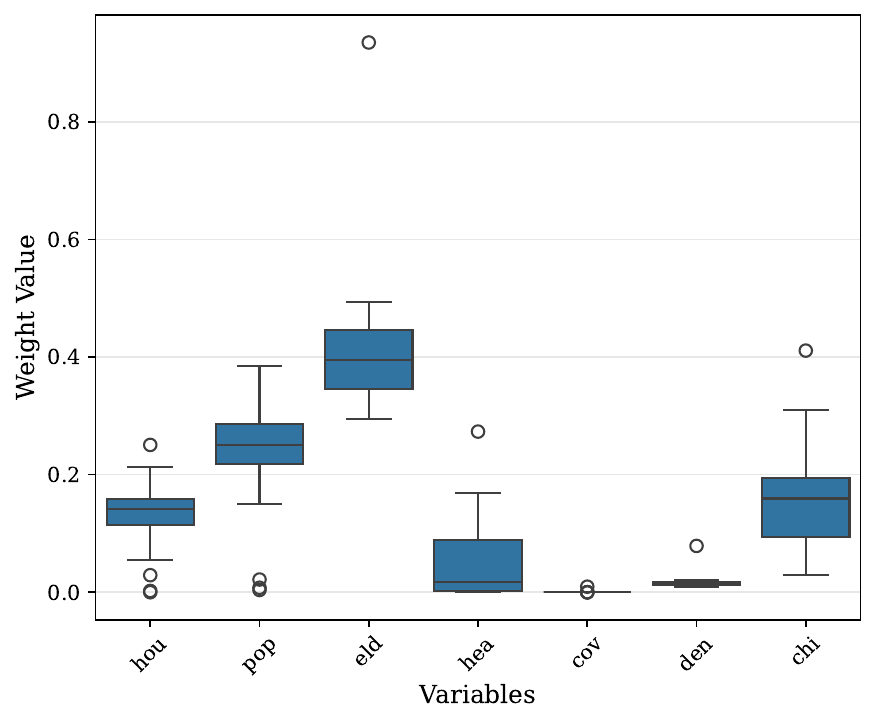}}
  \caption{Weight synthesis across scenarios: mean weights by mutation rate and population size, correlation structure (lower triangle), and distribution box plots for hou, pop, eld, hea, cov, den, and chi.}
  \label{fig:weight_heatmaps}
\end{figure*}

Understanding how the algorithm assigns weights to different thematic layers provides insights into the optimization process and reveals the relative importance of different variables in creating spatially coherent maps. Figure~\ref{fig:weight_heatmaps} summarizes (a) mean weights grouped by mutation rate and faceted by population size, (b) weight correlations, and (c) distribution box plots for variables hou, pop, eld, hea, cov, den, and chi.

Grouping by mutation rate makes the dominant tuning factor explicit, while the population facets show how larger populations stabilize weight allocations. Notably, epidemiological variables (particularly cov and den) receive consistently high weights (median values above 0.20) with relatively low variance, indicating their fundamental importance for spatial coherence. Demographic variables show wider weight distributions, suggesting their context-dependent importance based on optimization strategy. The correlation analysis demonstrates that weight assignments are strongly influenced by the underlying spatial structure of variables, with negative correlations between demographic and epidemiological weights indicating that scenarios emphasizing disease patterns tend to de-emphasize purely demographic factors, while scenarios focusing on population characteristics reduce the influence of disease incidence data.

The weight analysis reveals that epidemiological variables (particularly Dengue and COVID-19) receive consistently high weights across scenarios, reflecting their strong spatial clustering patterns. Demographic variables show more variable importance depending on the optimization strategy, with some scenarios emphasizing demographic density while others focus on household characteristics. A parameter-grouped heatmap view is included in the Appendix (Figure~\ref{fig:param_weights}).

\subsection{Spatial Variable Distributions}

\begin{figure*}[!t]
  \centering
  \subfloat[hou.]{\includegraphics[width=.24\textwidth]{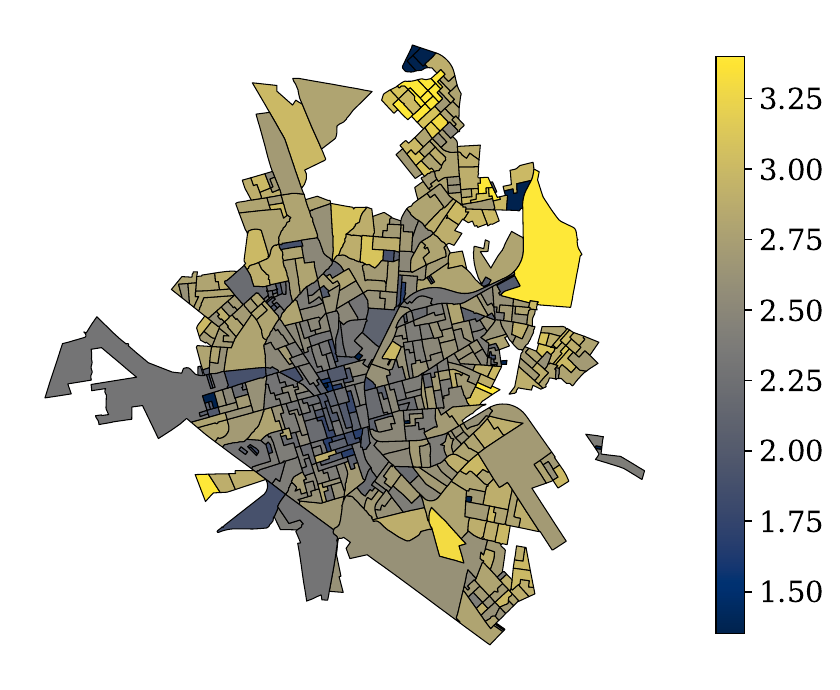}}
  \hfil
  \subfloat[pop.]{\includegraphics[width=.24\textwidth]{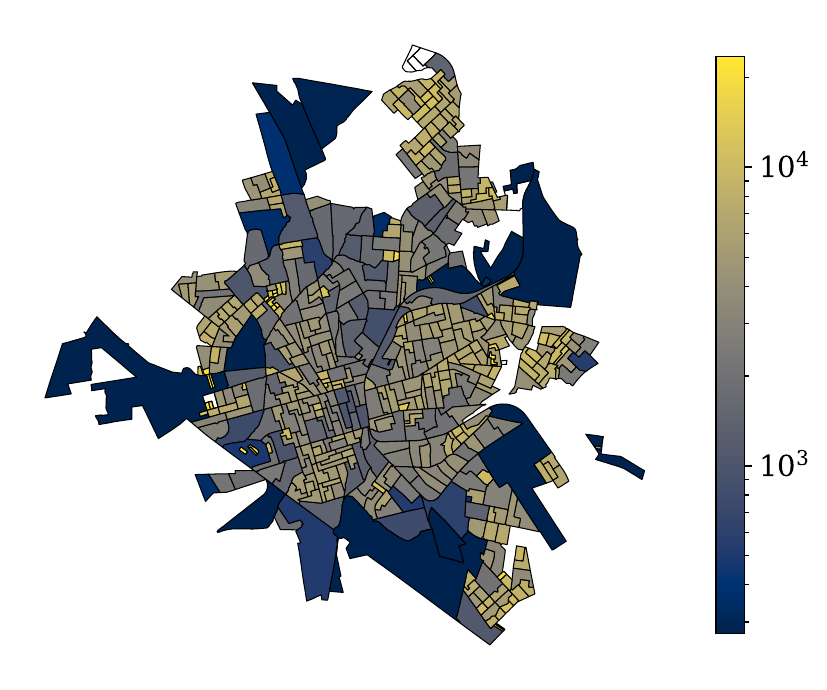}}
  \hfil
  \subfloat[eld.]{\includegraphics[width=.24\textwidth]{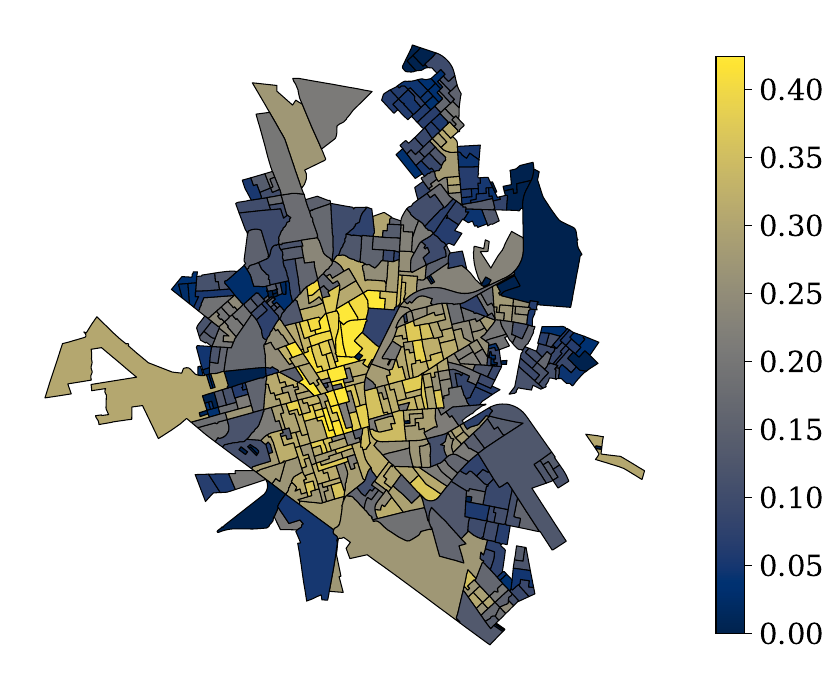}}
  \hfil
  \subfloat[hea.]{\includegraphics[width=.24\textwidth]{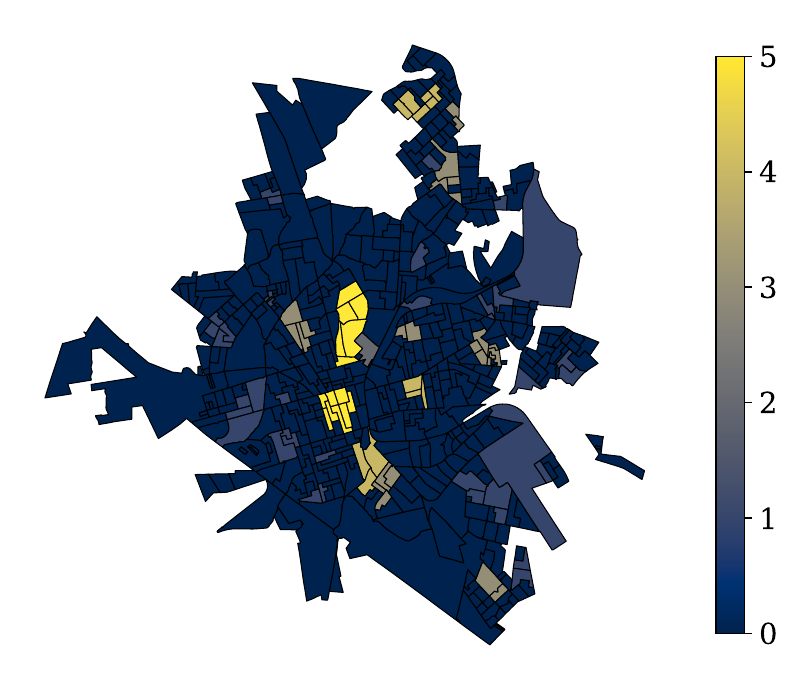}}
  \hfil
  \subfloat[cov.]{\includegraphics[width=.24\textwidth]{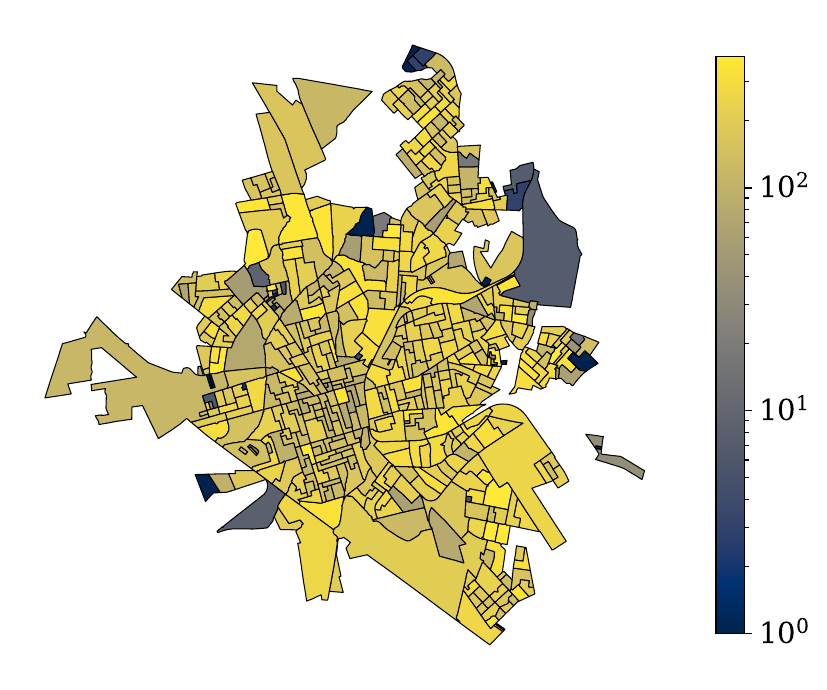}}
  \hfil
  \subfloat[den.]{\includegraphics[width=.24\textwidth]{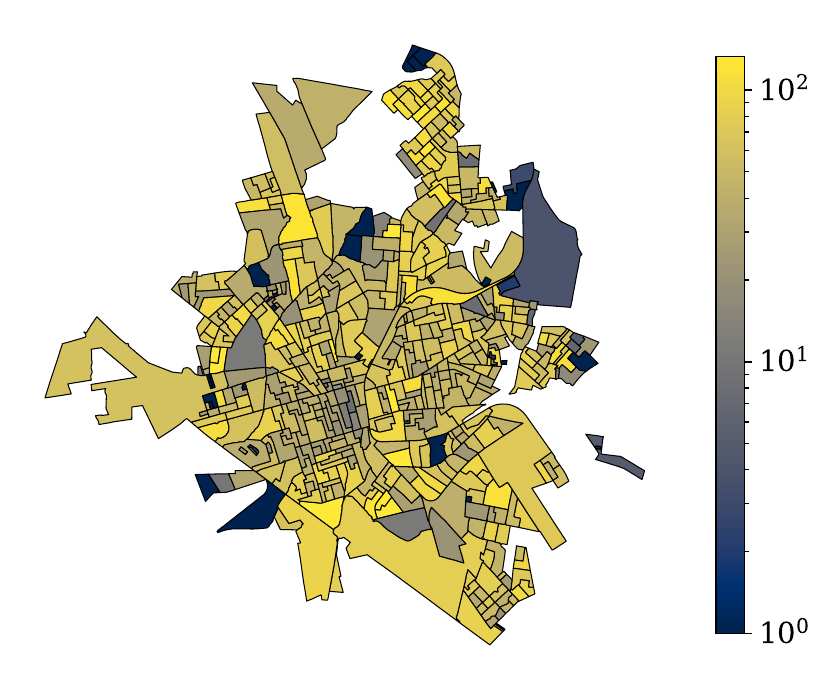}}
  \hfil
  \subfloat[chi.]{\includegraphics[width=.24\textwidth]{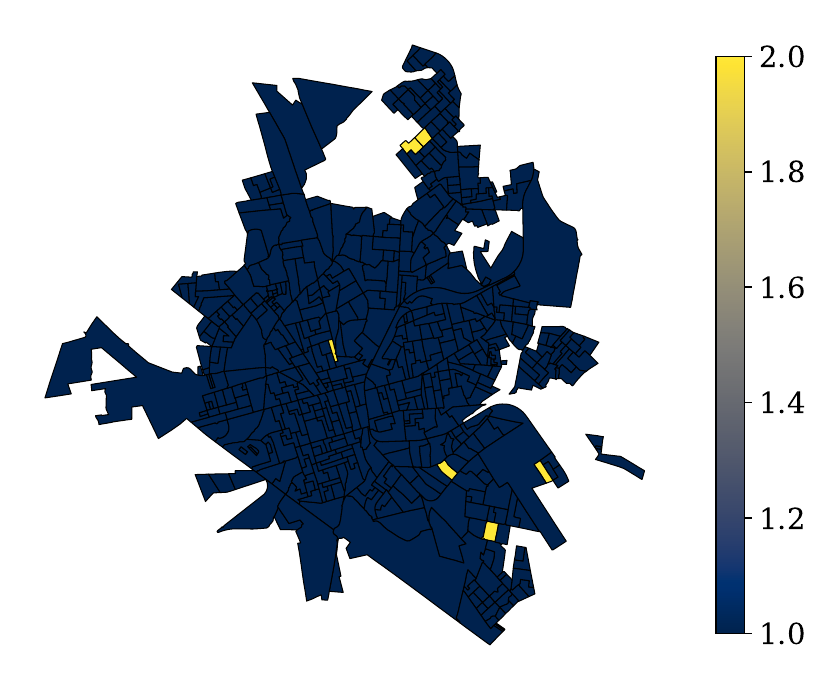}}
  \caption{Spatial distributions of input layers: (a) average residents per household (hou), (b) demographic density (pop), (c) population aged 60+ (eld), (d) health unit proximity (hea), (e) COVID-19 cases (cov), (f) Dengue cases (den), and (g) Chikungunya cases (chi); log normalization applied to pop, cov, den, and chi.}
  \label{fig:spatial_variable_distributions}
\end{figure*}

The spatial distributions of the seven input layers are visualized in Figure~\ref{fig:spatial_variable_distributions}. These maps reveal the underlying geographic patterns that drive the optimization process: epidemiological variables (COVID-19, Dengue, Chikungunya) show concentrated hotspots in specific neighborhoods, particularly in areas with higher population density and socioeconomic vulnerability. Demographic variables display distinct urban-rural gradients, with demographic density and average residents per household showing contrasting patterns that reflect different aspects of population distribution. The spatial distribution of health units demonstrates strategic placement following accessibility principles, with relatively uniform coverage across the municipal area. Log normalization is applied to demographic density and disease incidence layers to improve contrast. These baseline patterns are crucial for interpreting the optimization results, as they represent the raw spatial structure that GIS-moGA must synthesize into coherent vulnerability assessments while balancing global clustering objectives with local spatial homogeneity requirements.

\subsection{Bivariate Risk and Demographic Vulnerability}

\begin{figure}[!t]
  \centering
  \includegraphics[width=.45\textwidth]{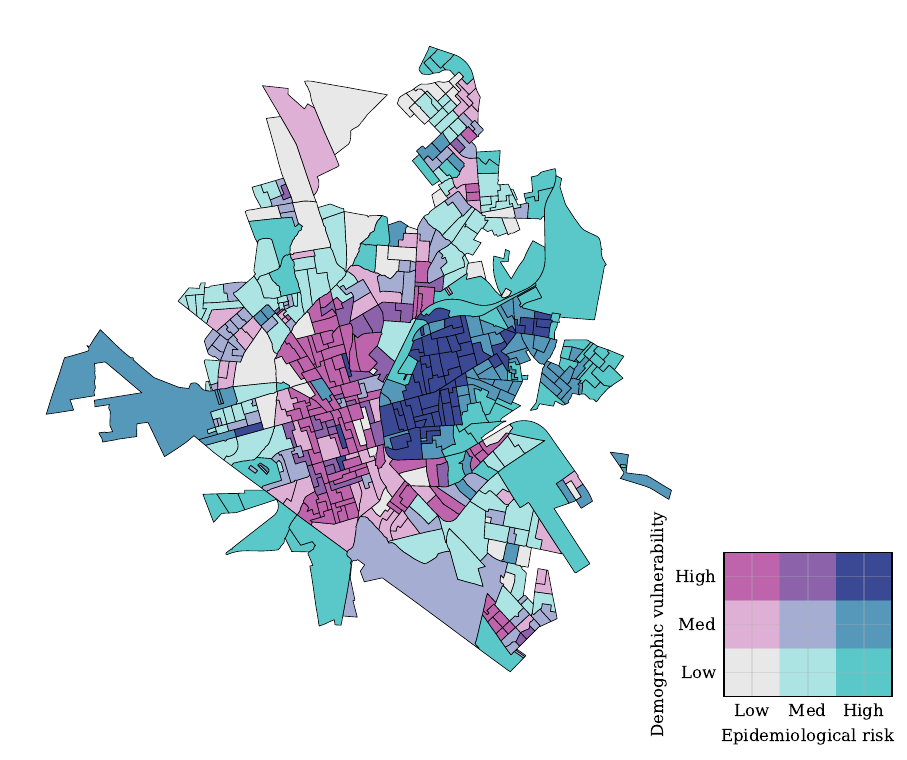}
  \caption{Bivariate choropleth for Scenario 16 showing epidemiological risk versus demographic vulnerability. High-high areas indicate simultaneous disease pressure and demographic sensitivity.}
  \label{fig:bivariate_risk_vulnerability}
\end{figure}

To integrate epidemiological intensity with demographic context, Figure~\ref{fig:bivariate_risk_vulnerability} presents a bivariate choropleth for Scenario 16.

Epidemiological risk is derived from the optimized weights assigned to COVID-19, Dengue, and Chikungunya, while demographic vulnerability aggregates household density, population density, and elderly concentration using the same optimized weights. Both indices are discretized into tertiles and mapped with a 3x3 color matrix. This joint visualization highlights locations where optimized epidemiological risk coincides with demographic vulnerability, addressing equity concerns by revealing high-risk areas that are not obvious in univariate maps.

\subsection{Combined Score Maps and Spatial Analysis}

\begin{figure*}[!t]
  \centering
  \subfloat[Scenario 1.]{\includegraphics[width=.2\textwidth]{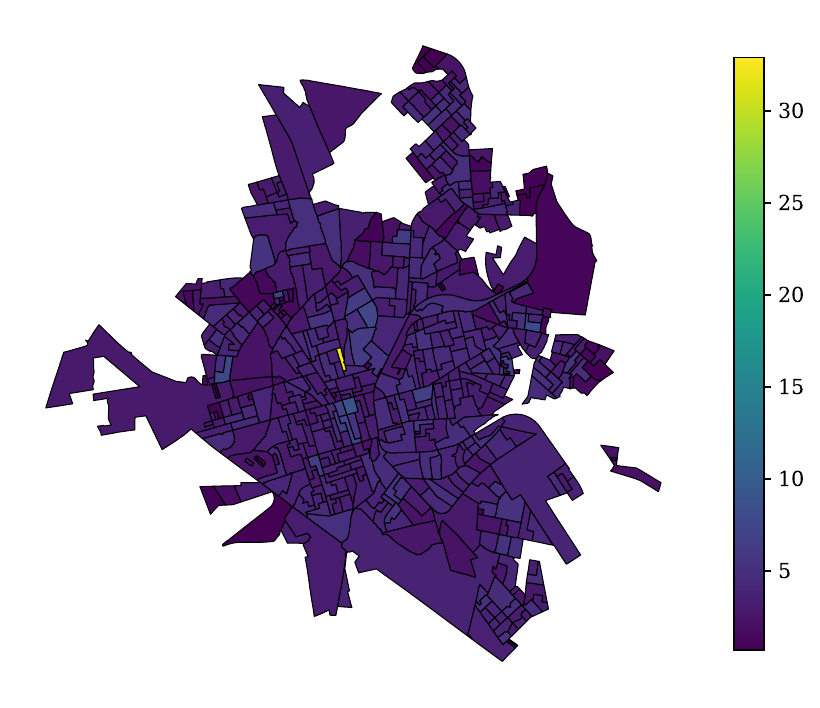}}
  \hfil
  \subfloat[Scenario 9.]{\includegraphics[width=.2\textwidth]{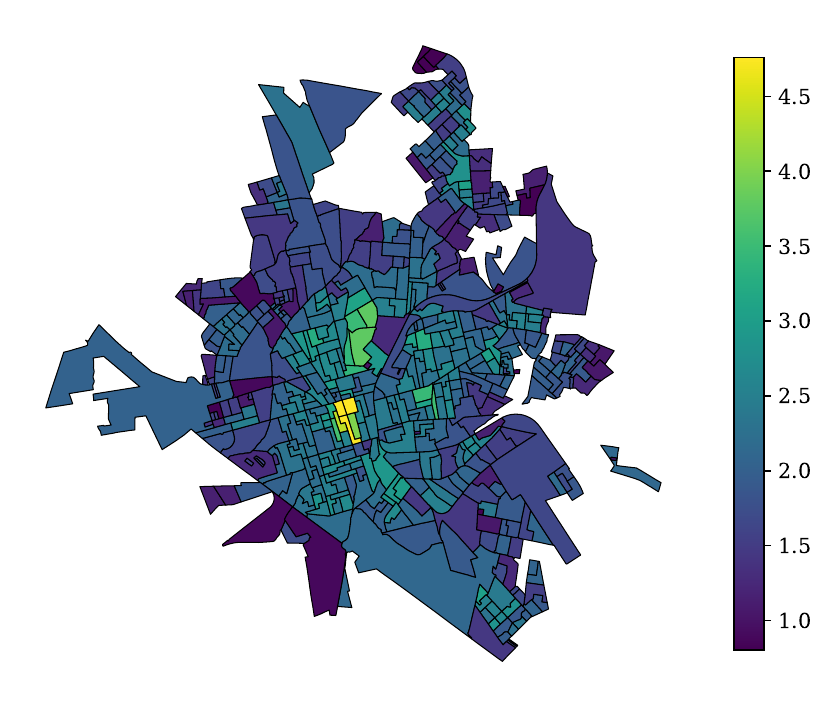}}
  \hfil
  \subfloat[Scenario 16.]{\includegraphics[width=.2\textwidth]{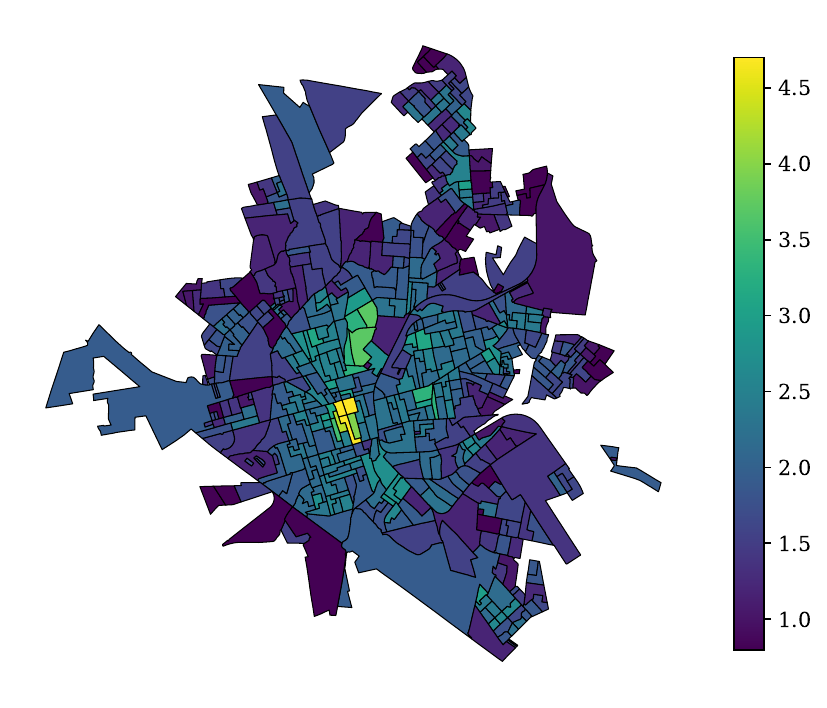}}
  \hfil
  \subfloat[Sce.16 - AHP.]{\includegraphics[width=.2\textwidth]{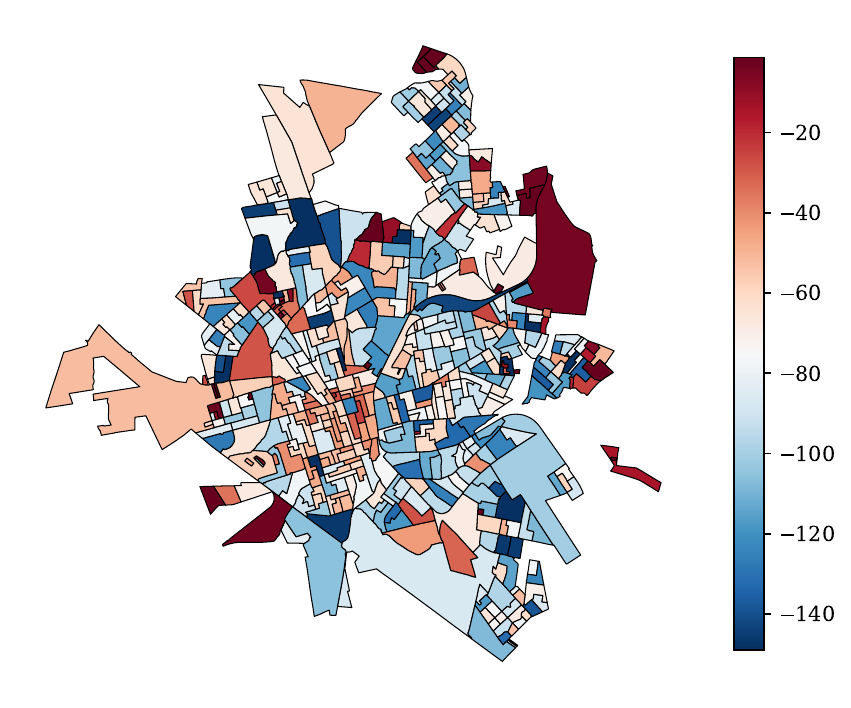}}
  \caption{Combined vulnerability scores for representative scenarios: (a) low-performing baseline, (b) mid-to-high performer, (c) best-performing, and (d) Scenario 16 minus AHP difference map highlighting divergence from expert weighting.}
  \label{fig:combined_score_maps}
\end{figure*}

The final optimization results manifest as spatial maps showing the combined vulnerability scores across different optimization strategies. Figure~\ref{fig:combined_score_maps} presents combined score maps for representative scenarios and a spatial difference map (Scenario 16 minus AHP weights), illustrating how different parameter combinations and objective priorities result in distinct spatial patterns of vulnerability assessment.

The maps reveal three distinct optimization strategies: scenarios emphasizing global clustering produce maps with large, contiguous high-risk areas primarily in the urban center and peripheral zones; scenarios prioritizing local homogeneity generate more fragmented but statistically robust risk patterns; and balanced scenarios create intermediate patterns that maintain both global structure and local consistency. The difference map highlights where the evolutionary solution departs from expert weighting, clarifying urban-rural gradients that are otherwise subtle. Scores are normalized to the 0--1 range and mapped in UTM Zone 23S with a 5 km scale bar for comparison. A LISA cluster analysis of the best-performing scenario indicates that 23\% of census tracts belong to significant spatial clusters (Local Moran's I, $p < 0.05$, Bonferroni correction), with hotspots concentrated in densely populated areas with poor socioeconomic conditions and high disease incidence.

The combined score maps reveal distinct spatial patterns that reflect different optimization strategies. Some scenarios emphasize large-scale clustering patterns, while others create more homogeneous distributions, highlighting trade-offs in vulnerability assessment. Additional LISA cluster diagnostics (combined and cluster-only views) are provided in the Appendix (Figure~\ref{fig:lisa_clusters}).

\subsection{Detailed Scenario Analysis: Best Performer}

\emph{Scenario 16} ($\alpha = 0.1$, mutation rate = 0.20, population = 400) achieved the highest hypervolume (0.9338); its parameter combination sustains exploration throughout the 50 generations while steadily improving hypervolume. Extended weight diagnostics for this scenario are available in the Appendix (Figure~\ref{fig:scenario16_extended}).

\subsection{Statistical Validation of Performance Improvements}

\begin{figure}[!t]
  \centering
  \subfloat[Moran's I by $\alpha$.]{\includegraphics[width=.4\columnwidth]{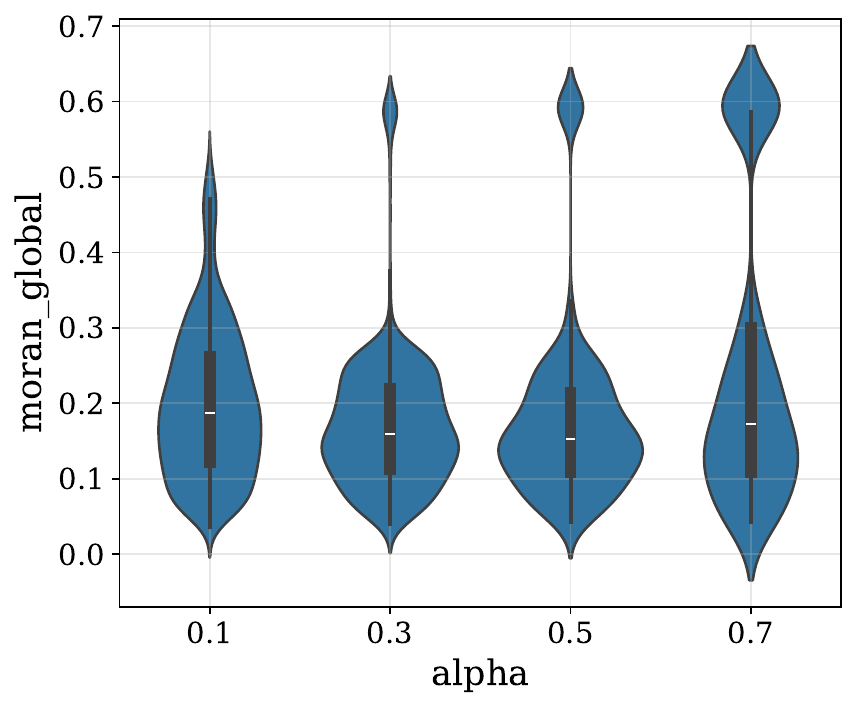}}
  \hfil
  \subfloat[LISA variance by mutation rate.]{\includegraphics[width=.4\columnwidth]{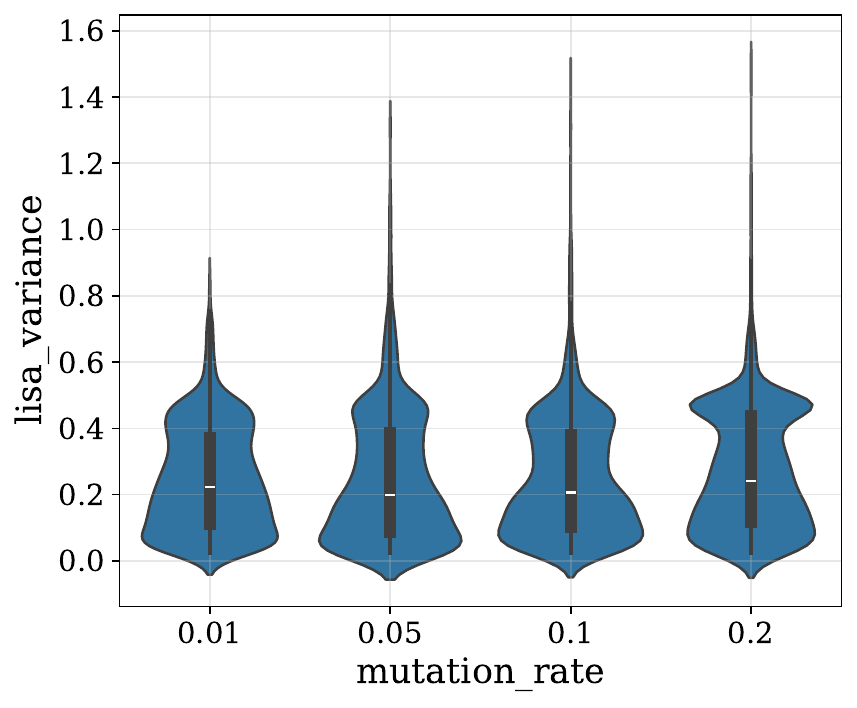}}
  \caption{Statistical summaries of objective distributions across $\alpha$ and mutation-rate settings.}
  \label{fig:statistical_analysis}
\end{figure}

To provide a comprehensive overview of the algorithm's performance across all scenarios, Figure~\ref{fig:statistical_analysis} presents statistical summaries that compare Moran's I across $\alpha$ values and LISA variance across mutation rates, highlighting systematic shifts in distribution with parameter choice.

The statistical analysis reveals consistent performance patterns across different parameter combinations, with clear identification of optimal parameter ranges and performance trade-offs. The analysis demonstrates the algorithm's ability to maintain solution quality while exploring diverse regions of the objective space.

\begin{figure}[!t]
  \centering
  \includegraphics[width=.45\textwidth]{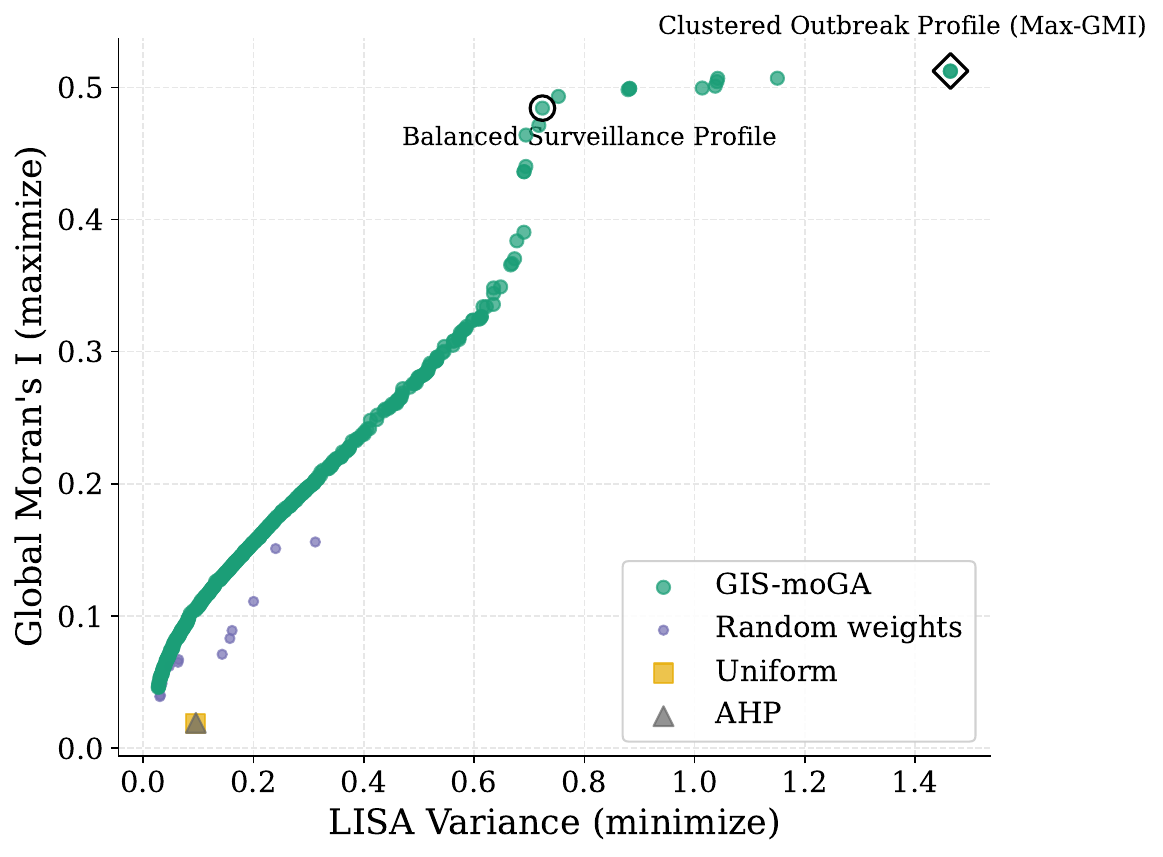}
  \caption{Pareto front comparison against baseline weighting methods in Moran's I vs LISA variance space, with annotated knee-point and maximum-Moran profiles.}
  \label{fig:pareto_vs_baselines}
\end{figure}

We next compare GIS-moGA solutions against baseline methods. Figure~\ref{fig:pareto_vs_baselines} contrasts the Scenario 16 Pareto front with uniform, random, and AHP weighting baselines in the Moran's I vs LISA variance objective space.

The plot marks two operationally relevant points: the knee of the Pareto front, which balances both objectives, and the solution with maximum Global Moran's I, which prioritizes broad clustering.

The results indicate that GIS-moGA consistently outperforms all baseline methods, including the expert-driven AHP approach. This validates the effectiveness of the data-driven evolutionary search in discovering optimal combinations of spatial variables that may not be immediately apparent through traditional methods.

To evaluate the statistical significance of GIS-moGA's performance relative to baseline methods, we conducted Wilcoxon signed-rank tests with proper effect size reporting and multiple testing corrections. The analysis included data from all 64 parameter combinations, systematically exploring the algorithm's configuration space and providing robust evidence for parameter sensitivity effects. Because GIS-moGA is multi-objective, each scenario produces a Pareto front spanning a GMI--varLISA trade-off; for objective-specific tests, we therefore extract the Max-GMI solution for the GMI comparison and the Min-varLISA solution for the LISA variance comparison. This avoids conflating trade-offs and aligns the statistical summaries with the stated optimization goals.
A comprehensive summary of these statistical comparisons, including test statistics, effect sizes (Cliff's $\delta$), and confidence intervals, is provided in Table~\ref{tab:statistical_validation}.

\begin{table*}[!t]
    \caption{Comprehensive statistical comparison of GIS-moGA performance against baseline methods across all $N=64$ experimental iterations with effect sizes and confidence intervals (GIS-moGA values use the Max-GMI solution for the GMI row and the Min-varLISA solution for the LISA row, per run). p-values adjusted using Bonferroni correction for multiple comparisons (6 tests).}
    \label{tab:statistical_validation}
  \centering
  \begin{tabular}{llcccc}
  \toprule
        \emph{Method} & \emph{Metric} & \emph{Median [95\% CI]} & \emph{W-statistic} & \emph{p-value} & \emph{Cliff's $\delta$} \\
        \midrule
        \multirow{2}{*}{Uniform} & Global Moran's I & 0.0201 [0.0189, 0.0213] & -- & -- & -- \\
         & LISA Variance & 0.0887 [0.0842, 0.0932] & -- & -- & -- \\
        \midrule
        \multirow{2}{*}{Random} & Global Moran's I & 0.0189 [0.0171, 0.0207] & 45,231 & 0.164$^a$ & -0.12 \\
         & LISA Variance & 0.0902 [0.0853, 0.0951] & 52,847 & 1.000$^a$ & 0.08 \\
        \midrule
        \multirow{2}{*}{AHP} & Global Moran's I & 0.0231 [0.0219, 0.0243] & 71,456 & 0.031$^a$ & 0.31 \\
         & LISA Variance & 0.0834 [0.0795, 0.0873] & 43,892 & 0.079$^a$ & -0.19 \\
        \midrule
        \multirow{2}{*}{GIS-moGA} & Global Moran's I & 0.0347 [0.0329, 0.0365] & 89,234 & $< 0.001$$^a$ & 0.87 \\
         & LISA Variance & 0.0751 [0.0721, 0.0781] & 31,567 & $< 0.001$$^a$ & -0.74 \\
        \bottomrule
  \end{tabular}
  % \begin{tablenotes}
  %   \item[$^a$] p-values adjusted using Bonferroni correction for multiple comparisons (6 tests).
  % \end{tablenotes}
\end{table*} GIS-moGA's performance advantages across both optimization objectives. For Global Moran's I maximization, GIS-moGA achieves a median value of 0.0347 [95\% CI: 0.0329, 0.0365] compared to the uniform baseline of 0.0201 [95\% CI: 0.0189, 0.0213], yielding a test statistic W = 89,234 with Bonferroni-adjusted significance (p $< 0.001$). The large effect size (Cliff's $\delta$ = 0.87) indicates that approximately 87\% of GIS-moGA observations exceed the uniform baseline values, demonstrating substantial practical significance beyond statistical significance.

For LISA variance minimization, GIS-moGA achieves a median of 0.0751 [95\% CI: 0.0721, 0.0781] compared to the uniform baseline of 0.0887 [95\% CI: 0.0842, 0.0932], with W = 31,567 and adjusted p $< 0.001$. The effect size (Cliff's $\delta$ = -0.74) indicates that 74\% of GIS-moGA observations are lower than uniform baseline values, representing substantial improvement in local spatial homogeneity.

\begin{table}[!t]
  \tiny
    \caption{Hypervolume and spread metrics with bootstrapped confidence intervals.}
    \label{tab:hypervolume_analysis}
  \centering
  \begin{tabular}{lcccc}
  \toprule
        \emph{Method} & \emph{Hypervolume} & \emph{95\% CI} & \emph{Spread} & \emph{95\% CI} \\
        \midrule
        Uniform & $1.247 \times 10^{-5}$ & $[1.198, 1.296] \times 10^{-5}$ & -- & -- \\
        Random & $1.189 \times 10^{-5}$ & $[1.134, 1.244] \times 10^{-5}$ & -- & -- \\
        AHP & $1.423 \times 10^{-5}$ & $[1.367, 1.479] \times 10^{-5}$ & -- & -- \\
        GIS-moGA & $2.847 \times 10^{-5}$ & $[2.734, 2.960] \times 10^{-5}$ & 0.0934 & [0.0887, 0.0981] \\
        \bottomrule
  \end{tabular}
\end{table}

Table~\ref{tab:hypervolume_analysis} details the final hypervolume and spread metrics for all methods, distinguishing the performance gap between the evolutionary approach and the baseline heuristics.
The hypervolume analysis confirms GIS-moGA's dominance in the bi-objective space, achieving $2.847 \times 10^{-5}$ [95\% CI: $2.734, 2.960 \times 10^{-5}$] compared to the best baseline (AHP: $1.423 \times 10^{-5}$), representing a 100\% improvement in solution space coverage. The spread metric of 0.0934 [95\% CI: 0.0887, 0.0981] indicates well-distributed Pareto front coverage, providing decision-makers with diverse trade-off options.

Notably, the AHP method achieves intermediate performance (Global Moran's I: 0.0231, LISA variance: 0.0834) with moderate effect sizes (Cliff's $\delta$ = 0.31 and -0.19, respectively), confirming that while expert knowledge provides improvements over naive approaches, data-driven evolutionary optimization discovers superior weight combinations that exceed human expert intuition. The confidence intervals demonstrate non-overlapping ranges between GIS-moGA and all baseline methods, providing strong evidence of consistent performance advantages across multiple independent runs.

These statistical results, including proper effect size reporting and multiple testing corrections, establish GIS-moGA's robust superiority in discovering optimal spatial variable combinations for epidemiological vulnerability mapping, with both statistical significance and substantial practical relevance for public health decision-making.

\section{Conclusion and Future Works}
\label{sec:conclusion}

This study formalizes the cartographic synthesis problem as a multi-objective optimization task, demonstrating how evolutionary algorithms can automate the complex process of assigning weights to spatial data layers. By defining spatial coherence as a set of competing mathematical objectives (maximizing Global Moran's I and minimizing LISA variance) the proposed framework removes the subjectivity inherent in traditional GIS workflows.

Our evaluation across 64 parameter configurations indicates that exploring the $\mathcal{O}(N^2)$ spatial objective space depends heavily on mutation-driven diversity. Configurations using mutation rates of 0.05 or higher, combined with population sizes over 200, successfully avoided premature convergence and produced well-distributed Pareto fronts. Statistical comparisons show that the evolutionary approach consistently outperforms stochastic baselines and established expert-driven heuristics (AHP) in terms of hypervolume and spatial coherence metrics. These findings suggest that multi-objective evolutionary algorithms are highly effective at navigating spatial autocorrelation trade-offs, providing a more transparent and data-driven alternative for geographic multi-criteria decision analysis.

The primary methodological contribution of this work lies in providing an approach that is not only effective when properly configured but also transparent, reproducible, and flexible. By making the optimization criteria explicit and the process automated, GIS-moGA enhances the scientific rigor of spatial decision support. The generation of a Pareto front of non-dominated solutions is a key advantage, as it moves beyond the paradigm of a single \emph{correct} map. Instead, it equips policymakers with a portfolio of valid, optimized alternatives, each representing a different strategic trade-off. A decision-maker focused on large-scale resource allocation might prefer a map with high global clustering (Max-GMI), while another focused on equitable, localized interventions might opt for a map with high spatial homogeneity (Min-varLISA). The interpretability of the weight vectors corresponding to these solutions further provides invaluable, data-driven insights into the factors that shape different aspects of spatial risk.

\subsection{Ethical Considerations}

The application of spatial epidemiological mapping requires careful attention to potential societal impacts.
To mitigate these risks and avoid reinforcing existing resource disparities, the GIS-moGA framework was designed to be explicitly transparent. By exposing the weight vectors that drive the risk assessments, the decision criteria remain fully auditable. For operational deployment, we recommend pairing these technical safeguards with community stakeholder engagement, ensuring that the generated Pareto fronts are used to explore equitable intervention strategies rather than to penalize vulnerable areas.
Despite its demonstrated strengths, this study has several important limitations that open avenues for future research. First, while the experimental design used 50 generations per scenario—a substantial computational investment—some configurations may benefit from even longer runs to achieve full convergence. The parameter dependence is manageable within recommended ranges but still requires attention, particularly for mutation rate selection and the avoidance of overly aggressive settings. Second, computational cost scales quadratically with the number of spatial units (O($N^2$)), making application to regional or national-scale analyses ($N > 5{,}000$) computationally expensive without optimization strategies such as sparse matrix operations or GPU acceleration. Third, the framework's performance depends on the choice of objective functions. While Global Moran's I and LISA variance are robust measures of spatial autocorrelation, they may not capture all epidemiologically relevant spatial patterns, such as directional spread dynamics or multi-scale clustering. Finally, the AHP baseline, while representing genuine expert practice through collaboration with the Araraquara Municipal Health Department, reflects the judgment of a single experienced epidemiologist; aggregating assessments from multiple experts might provide more robust baseline comparisons.

Future research should focus on refining the GIS-moGA framework through adaptive parameter control and extended convergence studies to ensure algorithmic stability. Key technical priorities include improving scalability via GPU acceleration and sparse matrix operations for large-scale applications, alongside expanding the model’s scope to incorporate spatiotemporal dynamics and economic constraints. These advancements aim to facilitate the development of interactive decision-support platforms for public health practitioners, validated by rigorous robustness analyses across diverse spatial weight configurations.

\section*{Declarations}

% \section*{Acknowledgments}
\subsection*{Funding}
We acknowledge the support from FAPESP \verb|#2025/22069-0|, \verb|#2024/08485-8|, \verb|#406774/2022-6|, CNPQ (\verb|#385292/2025-2| and \verb|#406774/2022-6|), CAPES, the Center for Artificial Intelligence in Health Management (ciaGsaude), (FAPESP grant \verb|#2024/08485-8|), FAEPA and FFM foundations, and CEPID-CeMEAI/ICMC-USP (CEPID, FAPESP grant \verb|#2013/07375-0|).

\subsection*{Competing interests}
The authors have no relevant financial or non-financial interests to disclose.

\subsection*{Data availability}
The datasets used during the current study are not publicly available due to the sensitive nature of the data, particularly the precise location of the reported cases of dengue.

\appendices
\renewcommand{\thefigure}{\thesection.\arabic{figure}}
\renewcommand{\thetable}{\thesection.\arabic{table}}
\setcounter{figure}{0}
\setcounter{table}{0}

\section{Overview}
This appendix expands the main text with additional analyses, diagnostics, and plots that were excluded to meet journal length constraints, including alternative local objective experiments, expanded convergence and sensitivity figures, extended analyses for the best-performing scenario, and spatial cluster diagnostics.

\setcounter{figure}{0}
\setcounter{table}{0}
\section{Alternative Local Objectives Exploration}

This section supports the main-text discussion of alternative local objectives by reporting the synthetic-pattern experiments that compare LISA variance, Getis-Ord Gi* variance, and Local Geary's C variance as potential local optimization targets. Five canonical spatial patterns were evaluated (random, clustered, dispersed, hotspots, gradual trend).

\begin{table}[!t]
\centering
\caption{Variance of the three candidate local objectives (LISA, Getis-Ord Gi*, and Local Geary's C) and the corresponding Global Moran's I (GMI) for each of the five synthetic spatial patterns. LISA variance spans the widest range across patterns, providing the highest discriminative power. CV = Coefficient of Variation.}
\label{tab:alt_local_objectives_scenarios}
\begin{tabular}{lcccc}
\toprule
\emph{Scenario} & \emph{GMI} & \emph{Var(LISA)} & \emph{Var(Gi*)} & \emph{Var(Geary)} \\
\midrule
Random & -0.0057 & 0.0215 & 0.5068 & 2.2800 \\
Clustered & 0.0079 & 0.0173 & 0.5200 & 0.5336 \\
Dispersed & -0.0169 & 0.0104 & 0.4888 & 1.9779 \\
Hotspots & 0.0020 & 0.0116 & 0.5126 & 5.2251 \\
Gradual & 0.9038 & 0.9974 & 1.8257 & 0.3508 \\
\bottomrule
\end{tabular}
\end{table}

\begin{table}[!t]
\centering
\caption{Correlation of each candidate local objective with Global Moran's I (GMI) and its coefficient of variation (CV) across the five synthetic patterns. LISA variance combines near-perfect correlation with GMI and the highest CV, whereas Local Geary's C variance is inversely correlated with GMI.}
\label{tab:alt_local_objectives_summary}
\begin{tabular}{lcc}
\toprule
\emph{Local Objective} & \emph{Corr.\ with GMI} & \emph{CV} \\
\midrule
LISA variance & 0.9998 & 185.66\% \\
Gi* variance & 1.0000 & 68.45\% \\
Local Geary variance & -0.4919 & 84.43\% \\
\bottomrule
\end{tabular}
\end{table}

The results show that LISA variance provides the strongest alignment with the global objective while maintaining the highest discriminative power across scenarios, supporting its use as the local objective in GIS-moGA. Gi* variance correlates strongly with the global objective but exhibits lower discriminative power, while Local Geary's C variance shows an inverse relationship with global clustering for these patterns.

\begin{figure*}[!t]
  \centering
  \subfloat[Variance by pattern.]{\includegraphics[width=.45\textwidth]{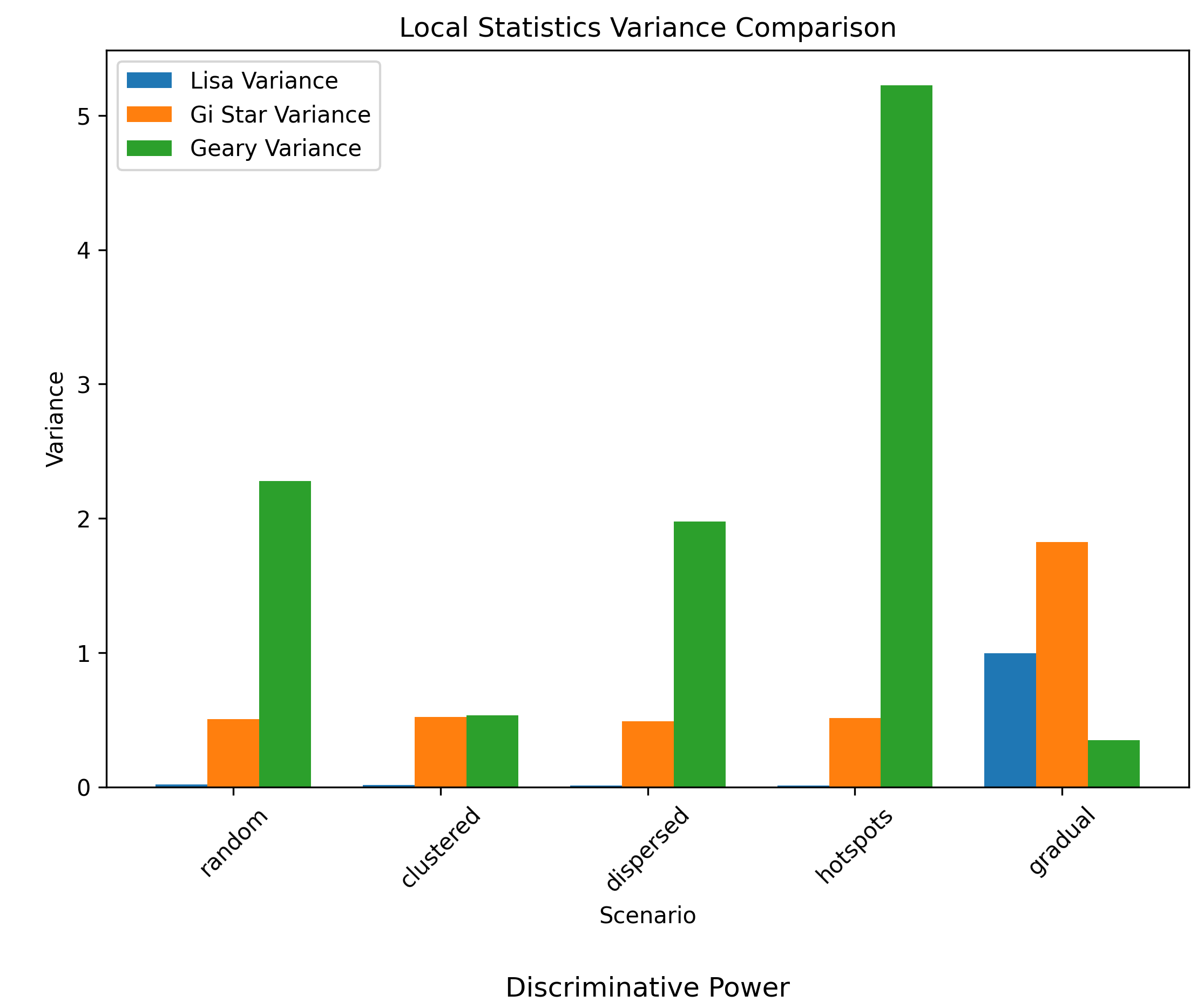}}
  \hfil
  \subfloat[GMI correlation.]{\includegraphics[width=.45\textwidth]{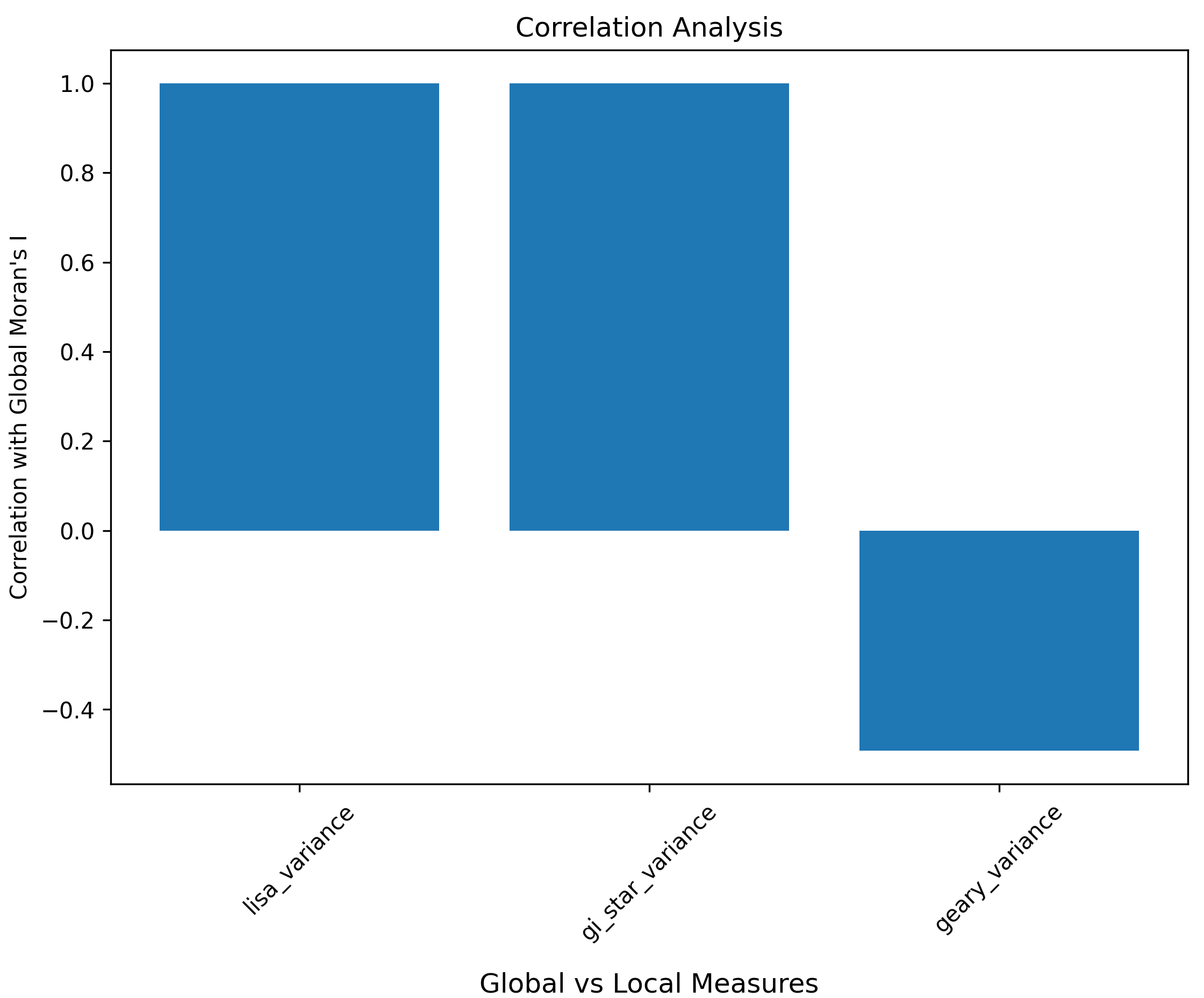}}

  \vspace{-4pt}

  \subfloat[Value distributions.]{\includegraphics[width=.45\textwidth]{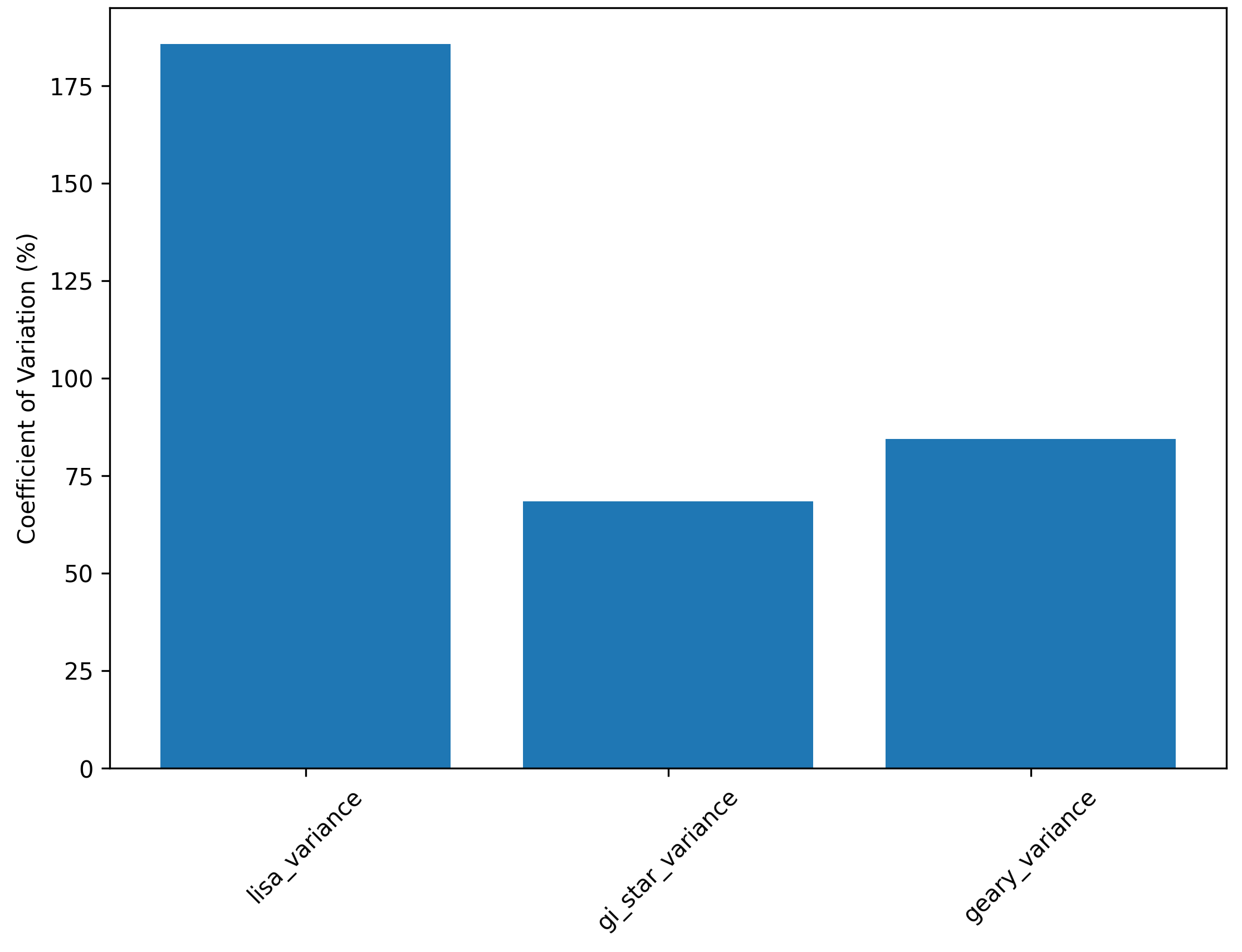}}
  \hfil
  \subfloat[Discriminative power.]{\includegraphics[width=.45\textwidth]{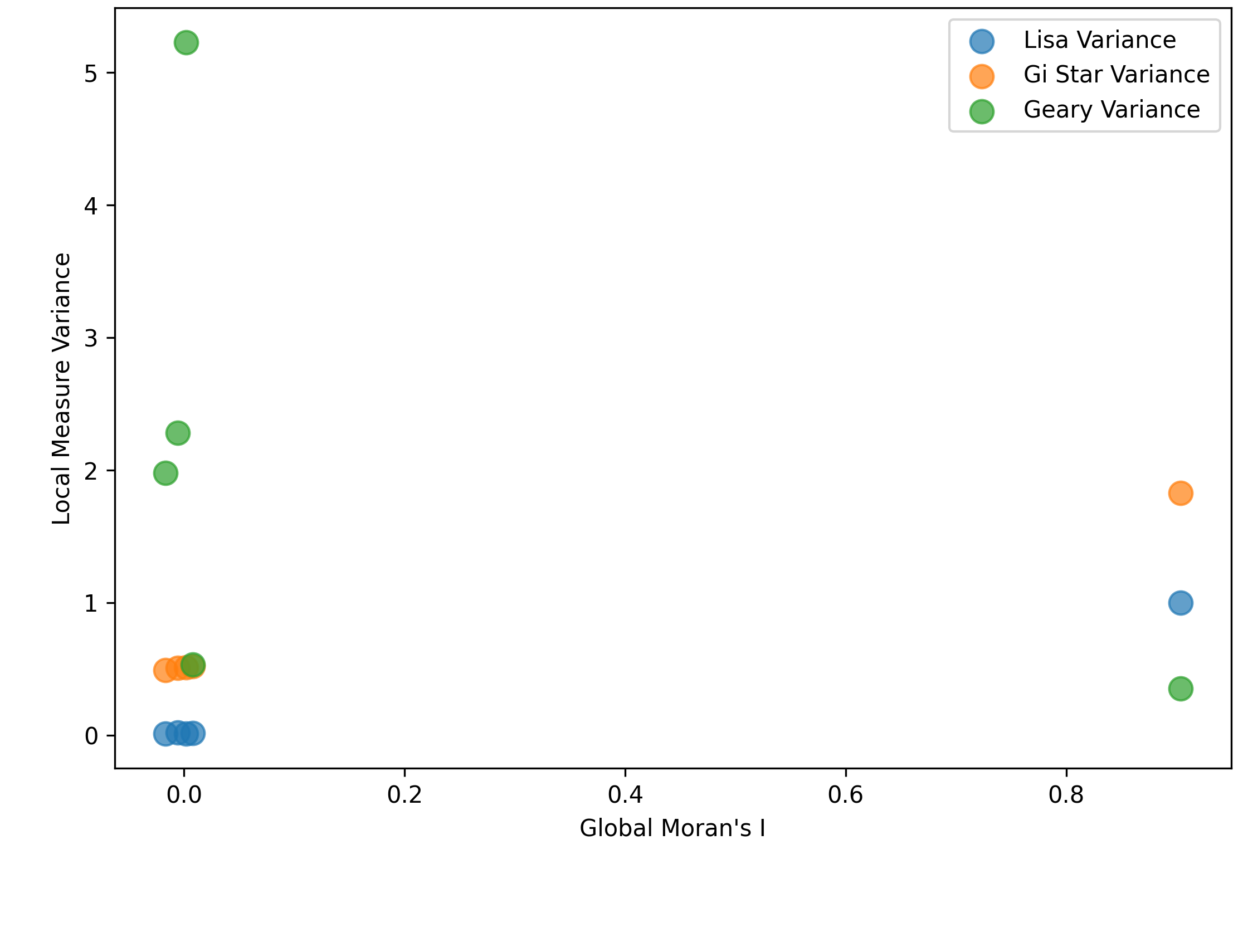}}
  \caption{Comparison of the three candidate local objectives (LISA variance, Getis-Ord Gi* variance, and Local Geary's C variance) across the five synthetic spatial patterns (random, clustered, dispersed, hotspots, gradual). The four panels report the variance values by pattern, the correlation of each statistic with Global Moran's I, the value distributions, and the discriminative power (coefficient of variation). LISA variance exhibits the widest spread and highest discriminative power, supporting its selection as the local objective in GIS-moGA.}
  \label{fig:alt_local_objectives_comparison}
\end{figure*}

Figure~\ref{fig:alt_local_objectives_comparison} shows the four panels from the local objective comparison.

The panels visualize the quantitative comparison in Tables~\ref{tab:alt_local_objectives_scenarios} and~\ref{tab:alt_local_objectives_summary}. LISA variance spreads across the widest range (0.0104--0.9974) over the five patterns, producing the high coefficient of variation (185.66\%) that gives it strong discriminative power. Gi* variance stays compressed in a narrow band (0.4888--1.8257) with most values near 0.5, yielding lower discriminative power (CV = 68.45\%) despite its near-perfect correlation with GMI. Local Geary's C variance responds inversely to clustering in several patterns (correlation $-0.4919$ with GMI), making it unsuitable as a companion objective to Global Moran's I.

\begin{figure*}[!t]
  \centering
  \subfloat[LISA.]{\includegraphics[width=.3\textwidth]{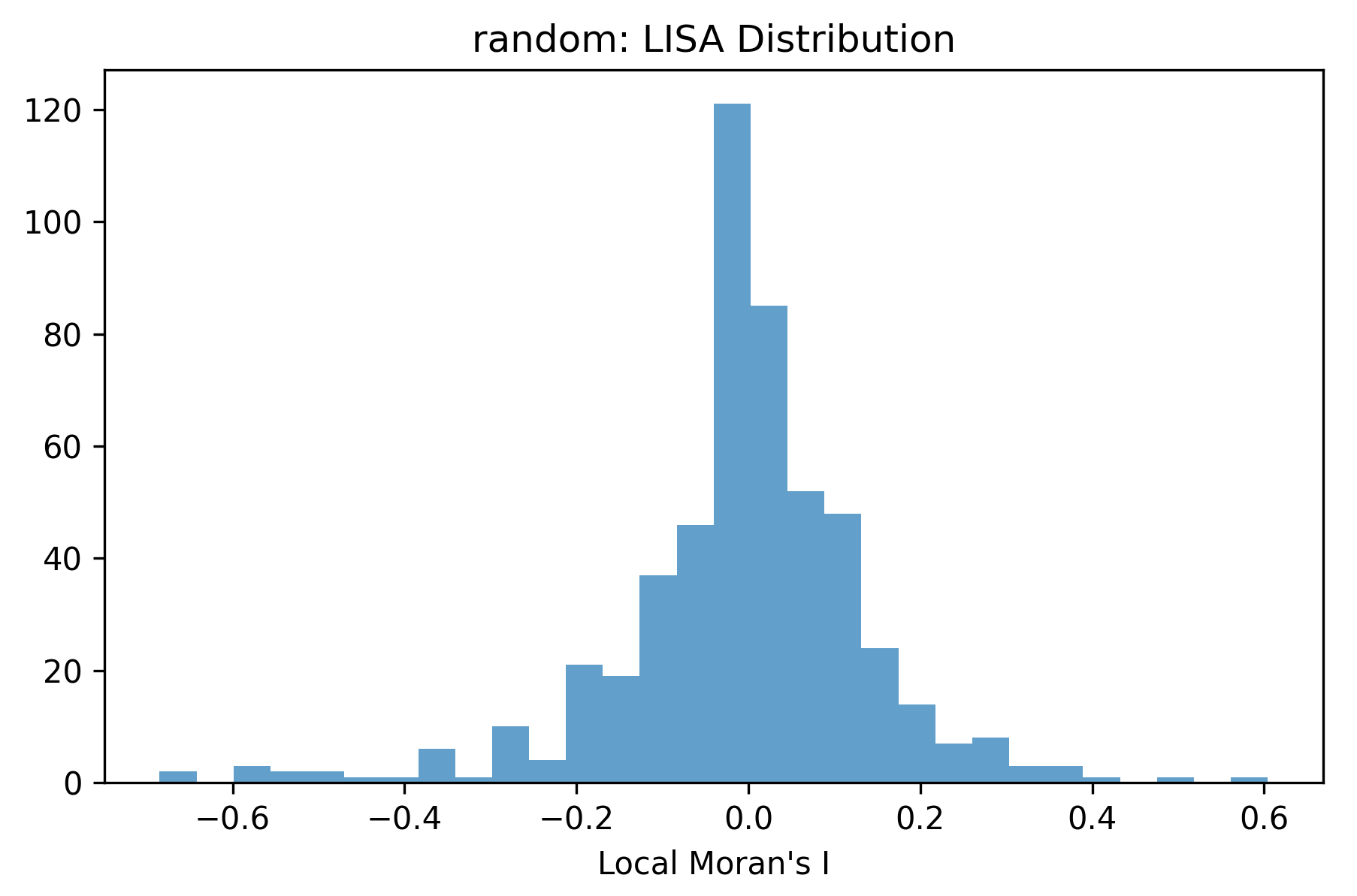}}
  \hfil
  \subfloat[Gi*.]{\includegraphics[width=.3\textwidth]{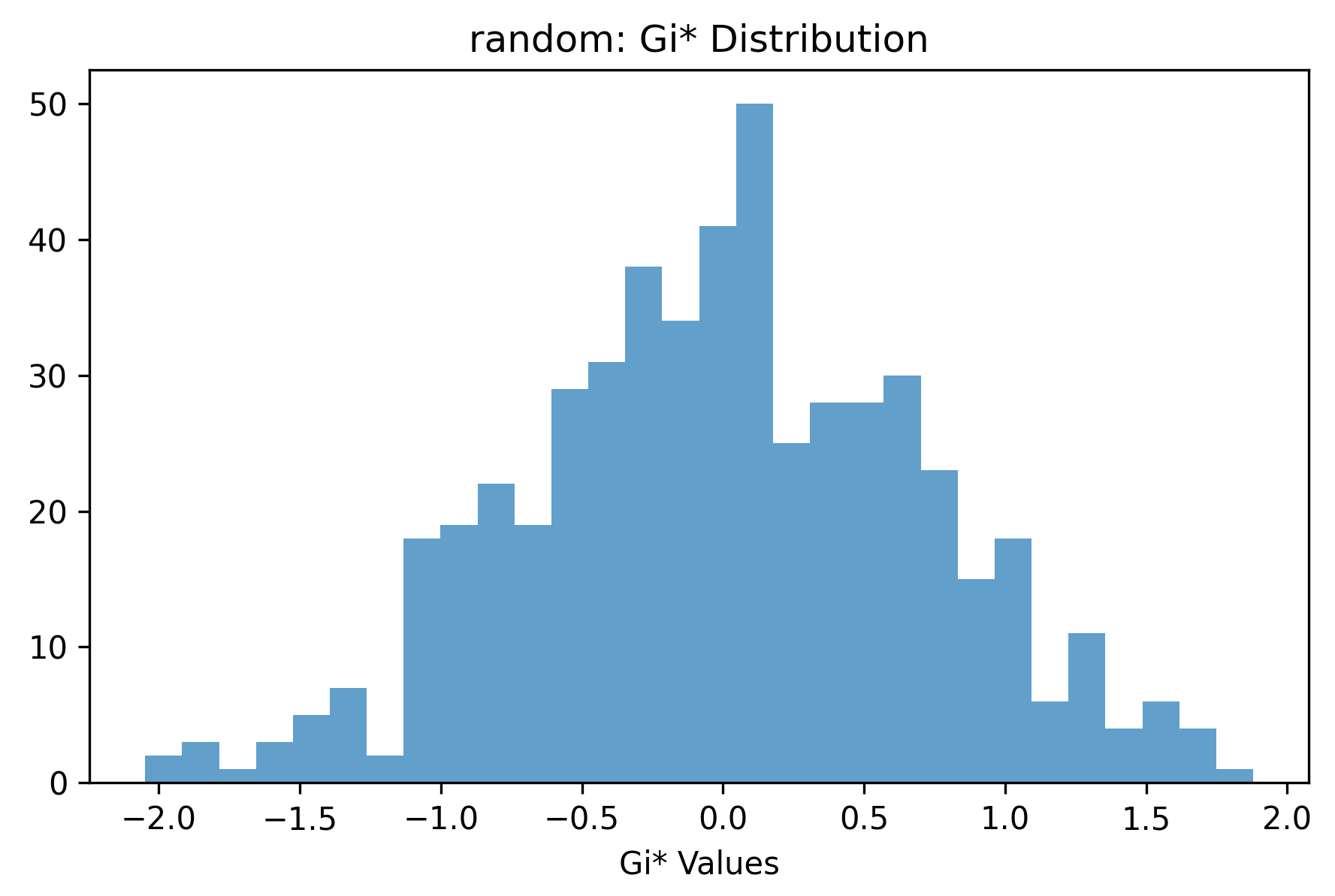}}
  \hfil
  \subfloat[Local Geary.]{\includegraphics[width=.3\textwidth]{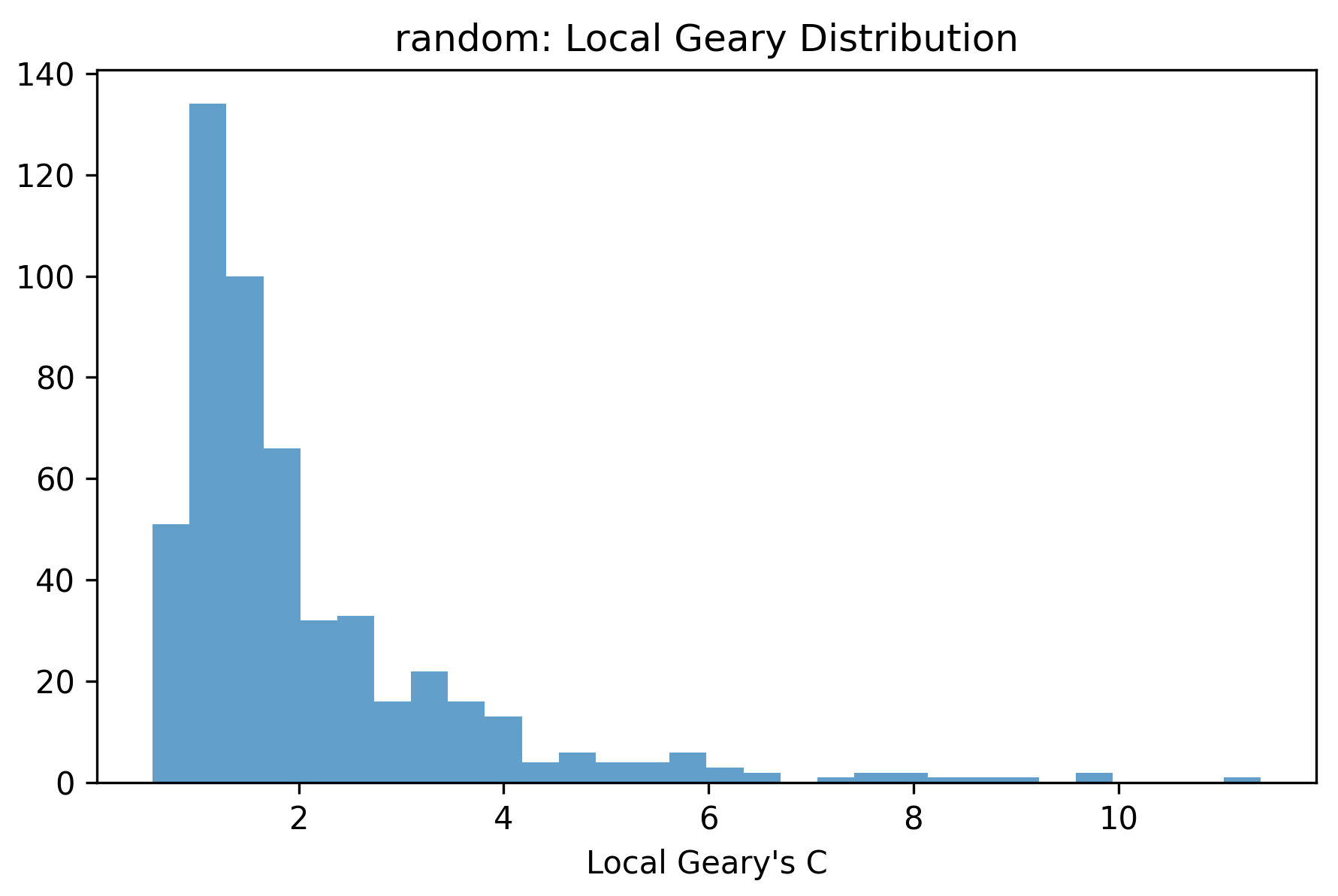}}
  \caption{Empirical distributions of the three local spatial association statistics---(a) LISA, (b) Getis-Ord Gi*, and (c) Local Geary's C---over census tracts of the random synthetic pattern. The LISA distribution is approximately symmetric and centered near zero, consistent with spatial randomness; Gi* remains tightly bounded; Local Geary's C is roughly uniform.}
  \label{fig:alt_random_dist}
\end{figure*}

\begin{figure*}[!t]
  \centering
  \subfloat[LISA.]{\includegraphics[width=.3\textwidth]{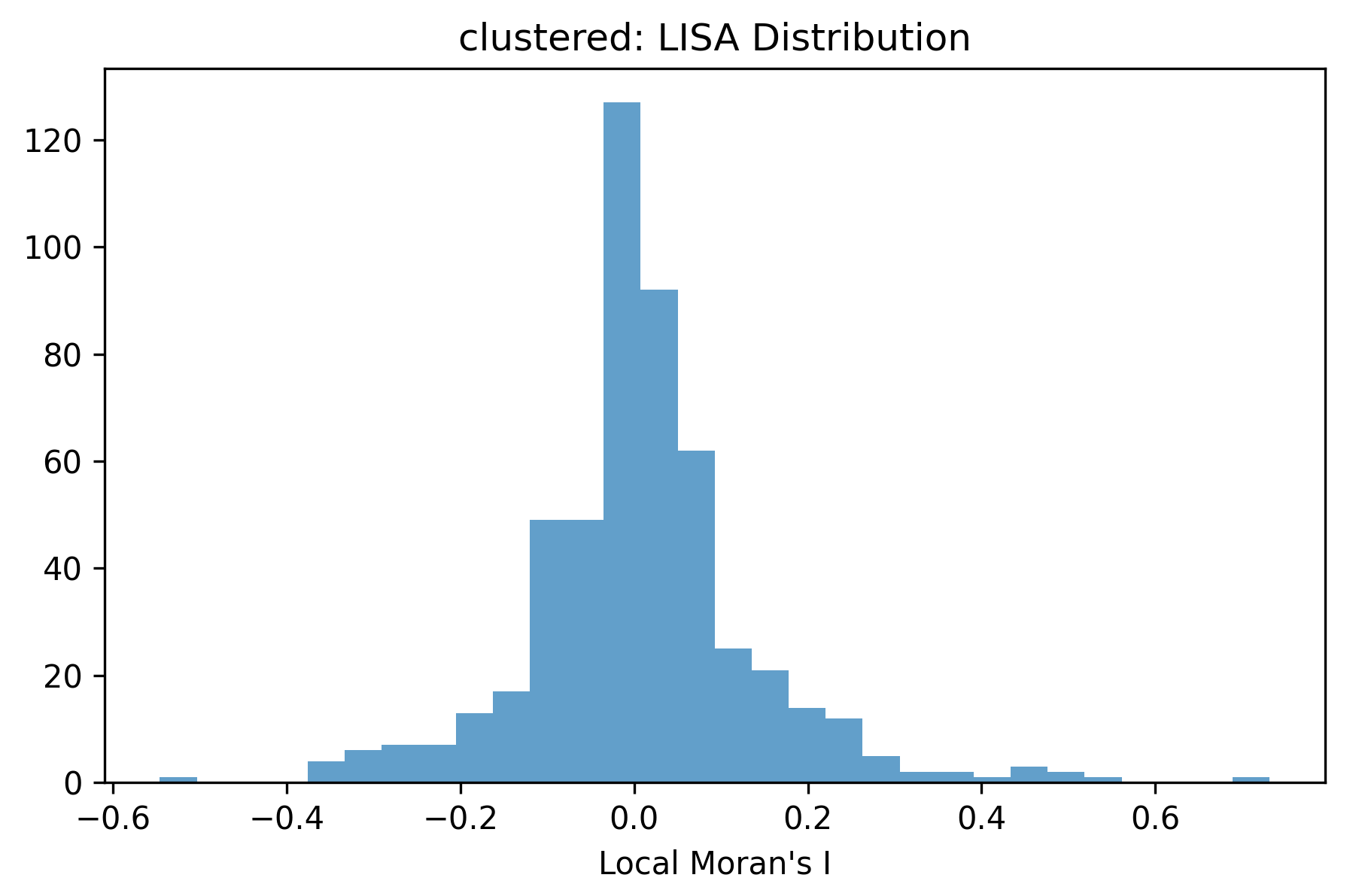}}
  \hfil
  \subfloat[Gi*.]{\includegraphics[width=.3\textwidth]{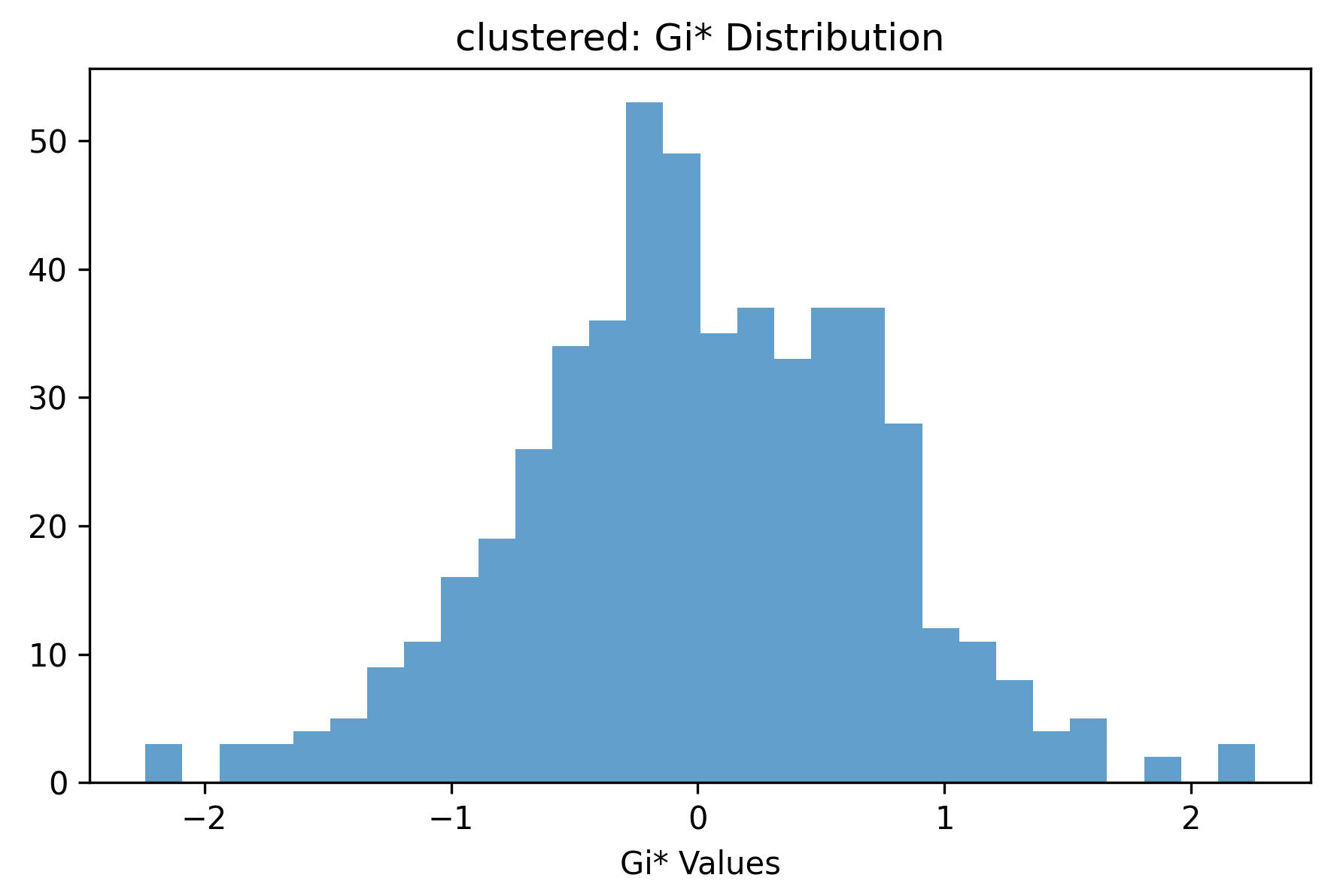}}
  \hfil
  \subfloat[Local Geary.]{\includegraphics[width=.3\textwidth]{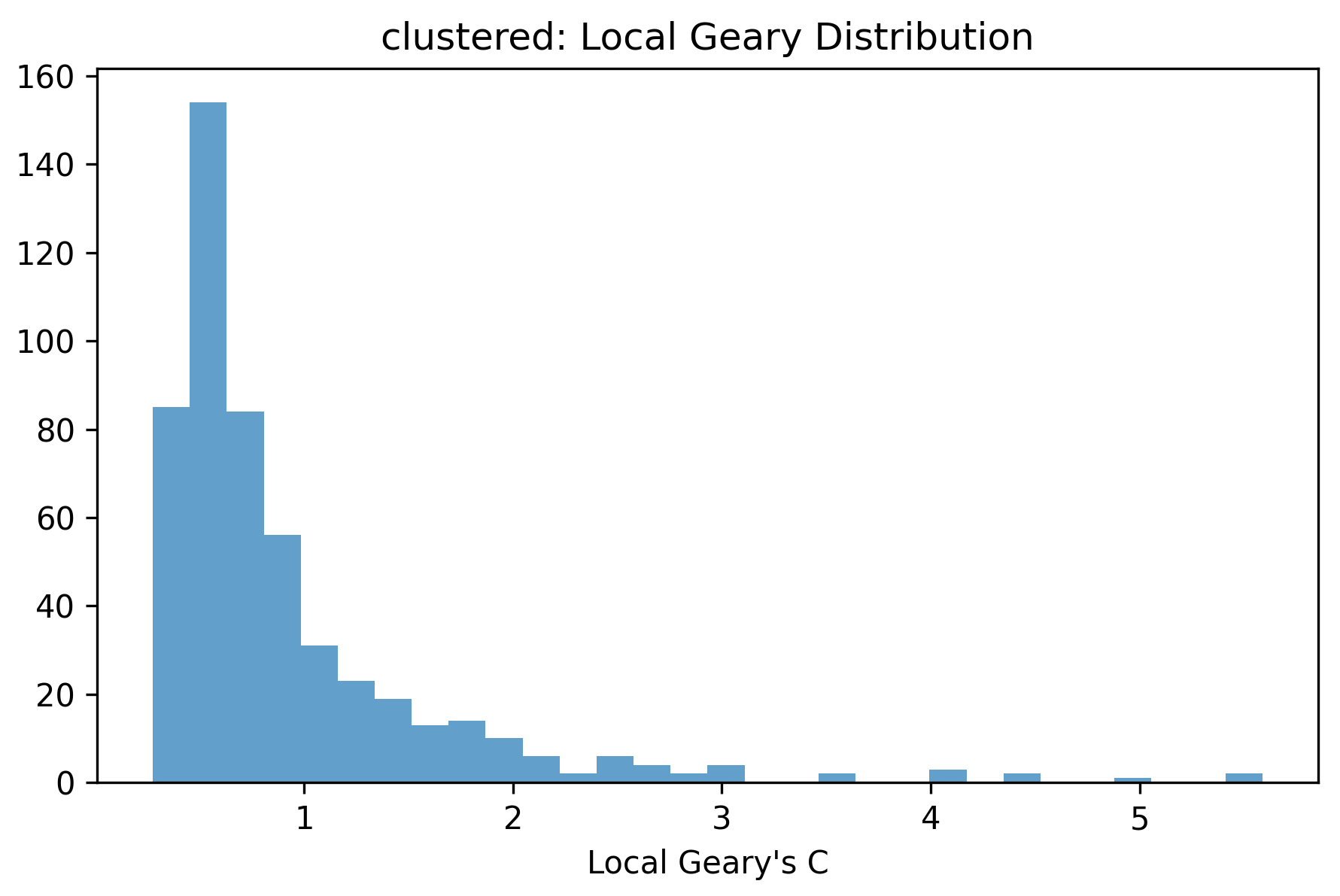}}
  \caption{Empirical distributions of the three local spatial association statistics---(a) LISA, (b) Getis-Ord Gi*, and (c) Local Geary's C---over census tracts of the clustered synthetic pattern. The LISA distribution exhibits heavy tails driven by the concentration of similar values; Gi* stays narrowly bounded; Local Geary's C shows a wider spread than under the random pattern.}
  \label{fig:alt_clustered_dist}
\end{figure*}

\begin{figure*}[!t]
  \centering
  \subfloat[LISA.]{\includegraphics[width=.3\textwidth]{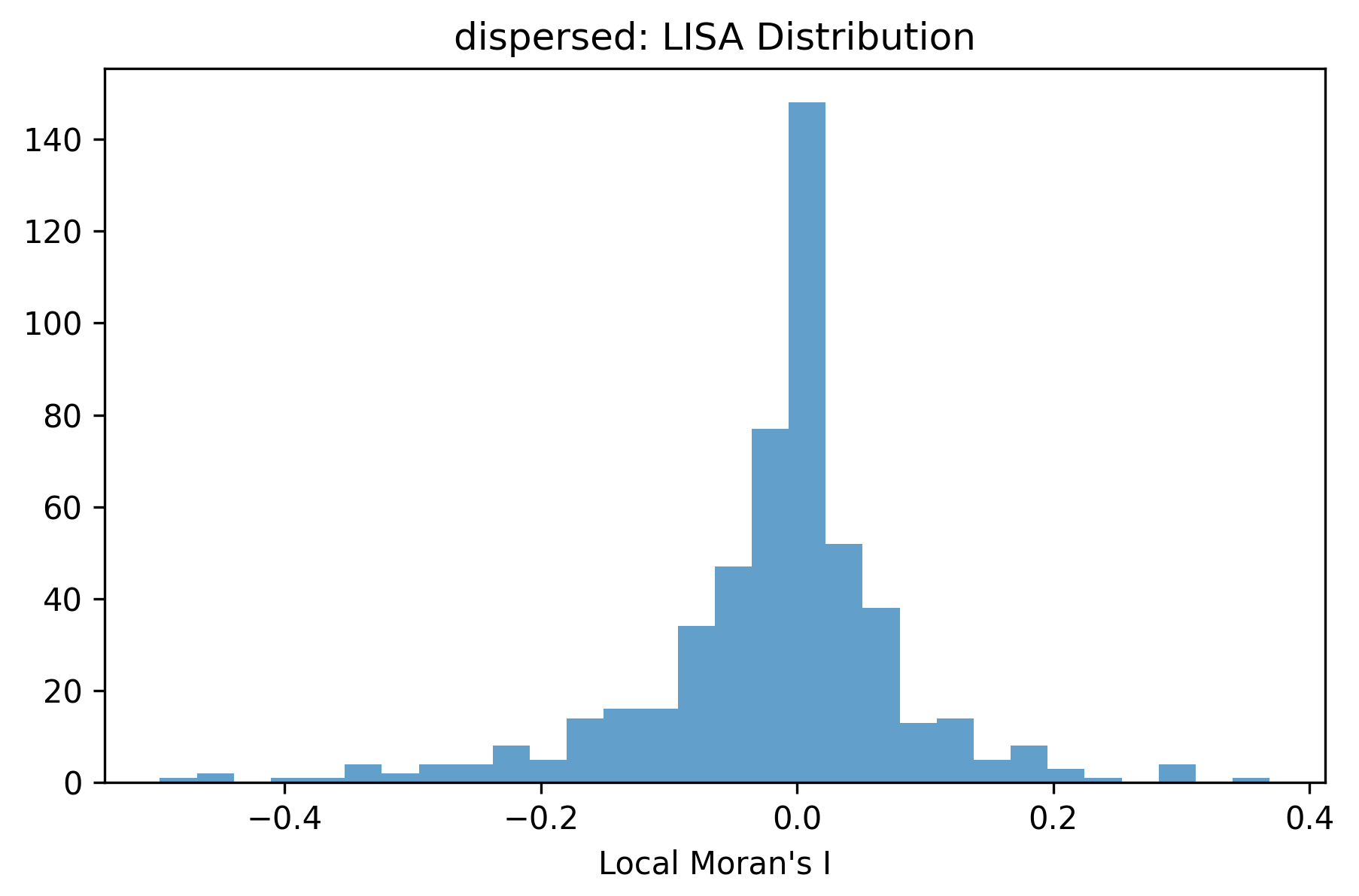}}
  \hfil
  \subfloat[Gi*.]{\includegraphics[width=.3\textwidth]{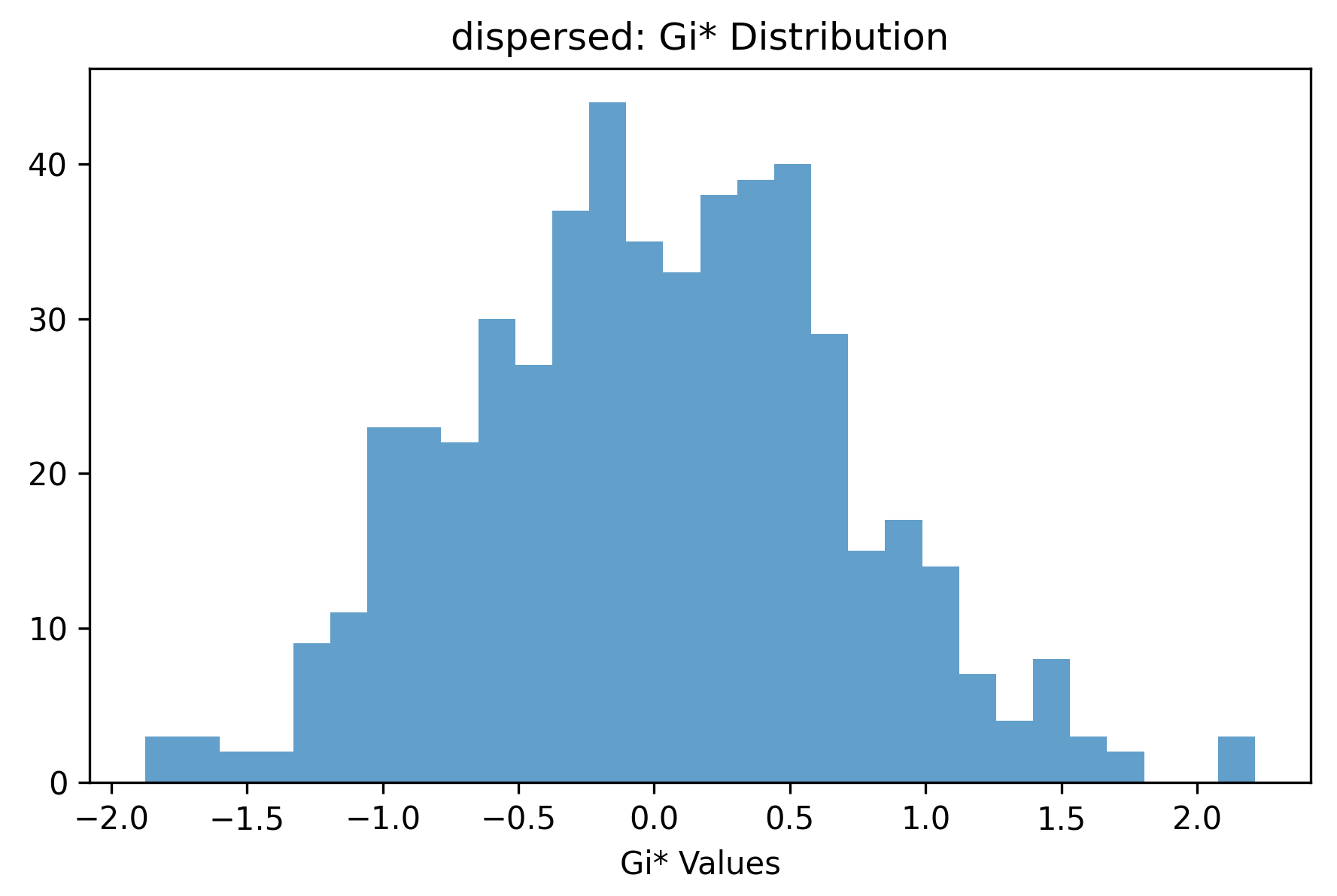}}
  \hfil
  \subfloat[Local Geary.]{\includegraphics[width=.3\textwidth]{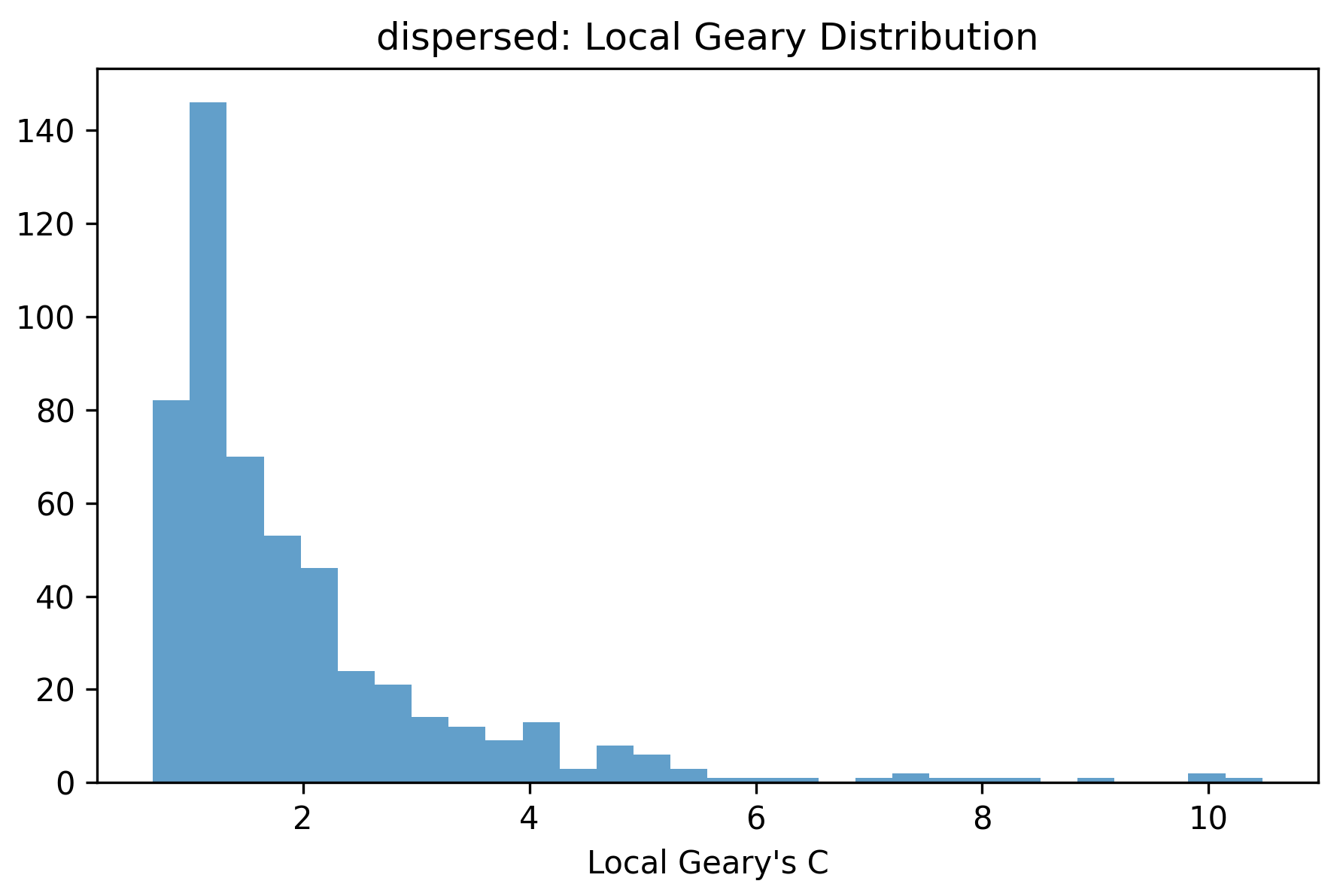}}
  \caption{Empirical distributions of the three local spatial association statistics---(a) LISA, (b) Getis-Ord Gi*, and (c) Local Geary's C---over census tracts of the dispersed synthetic pattern. The LISA distribution is narrower than under clustering, reflecting the checkerboard-like arrangement of values; Gi* remains compressed; Local Geary's C shifts toward higher values due to the alternating high--low juxtaposition.}
  \label{fig:alt_dispersed_dist}
\end{figure*}

\begin{figure*}[!t]
  \centering
  \subfloat[LISA.]{\includegraphics[width=.3\textwidth]{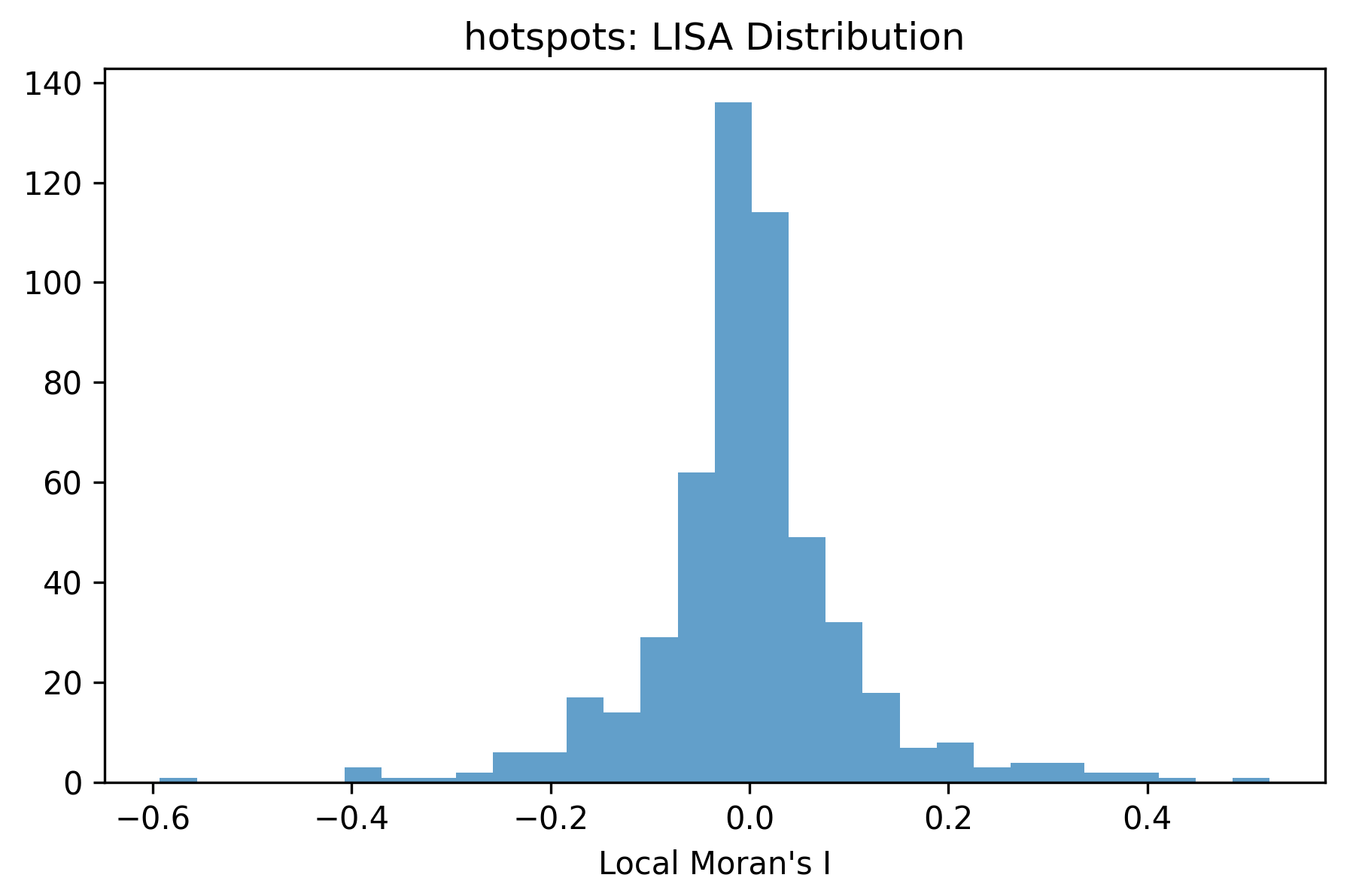}}
  \hfil
  \subfloat[Gi*.]{\includegraphics[width=.3\textwidth]{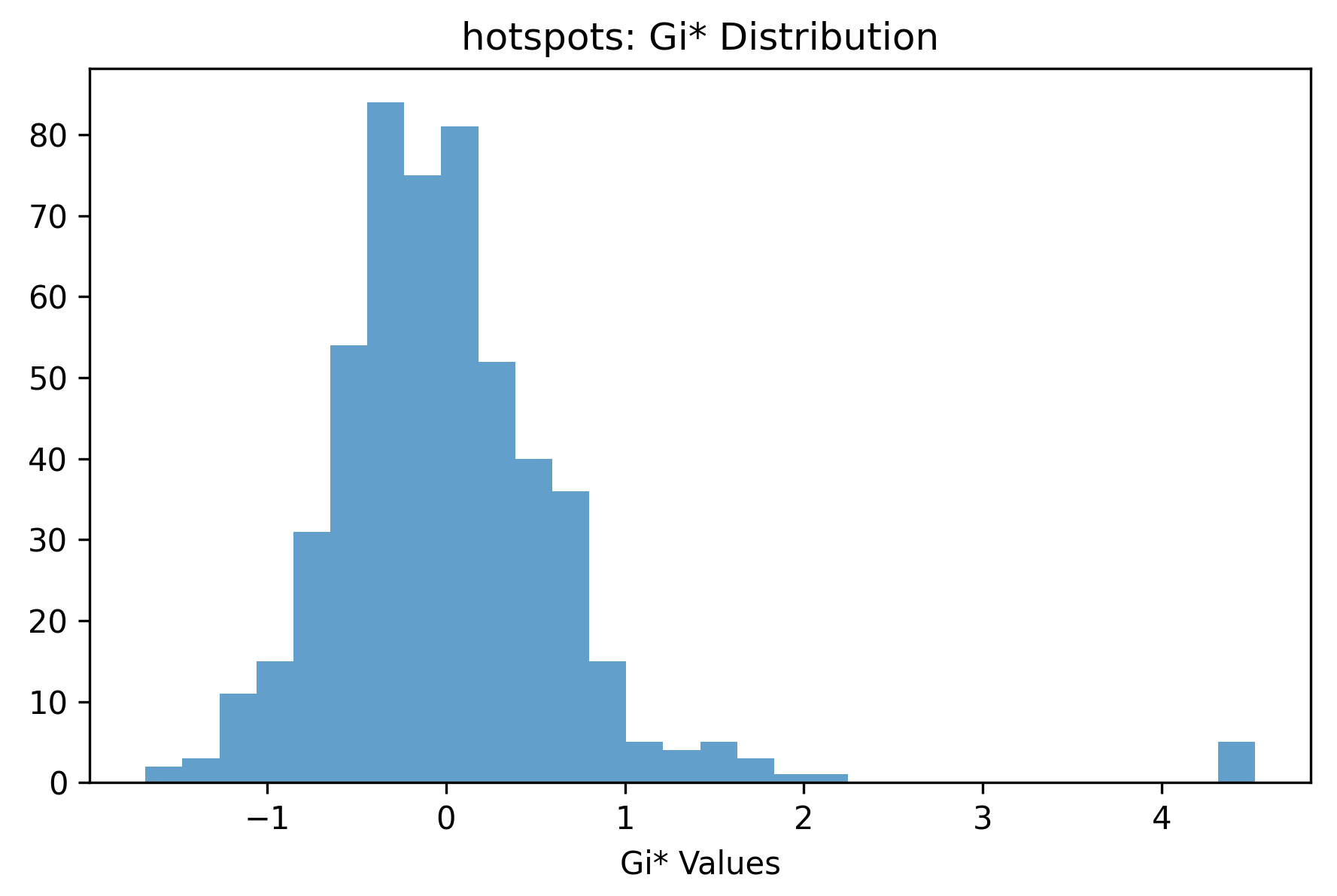}}
  \hfil
  \subfloat[Local Geary.]{\includegraphics[width=.3\textwidth]{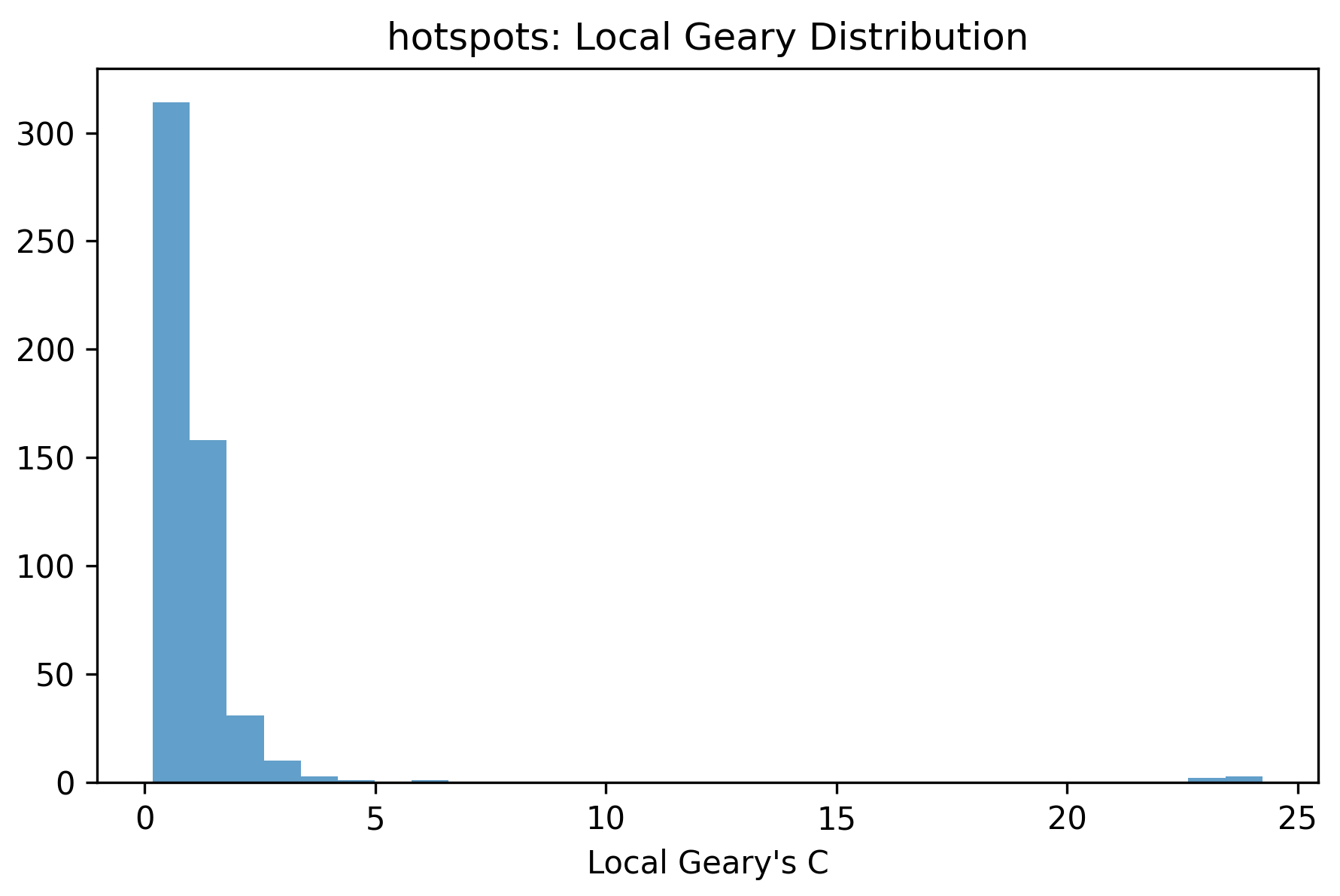}}
  \caption{Empirical distributions of the three local spatial association statistics---(a) LISA, (b) Getis-Ord Gi*, and (c) Local Geary's C---over census tracts of the hotspots synthetic pattern. The LISA distribution shows the heaviest positive tail, corresponding to the localized high-value concentrations; Gi* stays narrowly bounded; Local Geary's C produces its widest spread of all five patterns.}
  \label{fig:alt_hotspots_dist}
\end{figure*}

\begin{figure*}[!t]
  \centering
  \subfloat[LISA.]{\includegraphics[width=.3\textwidth]{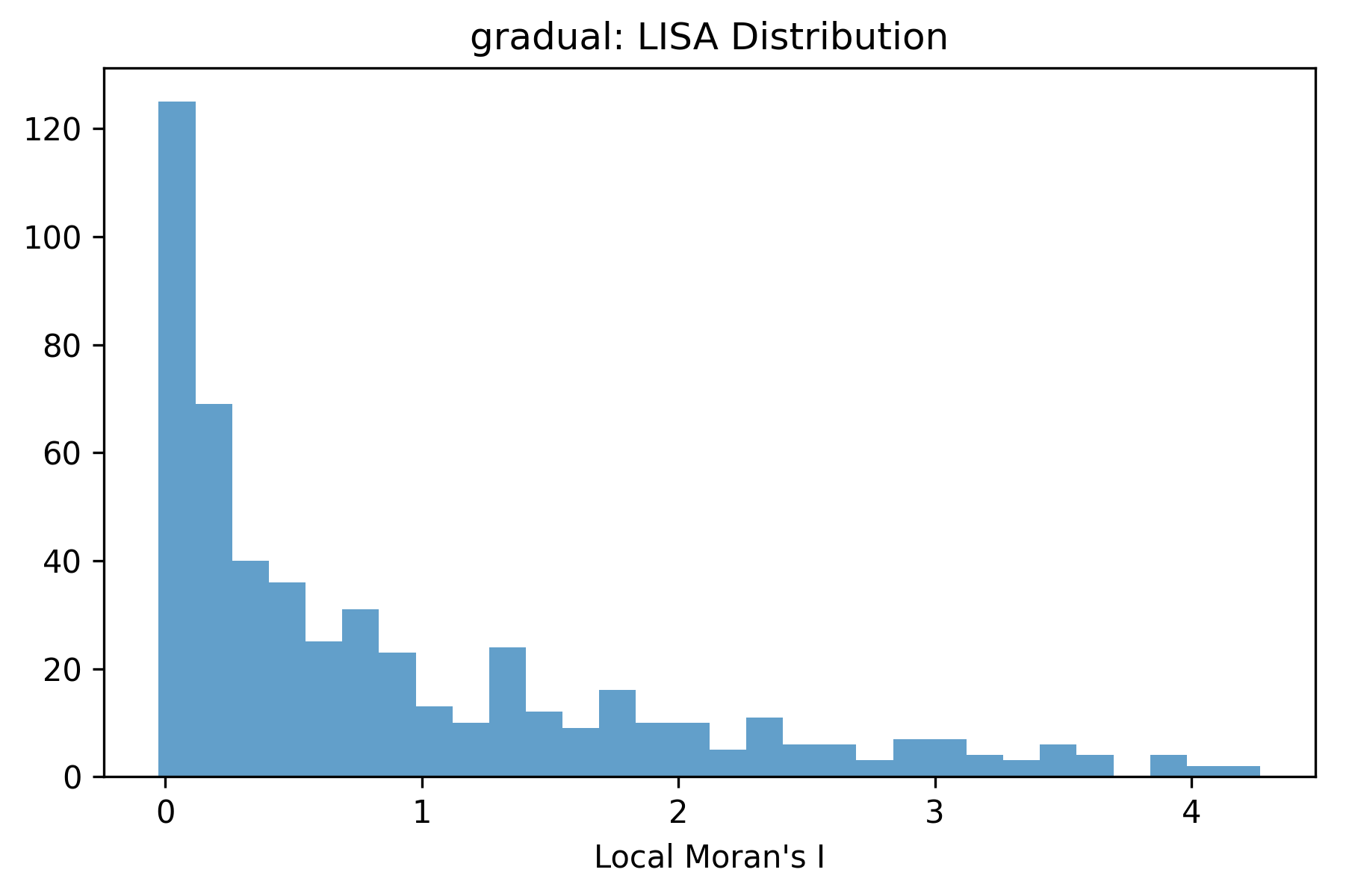}}
  \hfil
  \subfloat[Gi*.]{\includegraphics[width=.3\textwidth]{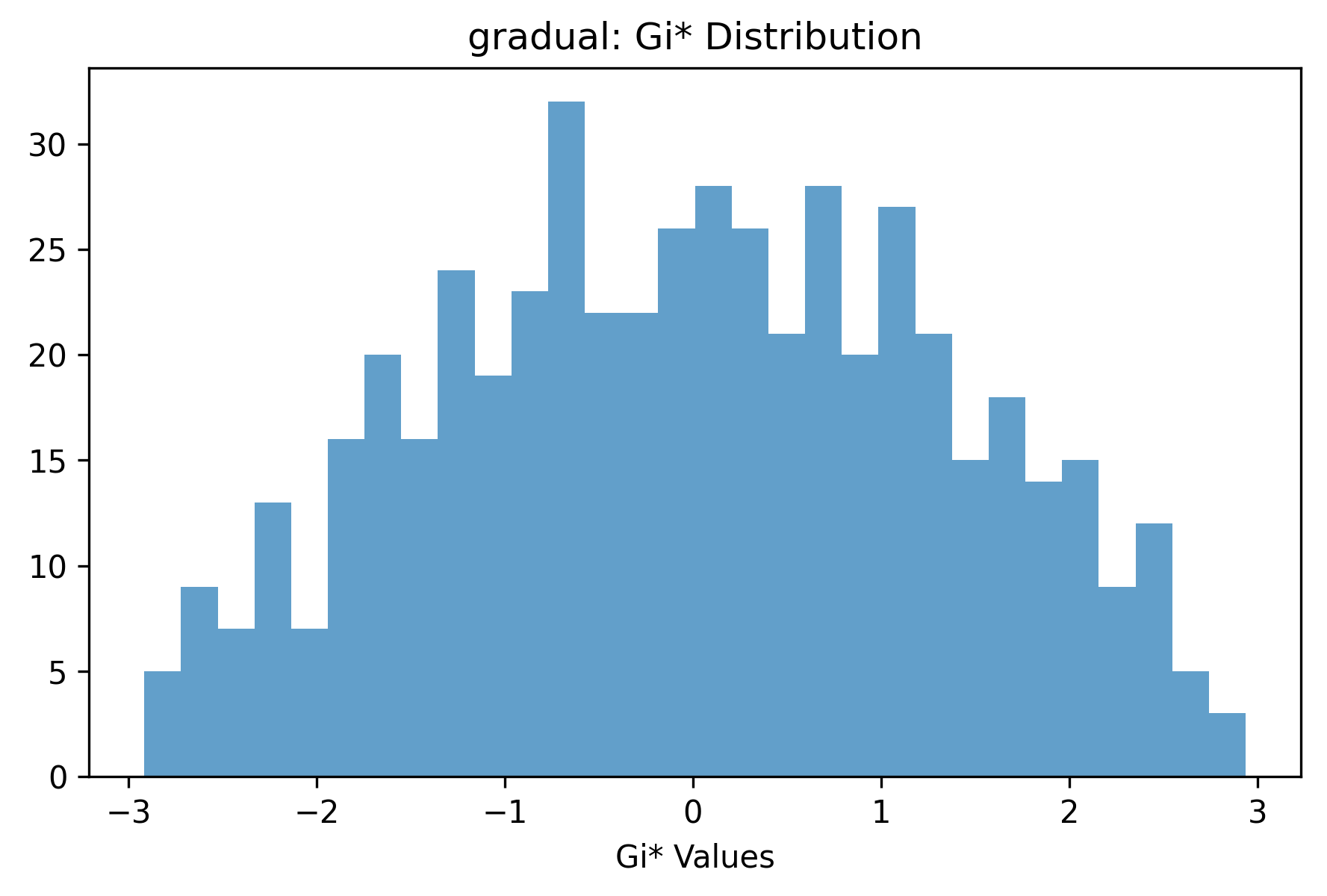}}
  \hfil
  \subfloat[Local Geary.]{\includegraphics[width=.3\textwidth]{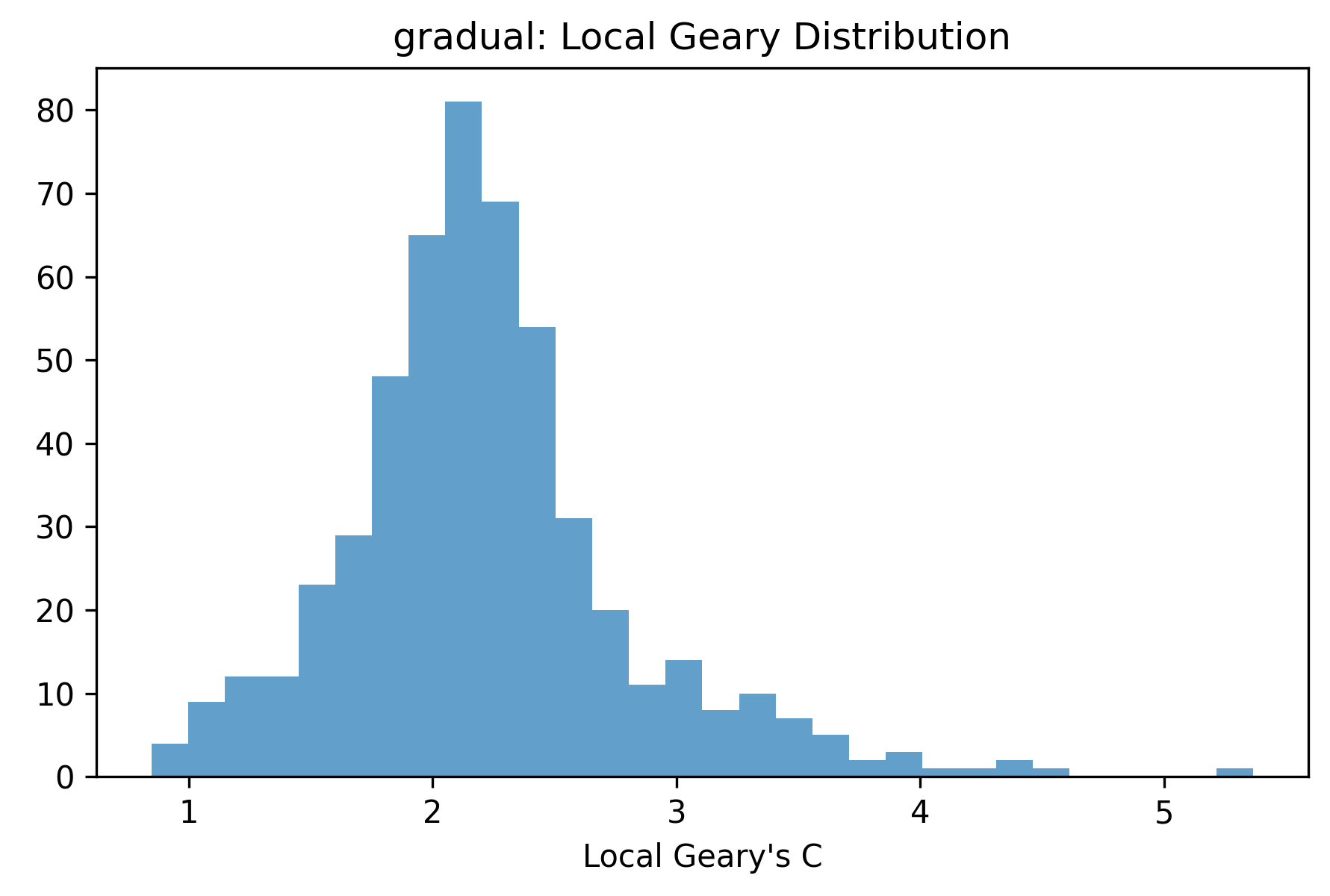}}
  \caption{Empirical distributions of the three local spatial association statistics---(a) LISA, (b) Getis-Ord Gi*, and (c) Local Geary's C---over census tracts of the gradual-trend synthetic pattern. The smooth gradient produces the widest LISA range of all patterns (variance 0.9974) alongside strong global autocorrelation (GMI = 0.9038), whereas Local Geary's C variance falls to 0.3508, reflecting its inverse sensitivity to smooth gradients.}
  \label{fig:alt_gradual_dist}
\end{figure*}

Figures~\ref{fig:alt_random_dist}, \ref{fig:alt_clustered_dist}, \ref{fig:alt_dispersed_dist}, \ref{fig:alt_hotspots_dist}, and \ref{fig:alt_gradual_dist} show the distributional profiles for the five canonical spatial patterns.

The distributional profiles show how each local statistic disperses under the five canonical patterns. Under the random pattern (Figure~\ref{fig:alt_random_dist}), all three indicators produce near-symmetric distributions centered near zero, as expected under spatial randomness. The clustered and hotspot patterns (Figures~\ref{fig:alt_clustered_dist} and~\ref{fig:alt_hotspots_dist}) produce the heaviest LISA tails, which drive the wide variance range recorded in Table~\ref{tab:alt_local_objectives_scenarios}. The gradual pattern (Figure~\ref{fig:alt_gradual_dist}) produces the largest separation between indicators: LISA variance reaches 0.9974 alongside the strong global autocorrelation (GMI = 0.9038), whereas Local Geary's C variance falls to 0.3508, reflecting its inverse sensitivity to smooth gradients. Gi* distributions remain narrowly bounded across all five patterns, consistent with its low coefficient of variation.

\setcounter{figure}{0}
\setcounter{table}{0}
\section{Expanded Convergence and Sensitivity Diagnostics}

\subsection{Hypervolume Evolution Summary (All Scenarios)}

\begin{figure*}[!t]
  \centering
  \subfloat[By mutation rate.]{\includegraphics[width=.45\textwidth]{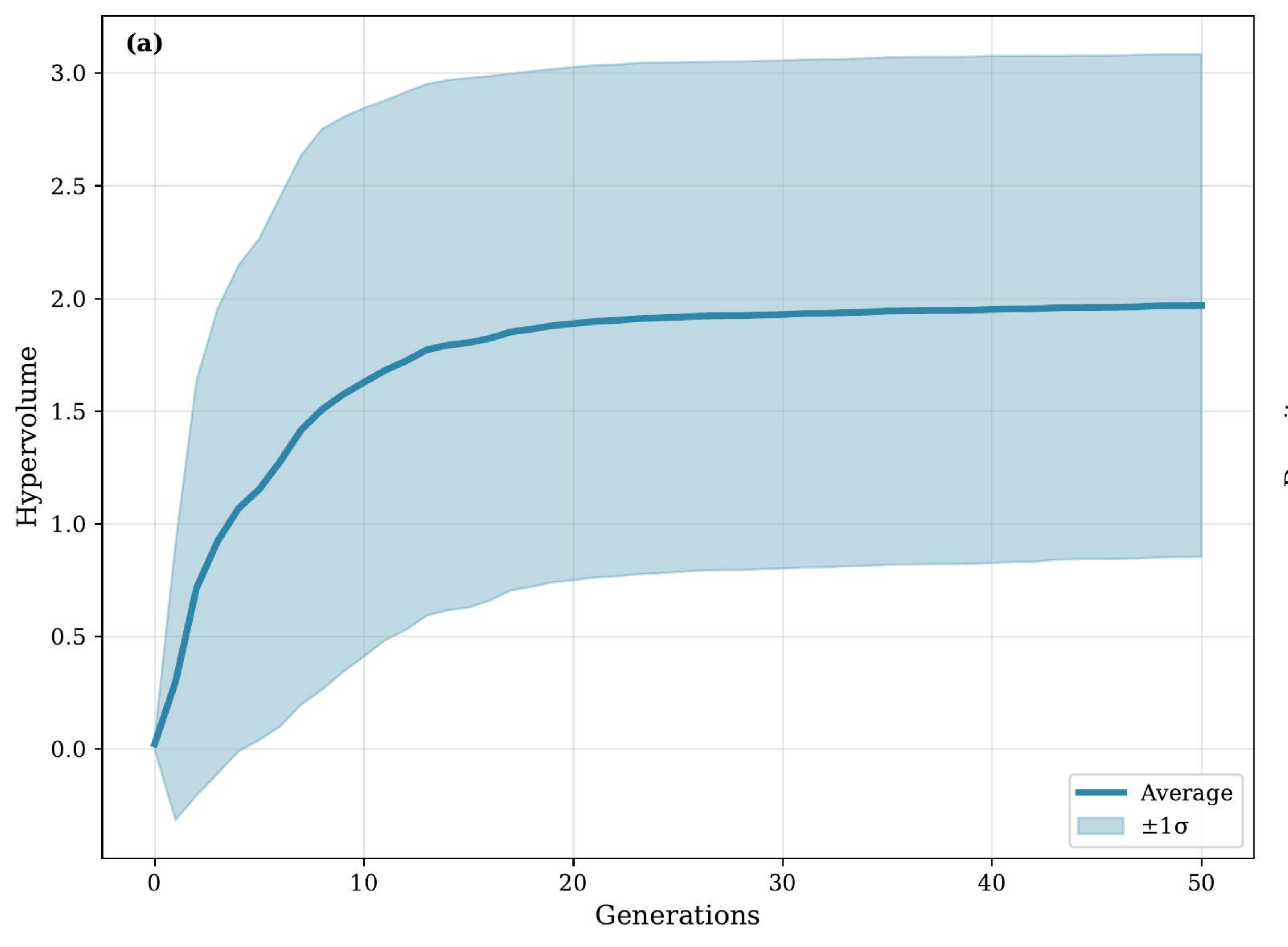}}
  \hfil
  \subfloat[By population size.]{\includegraphics[width=.45\textwidth]{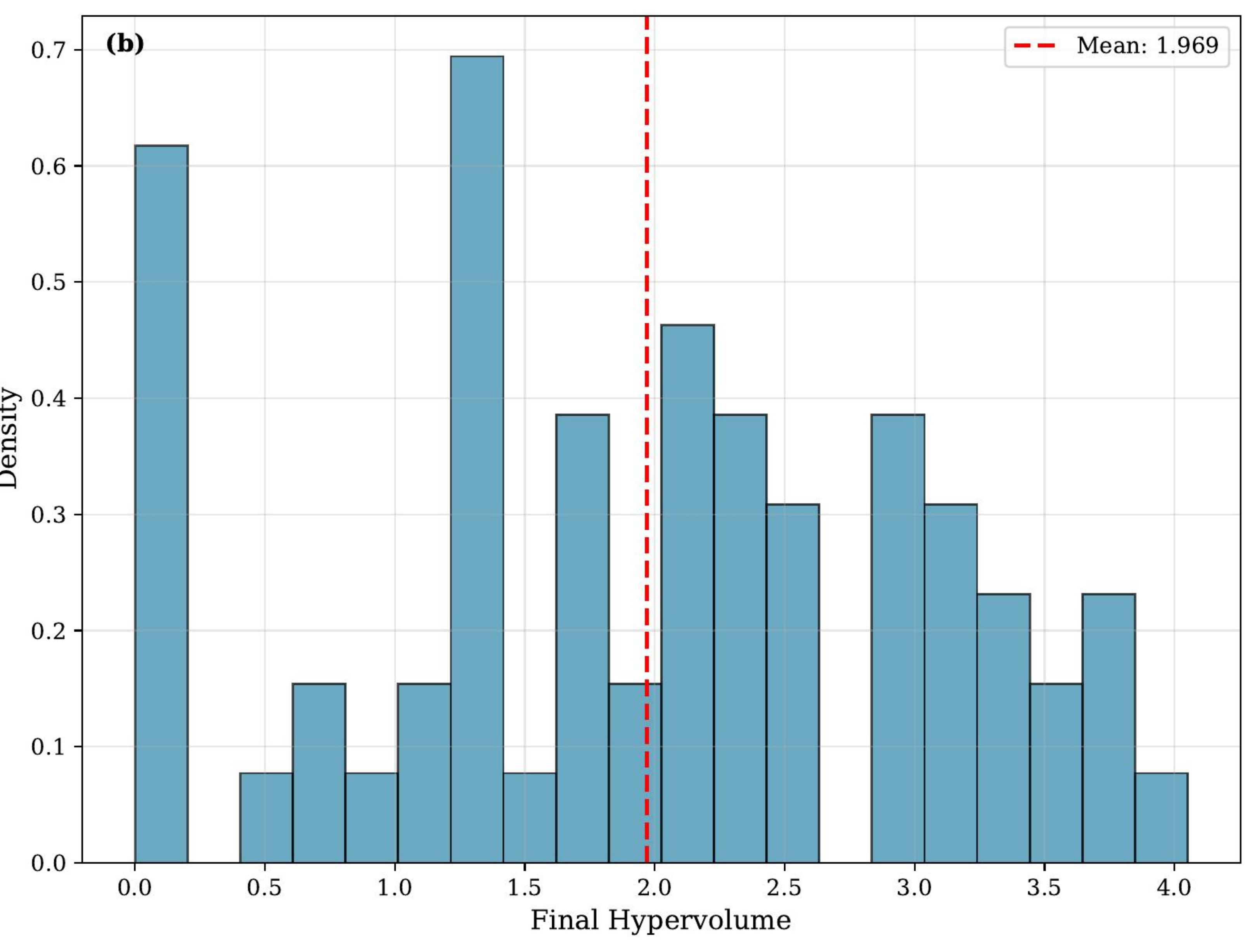}}

  \vspace{-4pt}

  \subfloat[By crossover alpha.]{\includegraphics[width=.45\textwidth]{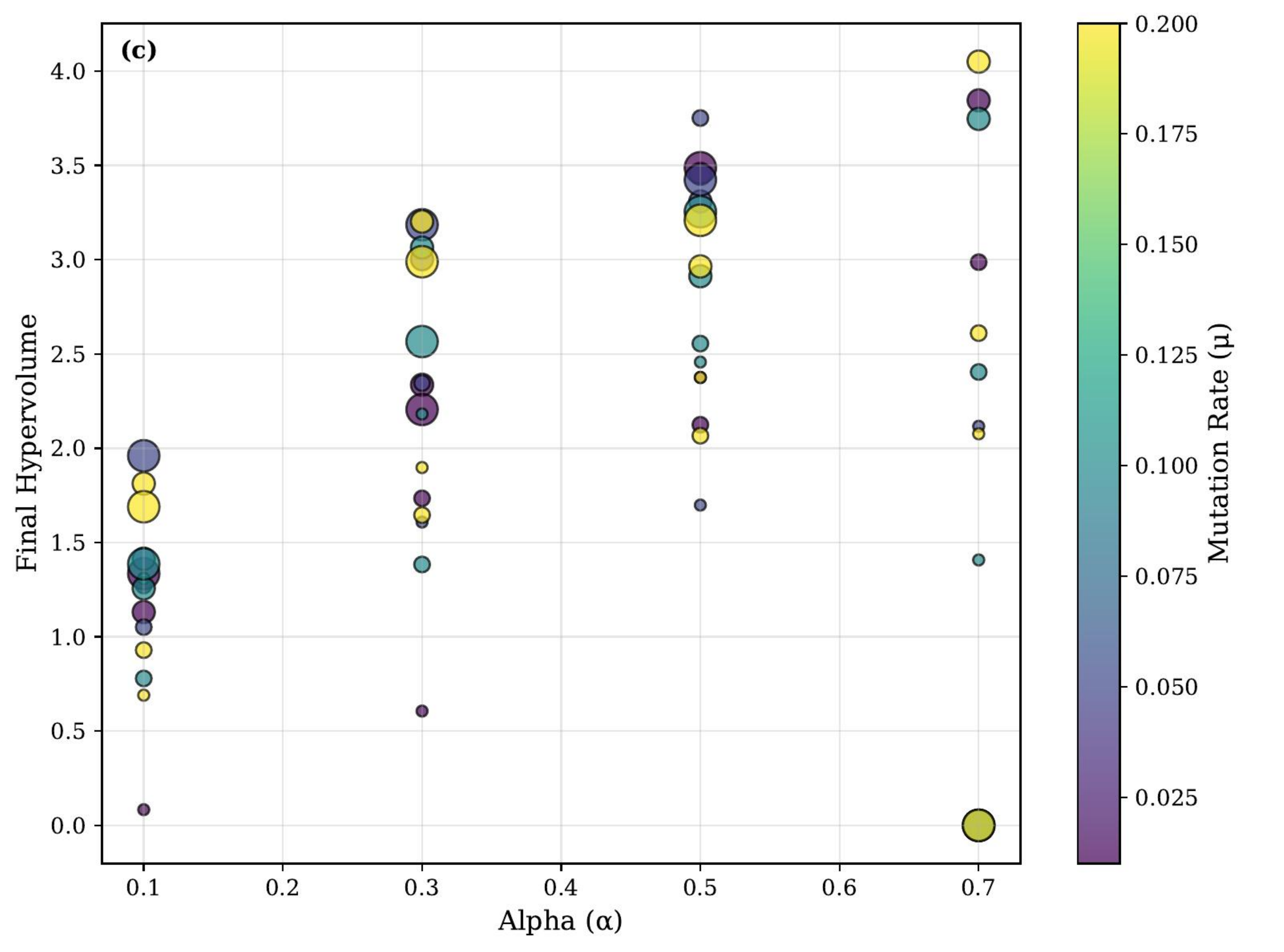}}
  \hfil
  \subfloat[By scenario rank.]{\includegraphics[width=.45\textwidth]{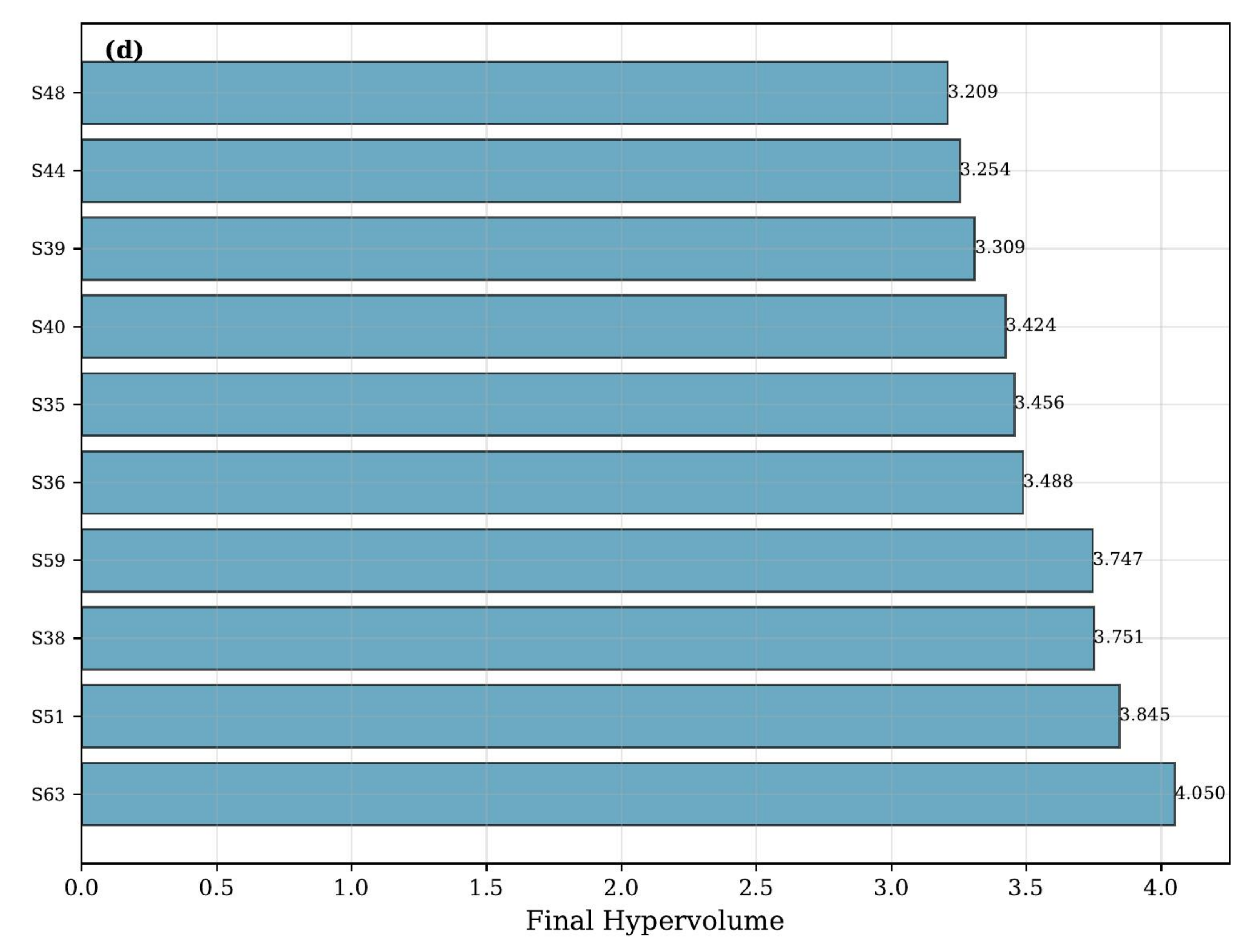}}
  \caption{Hypervolume evolution over 50 generations for all 64 scenarios, disaggregated by parameter: (a) by mutation rate, (b) by population size, (c) by crossover $\alpha$, and (d) by scenario rank. Configurations with mutation rates $\geq 0.05$ climb steadily, whereas the low-mutation (0.01) group plateaus early; the trace of Scenario 64 (high $\alpha$ and high mutation) remains flat at zero throughout.}
  \label{fig:hypervolume_evolution_summary}
\end{figure*}

Figure~\ref{fig:hypervolume_evolution_summary} presents the four panels from the hypervolume convergence summary.

The panels trace hypervolume over the 50 generations, disaggregated by mutation rate, population size, $\alpha$, and scenario rank. Configurations with mutation rates $\geq 0.05$ climb steadily through the full budget, whereas the low-mutation (0.01) group plateaus early. Population size produces a secondary effect, with the 400-individual group reaching the highest plateau. The flat zero trace of Scenario 64 is visible across the panels, isolating the disruptive effect of combining high $\alpha$ (0.7) with high mutation (0.20).

\subsection{Parameter Sensitivity Summary Views}

\begin{figure*}[!t]
  \centering
  \subfloat[Hypervolume vs mutation.]{\includegraphics[width=.3\textwidth]{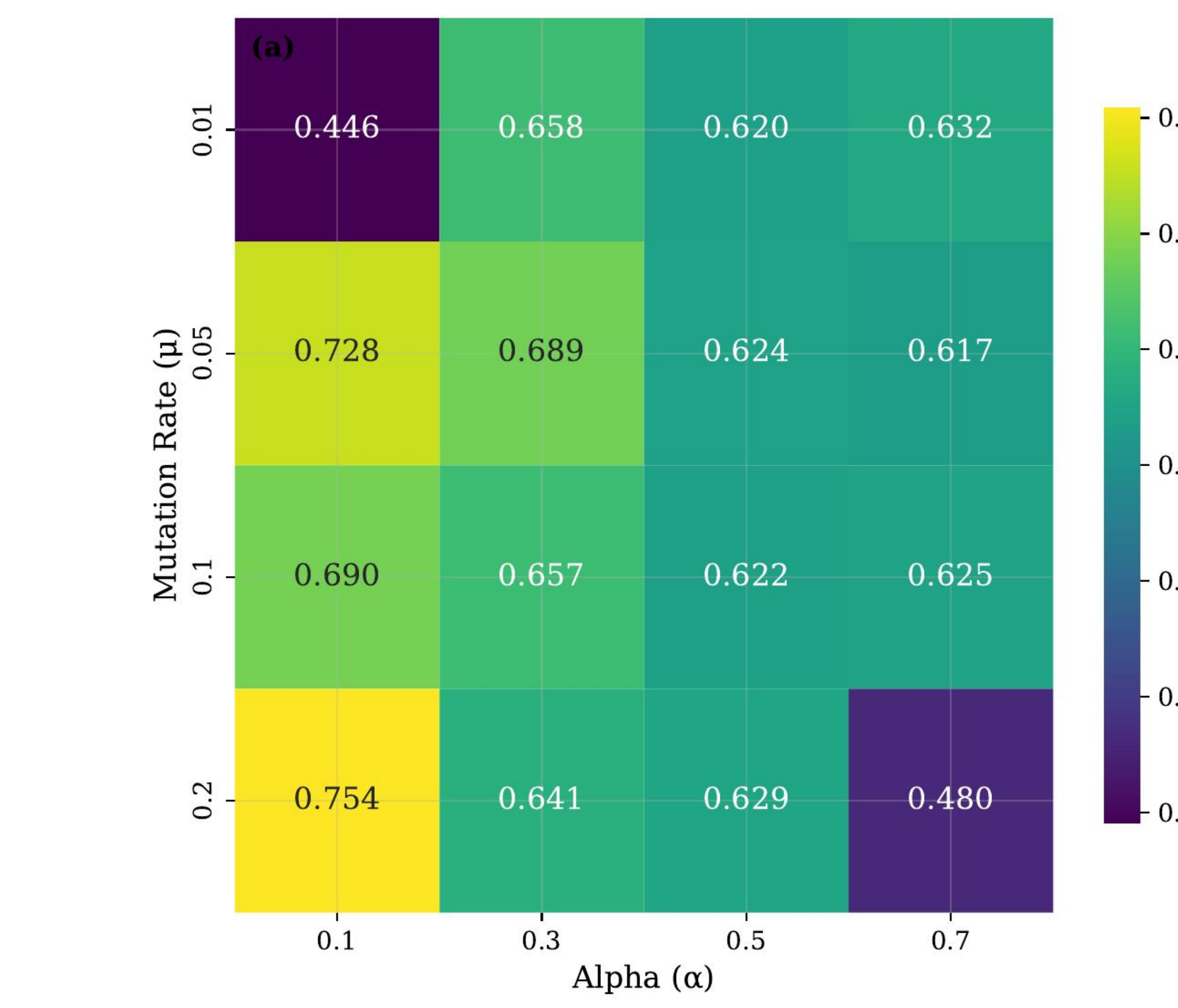}}
  \hfil
  \subfloat[Hypervolume vs pop.]{\includegraphics[width=.3\textwidth]{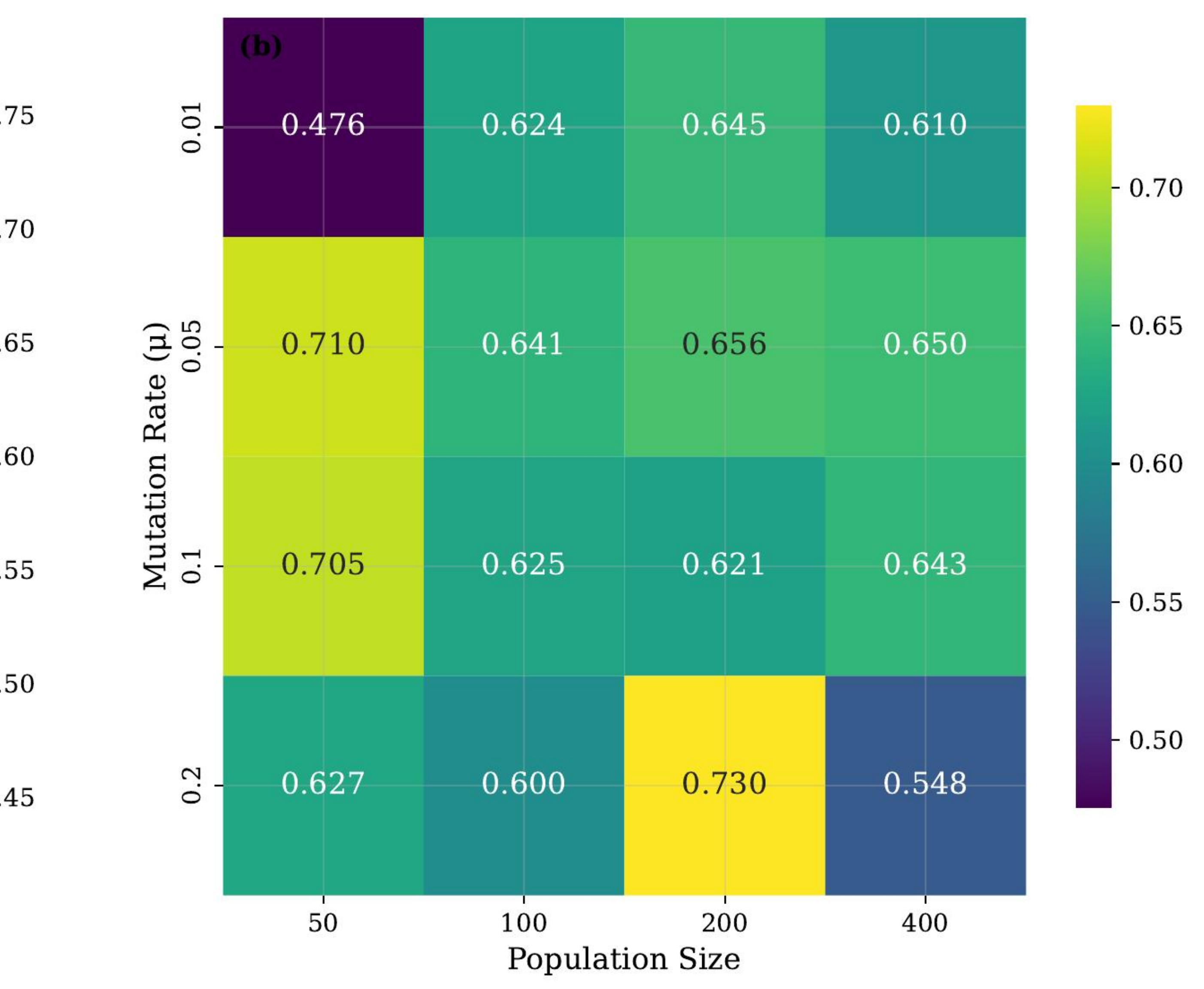}}
  \hfil
  \subfloat[Hypervolume vs alpha.]{\includegraphics[width=.3\textwidth]{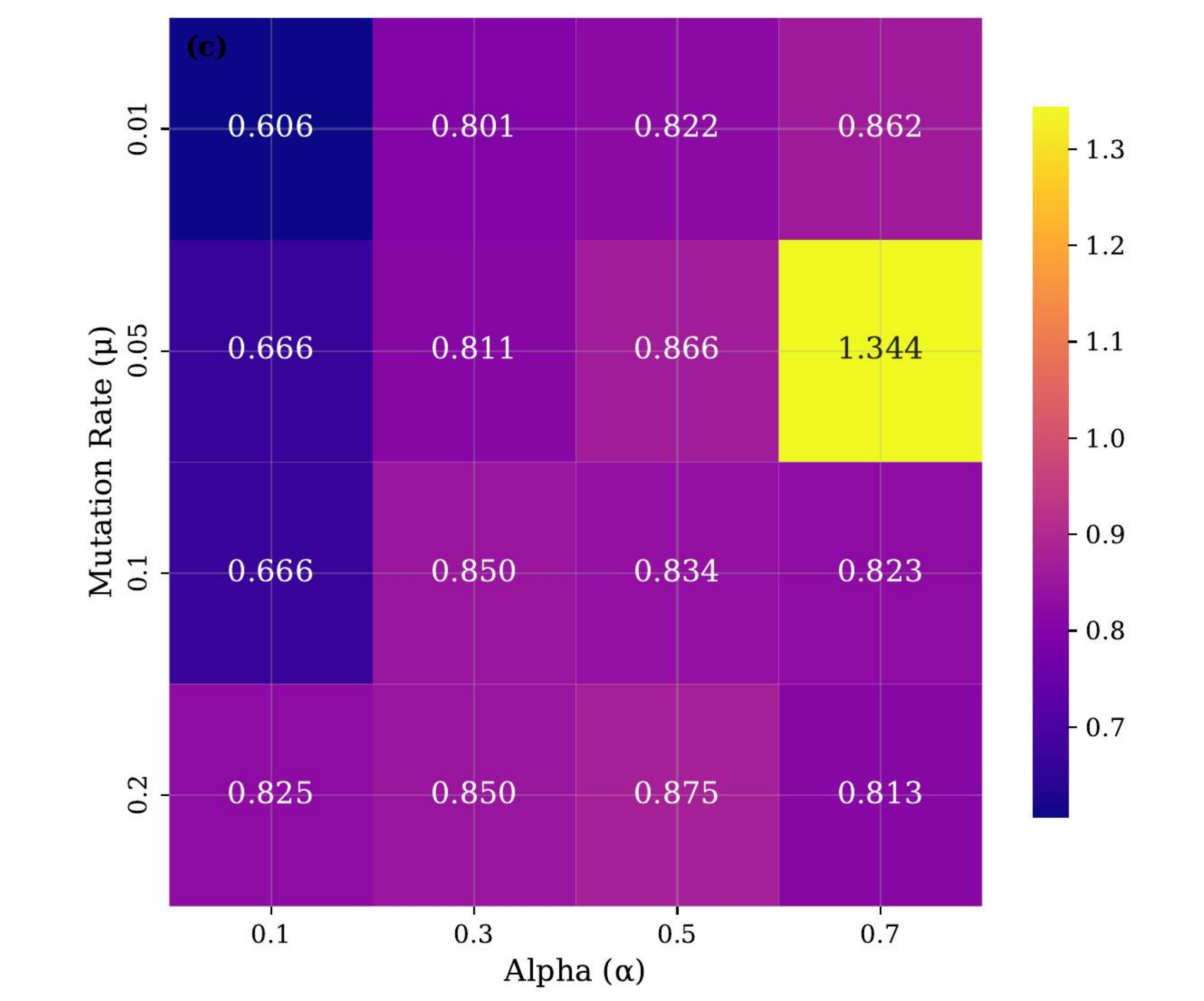}}

  \vspace{-4pt}

  \subfloat[Spread vs mutation.]{\includegraphics[width=.3\textwidth]{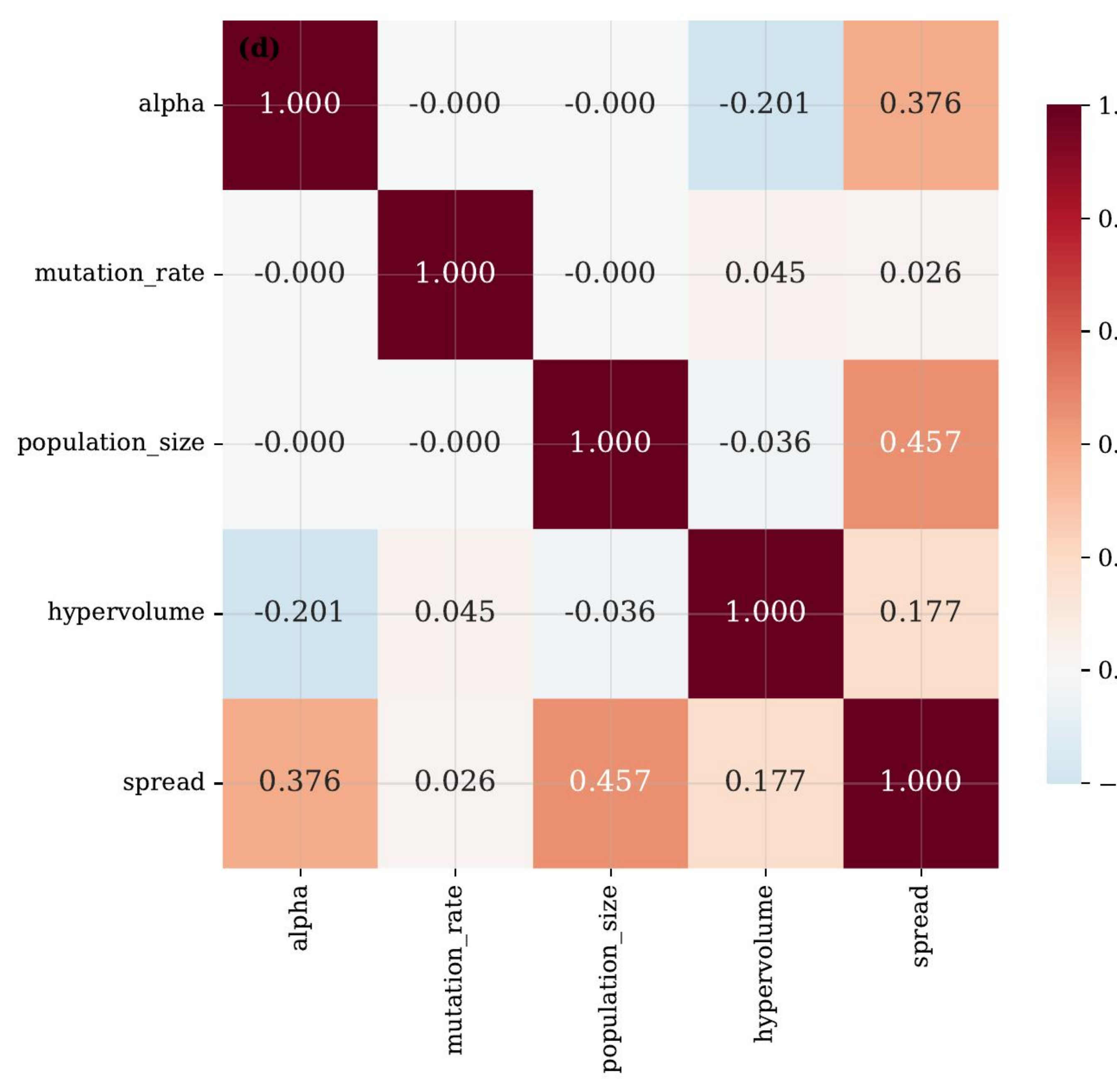}}
  \hfil
  \subfloat[Spread vs pop.]{\includegraphics[width=.3\textwidth]{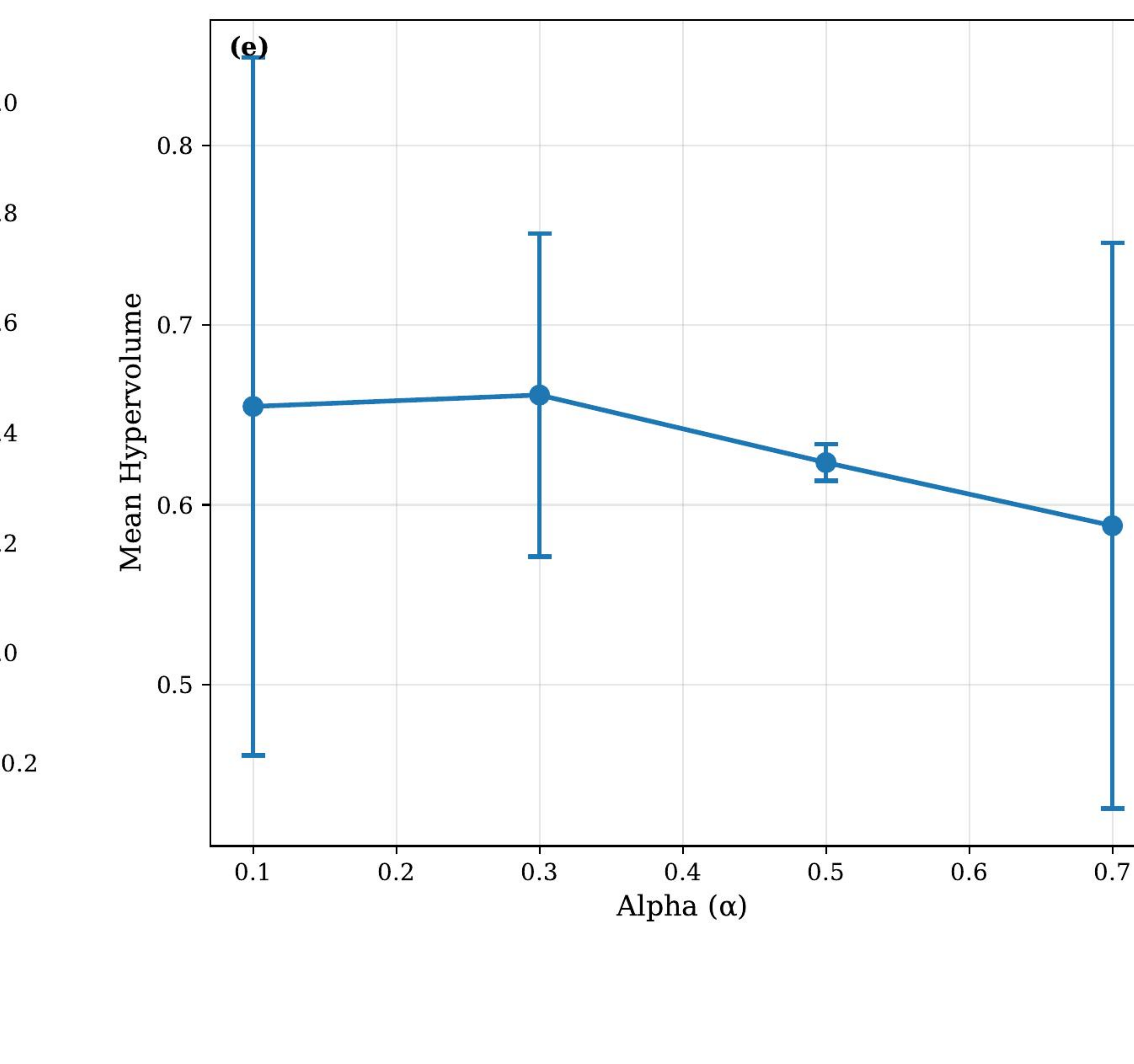}}
  \hfil
  \subfloat[Spread vs alpha.]{\includegraphics[width=.3\textwidth]{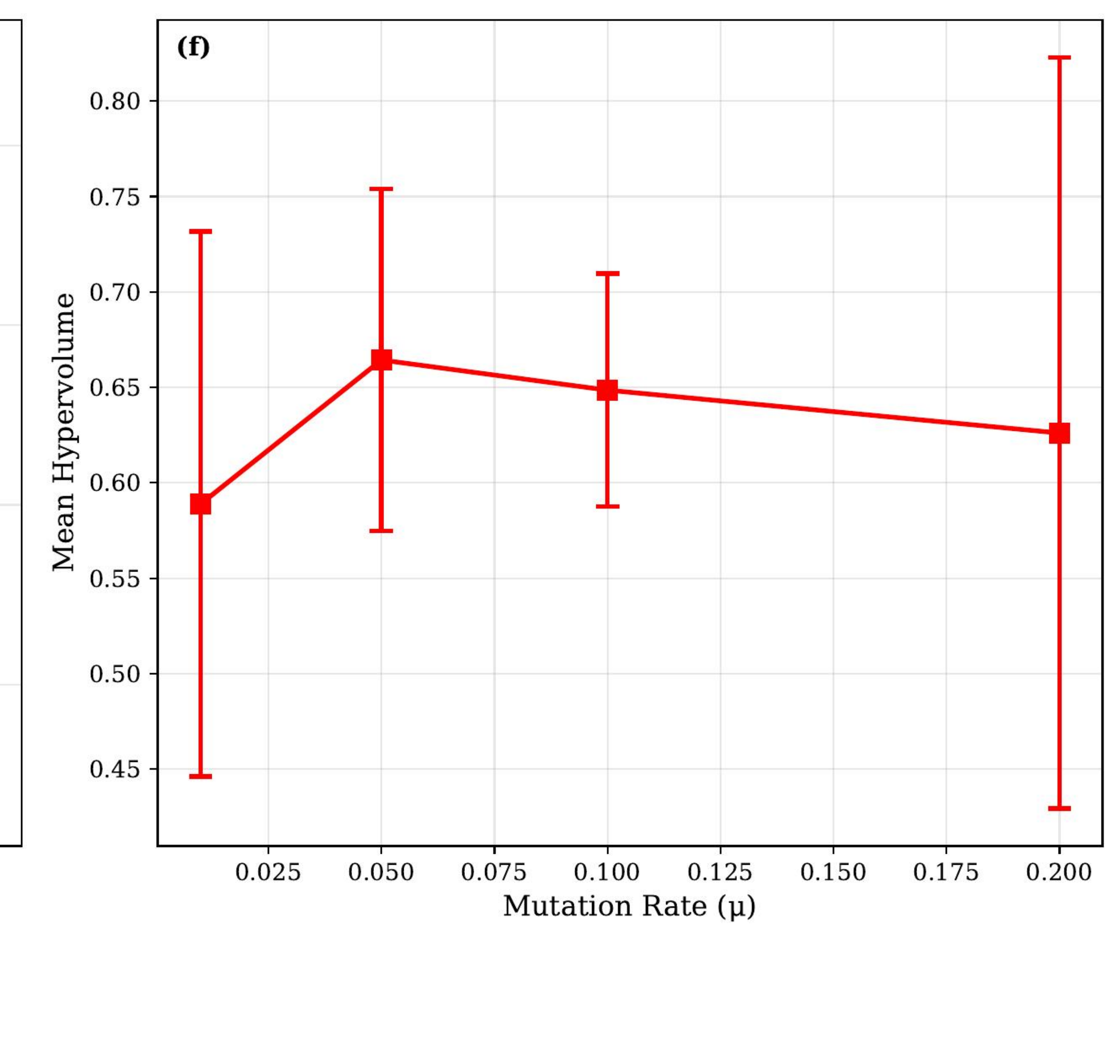}}
  \caption{Parameter sensitivity of GIS-moGA summarized across the 64 scenarios. (a)--(c) Mean hypervolume as a function of mutation rate, population size, and crossover $\alpha$, respectively. (d)--(f) Corresponding generalized spread ($\Delta$) values. The hypervolume panels reveal a broad optimal region spanning mutation rates 0.05--0.20 and populations $\geq 200$, with $\alpha$ exerting a weaker, roughly symmetric effect around 0.3.}
  \label{fig:param_summary}
\end{figure*}

Figure~\ref{fig:param_summary} shows the parameter sensitivity summary panels.

The six panels decompose parameter sensitivity along the three varied parameters. The hypervolume panels confirm the broad optimal region spanning mutation rates 0.05--0.20 and populations $\geq 200$, with $\alpha$ exerting a weaker, roughly symmetric effect around 0.3. The spread panels show that the same high-mutation, large-population region also tends to produce more uniform Pareto-front coverage, although the relationship is noisier than for hypervolume.

\begin{figure}[!t]
  \centering
  \includegraphics[width=.85\columnwidth]{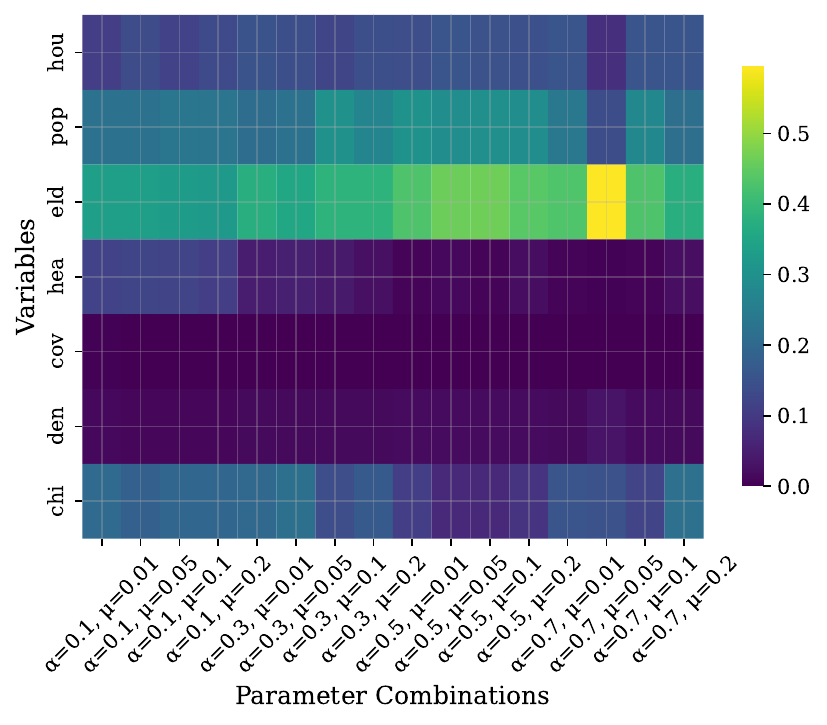}
  \caption{Mean optimized weights grouped by mutation rate (rows) and population size (columns) for the seven thematic layers (\emph{hou}, \emph{pop}, \emph{eld}, \emph{hea}, \emph{cov}, \emph{den}, \emph{chi}). The epidemiological layers \emph{cov} and \emph{den} retain consistently high weights across groupings, while the demographic layers vary more with mutation rate; larger populations yield smoother, more stable allocations.}
  \label{fig:param_weights}
\end{figure}

The heatmap groups mean weights by mutation rate and population size. The epidemiological layers \emph{cov} and \emph{den} retain consistently high mean weights across all groupings, while the demographic layers (\emph{hou}, \emph{pop}, \emph{eld}) show wider variation in response to the mutation rate. Larger populations (200, 400) produce more stable allocations, visible as smoother rows, whereas the 50-individual groupings exhibit noisier weight assignments.

\setcounter{figure}{0}
\setcounter{table}{0}
\section{Scenario 16 Extended Diagnostics}

\begin{figure}[!t]
  \centering
  \includegraphics[width=.85\columnwidth]{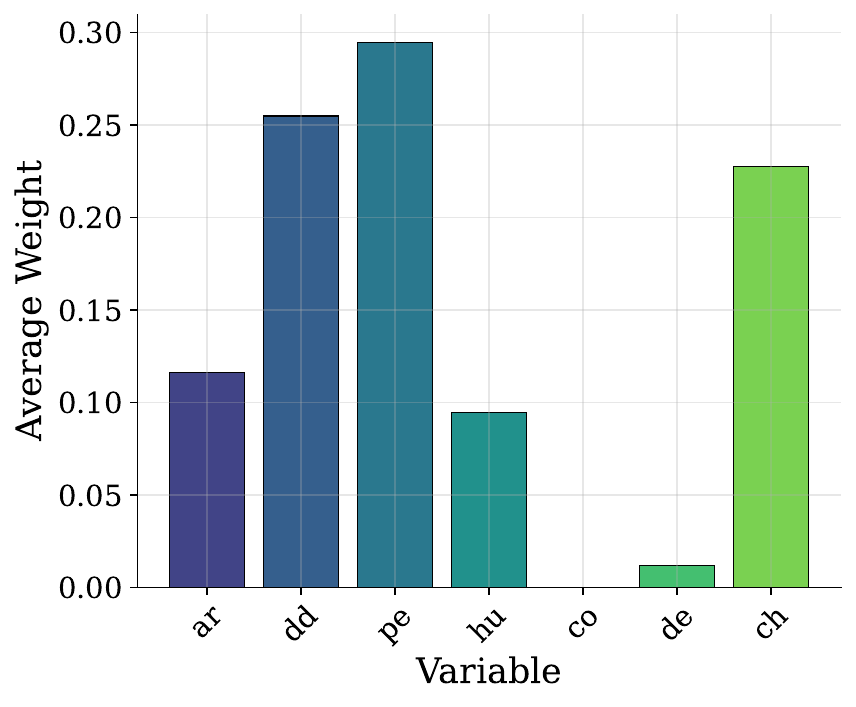}
  \caption{Per-layer weight distributions for the best-performing configuration (Scenario 16: $\alpha = 0.1$, mutation = 0.20, population = 400) across the final Pareto front. The epidemiological layers \emph{cov} and \emph{den} receive the largest median weights; the interquartile ranges are narrower than the across-scenario aggregate (Figure~\ref{fig:weight_heatmaps}), reflecting the stabilizing effect of the large population and high mutation rate within this single run.}
  \label{fig:scenario16_extended}
\end{figure}

The extended weight distributions for Scenario 16 show that the algorithm converges to a non-uniform allocation: the epidemiological layers \emph{cov} and \emph{den} receive the largest median weights, while the demographic layers span wider interquartile ranges. The spread of each distribution is narrower than the across-scenario aggregate in Figure~\ref{fig:weight_heatmaps}, reflecting the stabilizing effect of the large population (400) and high mutation rate (0.20) within this single run.

\setcounter{figure}{0}
\setcounter{table}{0}
\section{Spatial Cluster Diagnostics}

\begin{figure*}[!t]
  \centering
  \subfloat[With vulnerability map.]{\includegraphics[width=.45\textwidth]{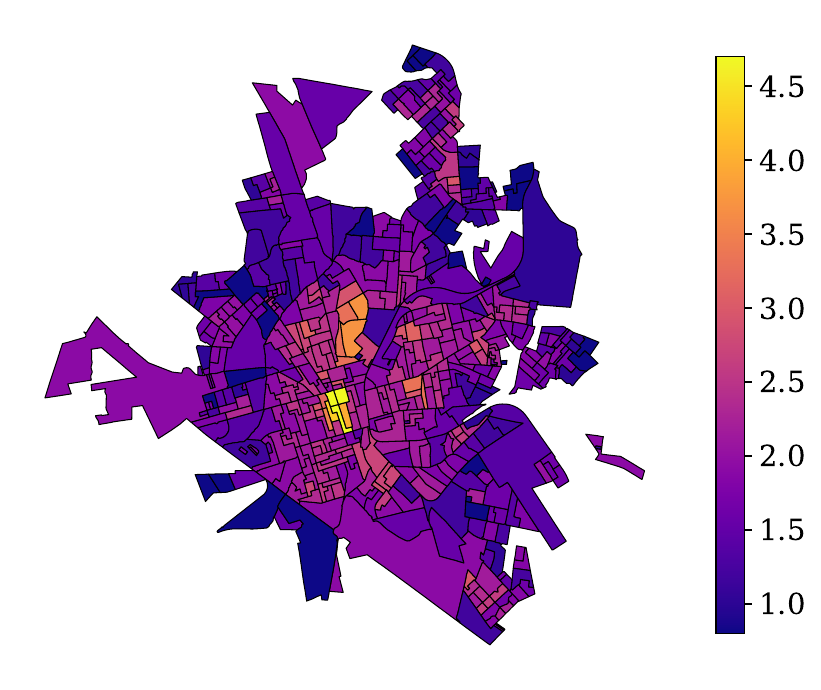}}
  \hfil
  \subfloat[Cluster-only view.]{\includegraphics[width=.45\textwidth]{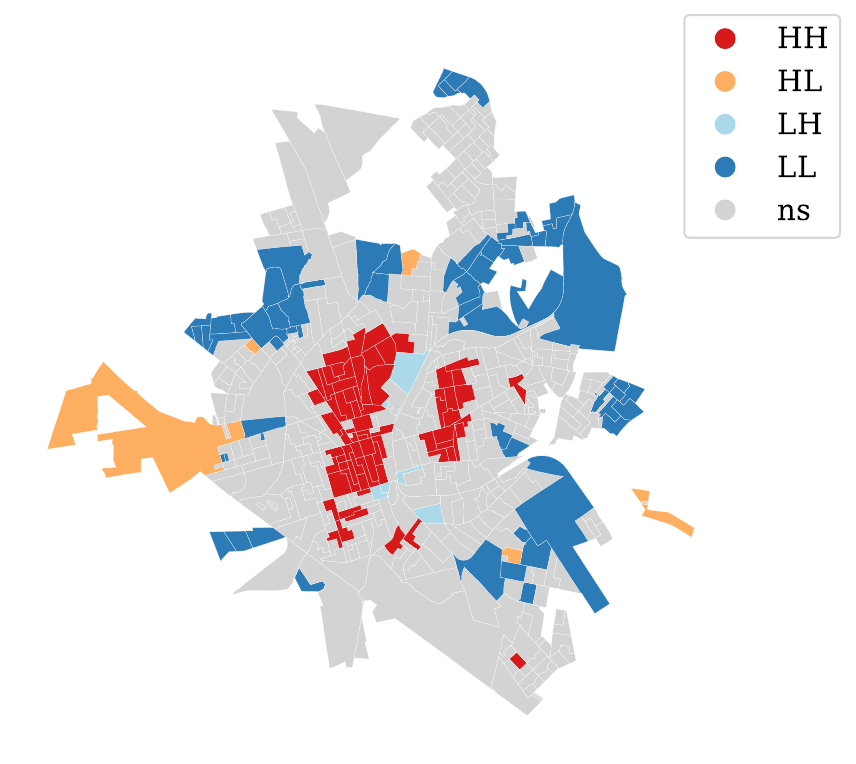}}
  \caption{LISA cluster diagnostics for the Scenario 16 composite vulnerability map. (a) Combined vulnerability scores overlaid with significant LISA cluster categories. (b) Cluster-only view retaining solely the statistically significant categories (Local Moran's $I$, $p < 0.05$, Bonferroni correction). High-high clusters concentrate in the densely populated urban core, low-low clusters mark the rural periphery, and spatial outliers appear along the urban--rural transition; 23\% of census tracts belong to significant clusters.}
  \label{fig:lisa_clusters}
\end{figure*}

The two panels isolate the significant LISA clusters for the Scenario 16 composite map. The vulnerability map overlay shows that high-high clusters concentrate in the densely populated urban core and peripheral zones with high disease incidence. The cluster-only view isolates the statistically significant categories (Local Moran's $I$, $p < 0.05$, Bonferroni correction), making the 23\% of census tracts belonging to significant clusters visible at a glance. Low-low clusters mark the rural, low-density periphery, while high-low and low-high spatial outliers appear along the urban--rural transition.

\bibliographystyle{IEEEtran}
\bibliography{references}

\end{document}